\documentclass{bmvc2k}

\usepackage{xcolor}
\usepackage{listings}
\usepackage{scalerel}
\usepackage{multirow}
\usepackage{adjustbox}
\usepackage{makecell}
\usepackage[most]{tcolorbox}
\tcbset{on line, 
        left=1pt,right=1pt,top=2pt,bottom=2pt,
        colback=red,boxrule=0pt,frame hidden,boxsep=0pt,
        highlight math style={enhanced}
        }
\usepackage{cascadia-code}
\usepackage[T1]{fontenc}
\usepackage{pifont}
\usepackage{nicematrix}
\usepackage{multicol}
\usepackage[longend,ruled,linesnumbered]{algorithm2e}
\usepackage{graphicx}
\usepackage{tabularx}
\usepackage{booktabs}
\usepackage{subcaption}
\usepackage{wrapfig}

\title{Solving Zero-Shot 3D Visual Grounding as Constraint Satisfaction Problems}

\addauthor{Qihao Yuan}{qihao.yuan@rug.nl}{1}
\addauthor{Kailai Li}{kailai.li@rug.nl}{1,*}
\addauthor{Jiaming Zhang}{jiaming.zhang@kit.edu}{2,*}
\addauthor{\\\footnotesize{$^*$Corresponding authors with equal contribution.}}{}{}

\addinstitution{
 Bernoulli Institute for Mathematics, Computer Science and Artificial Intelligence,\\
 University of Groningen,\\
 Groningen, The Netherlands
}
\addinstitution{
 Computer Vision for Human-Computer Interaction Lab (cv:hci),\\
 Karlsruhe Institute of Technology,\\
 Karlsruhe, Germany
}

\runninghead{QIHAO YUAN, KAILAI LI, JIAMING ZHANG}{CSVG}

\def\eg{\emph{e.g}\bmvaOneDot}

\begin{document}

\maketitle

\begin{abstract}
3D visual grounding (3DVG) aims to locate objects in a 3D scene with natural language descriptions. Supervised methods have achieved decent accuracy, but have a closed vocabulary and limited language understanding ability. Zero-shot methods utilize large language models (LLMs) to handle natural language descriptions, where the LLM either produces grounding results directly or generates programs that compute results (symbolically). In this work, we propose a zero-shot method that reformulates the 3DVG task as a Constraint Satisfaction Problem (CSP), where the variables and constraints represent objects and their spatial relations, respectively. This allows a global symbolic reasoning of all relevant objects, producing grounding results of both the target and anchor objects. Moreover, we demonstrate the flexibility of our framework by handling negation- and counting-based queries with only minor extra coding efforts. Our system, Constraint Satisfaction Visual Grounding (CSVG), has been extensively evaluated on the public datasets ScanRefer and Nr3D datasets using only open-source LLMs. Results show the effectiveness of CSVG and superior grounding accuracy over current state-of-the-art zero-shot 3DVG methods with improvements of $+7.0\%$ (Acc@0.5 score) and $+11.2\%$ on the ScanRefer and Nr3D datasets, respectively. The code of our system is available \href{https://asig-x.github.io/csvg_web}{here}.
\end{abstract}

\begin{figure}
\begin{center}
\includegraphics[width=1\linewidth]{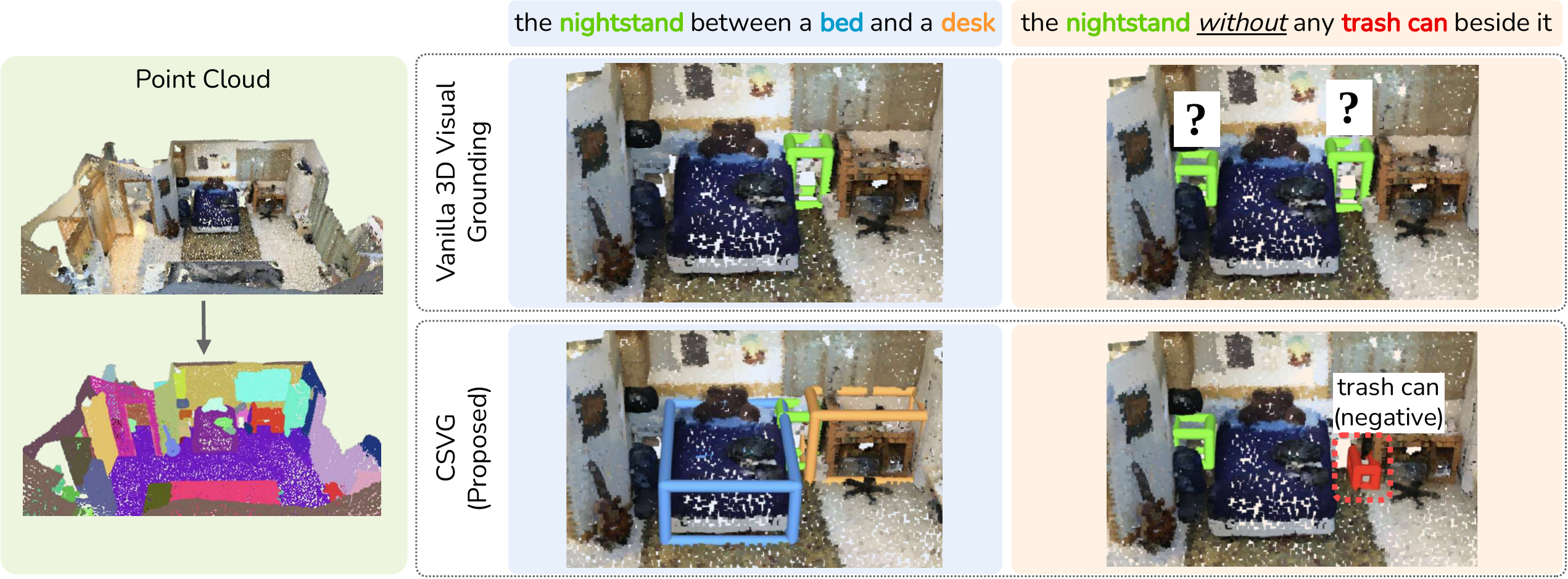}
\end{center}
\caption{Comparison of conventional 3DVG and our proposed Constraint Satisfaction Visual Grounding (CSVG) framework. On the first query, standard 3DVG only finds the target (\textit{\textcolor[HTML]{6ecd12}{nightstand}}), while our CSVG can locate and exploit anchor objects (\textit{\textcolor[HTML]{029ccc}{bed}} and \textit{\textcolor[HTML]{fc9e3f}{desk}}). The second query has a negation (``\textit{without any trash can beside it}''). Here, the standard 3DVG fails, whereas our method successfully locates the target (\textit{\textcolor[HTML]{ed1f1d}{trash can}}).}
\label{fig:teaser}
\end{figure}

\section{Introduction}
\label{sec:intro}


The 3D Visual Grounding (3DVG) task aims to locate objects in 3D scenes based on natural language descriptions \cite{vg_survey}.
To solve the 3DVG task, many supervised methods based on graph neural networks (GNNs)~\cite{gnn,tgnn} and transformers~\cite{transformer,3dvgtransformer,butddetr,3drpnet,3dsps} have been proposed. 
These methods achieve state-of-the-art performance~\cite{eda,g3lq,3dvista} on the datasets used for training, yet often have trouble handling the diversity of natural language descriptions~\cite{can3dvlm} due to limited model size and language patterns in the training data. 

To address these issues, zero-shot methods have been proposed~\cite{llmgrounder,visualprogramming,vlmgrounder}, which exploit the superior natural language understanding and reasoning abilities of Large Language Models (LLMs)~\cite{gpt4o,llama3}. 
In order to integrate LLMs into 3DVG systems, the point cloud must be converted into a textual format, with point cloud segmentation and object classification \cite{mask3d}.
This has the advantage that LLMs can handle very complex natural language queries, and no training dataset is required for the 3DVG task.

Given a query involving multiple object relationships (\eg, \textit{``the chair with two trash bins beside it''}), existing method like ZSVG3D~\cite{visualprogramming} reasons about the individual object relationship locally (\textit{``the chair with one trash bin beside it''}) one at a time, which lacks a global perspective all relationships as a whole. This often results in incorrect grounding given non-trivial queries in complex 3D scenes (Fig.~\ref{fig:teaser}).
In this paper, we introduce a novel zero-shot 3D visual grounding framework, Constraint Satisfaction Visual Grounding (CSVG), by enabling global interpretation of object spatial relations through the formulation of Constraint Satisfaction Problem (CSP)~\cite{cspbook}. 
Our contributions are summarized as follows:\\
\textbf{(1)} We formulate the 3DVG task as a constraint satisfaction problem, which allows for global reasoning about spatial relations of all relevant objects at once, leading to grounding of both the target and anchor objects.\\
\textbf{(2)} We demonstrate the flexibility of our framework by extending it to handle negation- and counting-based queries with only minor code modifications. In contrast, supervised methods would require a substantial amount of additional training data for theses capabilities.\\
\textbf{(3)} We evaluate our system on the well-known Nr3D~\cite{referit3d} and ScanRefer~\cite{scanrefer} datasets, achieving increases in accuracy (Acc@0.5 score) over the state of the art (ZSVG3D~\cite{visualprogramming}) by a margin of $+11.2\%$ on the Nr3D and $+7.0\%$ on the ScanRefer dataset.
\section{Related Work}
\label{sec:related_work}

\begin{figure*}[t]
\begin{center}
\includegraphics[width=1.0\linewidth]{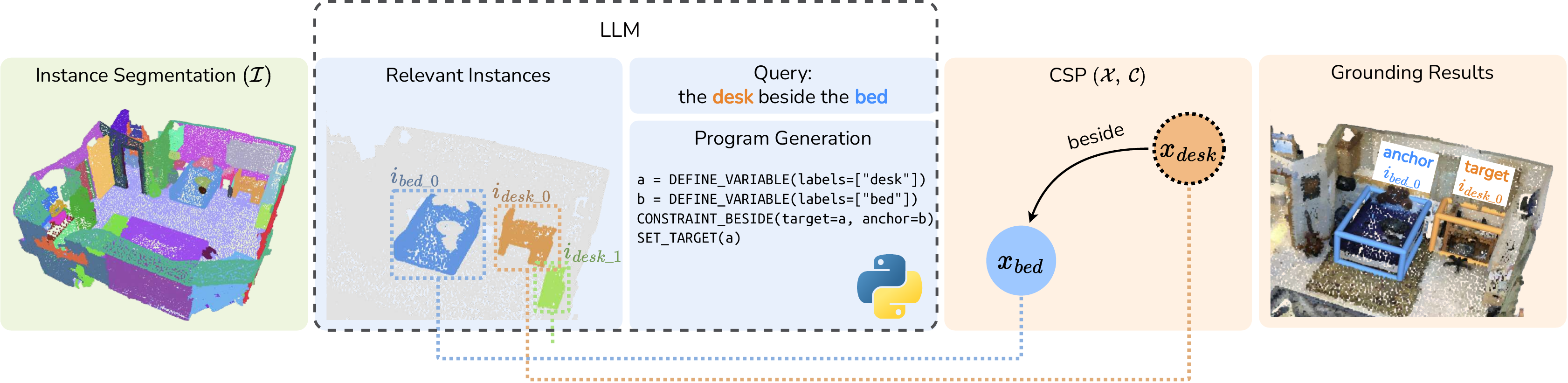}
\end{center}
\caption{Pipeline of CSVG framework. Given a point cloud, we first perform instance segmentation to obtain a list ${\mathcal{I}}$. Based thereon, an LLM is then employed to generate a Python program, and the CSP ($\mathcal{X}, \mathcal{C}$) is built (target indicated by dotted circle). Therein, variables represent objects, and constraints are created based on the object spatial relations. Solving the CSP then produces the 3D grounding results.}
\label{fig:pipeline}
\end{figure*}

\noindent\textbf{Two-Stage Supervised 3DVG.} Many early 3DVG methods employ a two-stage architecture, consisting of an object proposal module followed by a multimodal fusion module, where fused features are used to predict scores for object proposals. ScanRefer~\cite{scanrefer} applies simple Multi-Layer Perceptrons (MLP) in the fusion module, while ReferIt3D~\cite{referit3d} leverages Dynamic Graph-Convolutional Network (DGCN)~\cite{dgcn}. TGNN~\cite{tgnn} further explores Graph Neural Networks (GNN) for multimodal feature fusion. To improve grounding accuracy, 3D-VisTA~\cite{twostage4} trains a unified model across multiple 3D tasks, and Multi3DRefer~\cite{multi3drefer} extends the 3DVG task to handle cases with zero, one, or multiple targets. A major drawback of two-stage methods is the separation of the object proposal and multimodal fusion, which prevents language features from guiding object detection, often resulting in missed objects.

\noindent\textbf{Single-Stage Supervised 3DVG.} Single-stage 3DVG methods train models end-to-end, directly predicting target bounding boxes without relying on a pretrained object proposal module. For instance, 3D-SPS~\cite{3dsps} uses a coarse-to-fine key-point selection mechanism to regress target bounding boxes, while BUTD-DETR~\cite{butddetr} employs a cross-modal encoder to encode visual, linguistic, and object proposal features, enabling it to decode a bounding box for each object in the query. EDA~\cite{eda} further parses queries into semantic components and aligns them with object proposals to guarantee correct anchor object selection. 3DRP-Net~\cite{3drpnet} introduces modules to model relative positions and orientations between object pairs, and G3-LQ~\cite{g3lq} makes extra efforts to ensure semantic-geometric consistency. Single-stage methods generally outperform two-stage methods; however, they often struggle with diverse natural language descriptions and may suffer from accuracy loss due to limited data availability in the training data.

\noindent\textbf{Zero-Shot 3DVG.} To enhance the generalizability of 3DVG methods, several zero-shot pipelines have been proposed using LLMs to leverage their strong natural language understanding and reasoning capabilities. For instance, LLM-Grounder~\cite{llmgrounder} prompts the LLM to generate a grounding plan to locate the target object, utilizing open-vocabulary scene understanding models~\cite{openscene,lerf} to extract objects from the point cloud. However, it suffers from slow inference speed due to the complex grounding plans and reasoning processes generated by the LLM. More recently, VLM-Grounder~\cite{vlmgrounder} was proposed to leverage the image understanding ability of Visual Language Models (VLMs)~\cite{llava,gpt4o} for 3DVG in RGB-D sequences. It detects relevant objects and prompts the VLM to locate the target object, both of which are performed in 2D images. The bounding box is then predicted by projecting the depth values of the target object into 3D. Inspired by approaches in the 2D case~\cite{visproj,vipergpt}, ZSVG3D~\cite{visualprogramming} prompts an LLM to generate Python programs to solve the 3DVG task. Since the LLM only needs to generate a relatively short Python program, inference time is reduced, making evaluations on larger datasets more feasible. However, ZSVG3D only considers one single spatial relation at a time, which can lead to incorrect results in more complex scenarios. In contrast, our CSVG transforms the 3DVG task into a CSP and find a global solution that satisfies all spatial relations at the same time, thereby achieving higher grounding accuracy.
\definecolor{codegreen}{rgb}{0,0.6,0}
\definecolor{codegray}{rgb}{0.5,0.5,0.5}
\definecolor{codepurple}{rgb}{0.58,0,0.82}
\definecolor{backcolour}{rgb}{0.95,0.95,0.92}

\lstdefinestyle{mystyle}{
    backgroundcolor=\color{backcolour},   
    commentstyle=\color{codegreen},
    keywordstyle=\color{magenta},
    numberstyle=\tiny\color{codegray},
    stringstyle=\color{codepurple},
    basicstyle=\ttfamily\footnotesize,
    breakatwhitespace=false,         
    breaklines=true,                 
    captionpos=b,                    
    keepspaces=true,                 
    numbers=left,                    
    numbersep=5pt,                  
    showspaces=false,                
    showstringspaces=false,
    showtabs=false,                  
    tabsize=2
}
\lstset{style=mystyle}

\section{Proposed Method}
\label{sec:method}

In this section, we elaborate the proposed CSVG framework. We first introduce CSP briefly, and then give an overview and go into the details of the proposed system.

\subsection{Preliminary}
\label{sec:csp_intro}

Conceptually, a Constraint Satisfaction Problem (CSP) consists of the following components~\cite{cspbook}: 
\textbf{variables} $\mathcal{X} = \{x_1, x_2, \ldots, x_n\}$; 
\textbf{domains} $\mathcal{D} = \{\mathcal{D}_1, \mathcal{D}_2, \ldots, \mathcal{D}_n\}$,
where each domain $\mathcal{D}_i$ denotes a finite set of possible values that variable $x_i$ may take;
\textbf{constraints} $\mathcal{C} = \{c_1, c_2, \ldots, c_m\}$,
where each constraint $c_j$ limits the values that a subset of variables may simultaneously take.
Each constraint $c_j$ is a pair $\langle\mathcal{S}_j,\mathcal{R}_j\rangle$, where
subset $\mathcal{S}_j\subseteq\mathcal{X}$ defines variables involved in this constraint, and relation $\mathcal{R}_j$ specifies the allowed combinations of values for variables in $\mathcal{S}_j$.

A solution to the CSP is an assignment to all variables in $\mathcal{X}$ such that each constraint $c_j$ is satisfied. A constraint is satisfied if the assignment of $\mathcal{S}_j$ fulfills the relation $\mathcal{R}_j$. For instance, suppose $\mathcal{S}_j=\{x_1,x_2\}$ and $\mathcal{R}_j=\langle(x_1,x_2), x_1 > x_2 \rangle$, then assignment $\{x_1=2, x_2=1\}$ satisfies $c_j$, while assignment $\{x_1=1, x_2=2\}$ does not.
Numerous algorithms exist for solving CSPs~\cite{cspbook}, e.g., backtracking, constraint propagation and local search.

\subsection{3DVG through Constraint Satisfaction}
\label{sec:method_overview}
Our proposed method reformulates the 3DVG task as a CSP, where the solution is derived by simultaneously considering all relevant objects and spatial relations. As illustrated in Fig.~\ref{fig:pipeline}, given the query \textit{``the desk beside the bed''}, we first perform instance segmentation on the input point cloud, producing a list of instances $\mathcal{I} = \{i_1, i_2, ..., i_L\}$, each of which ($i_k$) has a point cloud $pc\left(i_k\right)$ and a label $label\left(i_k\right)$. Finally, the query and $\mathcal{I}$ are given to the LLM to generate a Python program constructing the CSP for 3DVG, including the identified variables $\mathcal{X}$ and related constraints $\mathcal{C}$, with each variable $x_j\in\mathcal{X}$ given a label by LLM.
The reason for choosing Python as the generated language is $\left(1\right)$ LLM could potentially generate Python programs better as it is a popular language and has a lot of code samples on the Internet $\left(2\right)$ it provides easy access to the interpreter, eliminating the need to implement an interpreter for a custom language designed for this task.

Specifically, we construct a prompt containing the query, a list of labels (object category names) of all instances in the scene, and instructions.
Each variable is assigned one (or in rare cases, multiple) instance labels, and the domain $\mathcal{D}$ of that variable consists of all instances in $\mathcal{I}$ with the assigned label(s). 
For the constraints, we have a predefined set of constraints for spatial relationships, and the LLM is asked to only choose from them, instead of inventing constraints on-the-fly.
The argument is that there is only a limited number of words for spatial relationships that are commonly used, e.g., "on", "under", etc. Having a predefined set simplifies the system and faciliates validation of the generated CSP.
More details will be given later and also in the supplementary material.

As shown in Fig.~\ref{fig:pipeline}, given the query and instance segmentation, the LLM implicitly determines the relevant instances, including a bed and two desks, namely, $\{i_{bed\_0}, i_{desk\_0}, i_{desk\_1}\}$. Further, the LLM generates a CSP that consists of variables $\mathcal{X} = \{x_{desk}, x_{bed}\}$, including a desk and a bed, and a constraint $\mathcal{C} = \{c_{beside}\}$ stating that the desk is beside the bed. Given the instances $\mathcal{I}$, the domains of $x_{desk}$ and $x_{bed}$ follow $\mathcal{D}_{desk} = \{i_{desk\_0}, i_{desk\_1}\}$ and $\mathcal{D}_{bed} = \{i_{bed\_0}\}$, respectively. Solving the CSP (via backtracking) yields an assignment $\{x_{desk}=i_{desk\_0},x_{bed}=i_{bed\_0}\}$. In this manner, we obtain not only the instance assigned to the target variable $x_{desk}$, but also the anchor variable $x_{bed}$, both of which are involved in the grounding process. The LLM prompts used to generate Python programs are provided in the supplementary material.

\begin{figure}[t]
\begin{center}
\begin{tabular}{cc}
\makecell{\bmvaHangBox{\includegraphics[width=0.5\textwidth]{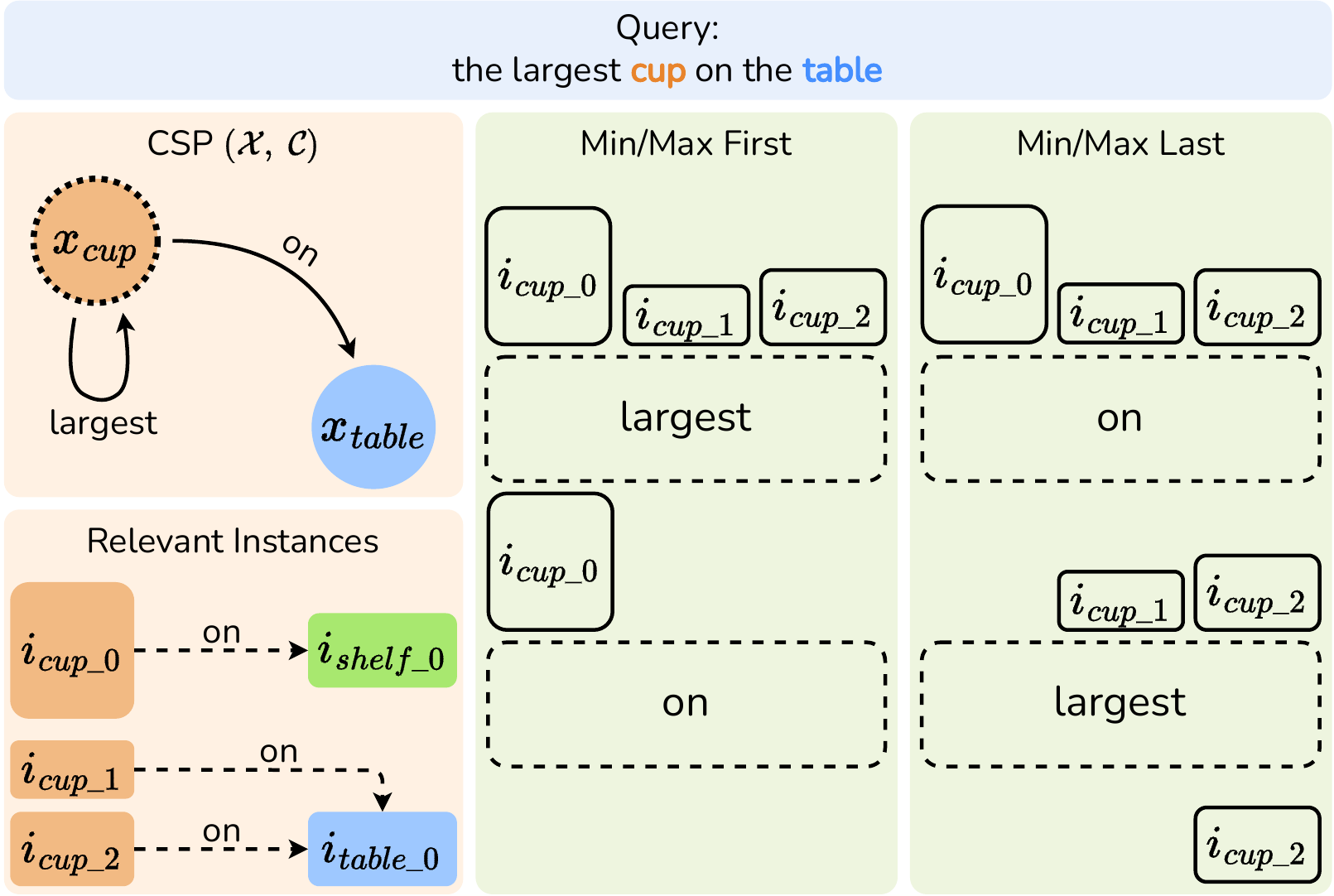}}}&
\makecell{\bmvaHangBox{\includegraphics[width=0.45\textwidth]{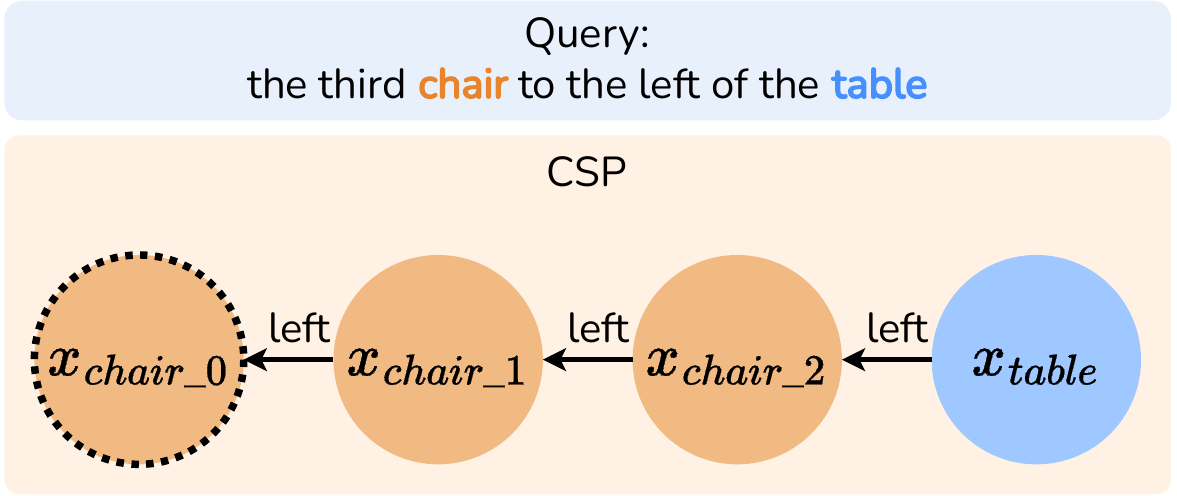}}}\\
(a)&(b)
\end{tabular}
\end{center}
\caption{\textbf{(a)} Illustration of combining min/max and spatial constraints. Suppose we have one cup ($i_{cup\_0}$) on the shelf ($i_{shelf\_0}$) and two cups ($i_{cup\_1}$ and $i_{cup\_2}$) on the table ($i_{table\_0}$). A CSP is formulated given the query \textit{``the largest cup on the table''}. The size of the orange boxes enclosing the cups indicates their respective size, with $i_{cup\_0}$ being the largest. Selecting the largest cup ($i_{cup\_0}$) first leads to failure, as $i_{cup\_0}$ is on the shelf, not the table. However, if we first identify the cups on the table ($i_{cup\_1}$ and $i_{cup\_2}$) and then choose the largest one among them, we obtain the correct result ($i_{cup\_2}$). \textbf{(b)} A counting-based query and CSP. We add an in-context example for the LLM prompt, enabling it to generate appropriate structure when encountering terms like ``second'', ``third'' in the query. Consequently, the CSP solution will automatically locate the third chair to the left of the table.}
\label{fig:minmax_demo_and_counting_based_demo}
\end{figure}

\subsection{CSP Formulation and Solution}
\label{sec:csp_form}

\noindent\textbf{Variables.} Formulating 3DVG as a CSP requires defining variables $\mathcal{X}$, which are generated by prompting the LLM with the query and available instances $\mathcal{I}$. The domain of a variable $x_j\in\mathcal{X}$ is implicitly determined by the subset of instances with the same label, namely, $\mathcal{D}_j = \{i_k\in\mathcal{I}\,\vert\,label(i_k)=label(x_j)\}$. In certain cases, we may also specify multiple labels of similar meaning (\eg, \textit{``dishwasher''} and \textit{``dish washing machine''}) to cope with noise in the labels from instance segmentation. For this, the LLM generates a set of labels $label\_{set}(x_j)$ for the variable $x_j$, and its domain becomes $\mathcal{D}_j=\{i_k\in\mathcal{I}\,\vert\,label(i_k) \in label\_set(x_j)\}$. 

\noindent\textbf{Spatial Constraints.} Regarding real-world scenarios, most constraints we consider involve two or more instances of certain spatial relations in the 3D scene, \eg, $above\left(target,\;anchor\right)$, $right\left(target,\;anchor\right)$ or $between\left(target,\;anchor\_0,\;anchor\_1\right)$. The criteria for constraint satisfaction are based on the center coordinates or bounding box size of the involved instances, with predefined thresholds. Similar to ZSVG3D~\cite{visualprogramming}, we set up the viewpoint at the center of the scene for view-dependent relations (such as $left$ and $right$).

\noindent\textbf{Min/Max Constraints.} To enhance the expressiveness of our CSP formulation, we introduce min/max constraints to extend the semantics of classic CSP for addressing more complex scenarios. From instances with identical labels, a min/max constraint is generated by LLM to select a single instance based on a pre-defined score function. For example, if the query indicates \textit{``the largest cup''}, the score function \textit{``size''} will be applied to the variables of label \textit{``cup''} to select the desired variable. Theoretically, combining min/max and spatial constraints introduces additional complexities due to the freedom in ordering the different types of constraints. In practice, we have found that applying min/max constraints at the final step yields the best semantic accuracy, and we adopt this strategy in CSVG. A concrete example of this is given in Fig.~\ref{fig:minmax_demo_and_counting_based_demo}~(a).

\noindent\textbf{Solving CSP.} We employ the backtracking algorithm to search for solutions to a CSP~\cite{cspbook}, with modification to incorporate min/max constraints. More details on solving CSP is provided in the supplementary material, where we also elaborate the Python program generation with additional CSP examples. 

\noindent\textbf{Heuristic for Solution Selection.} Given our CSP formulation, multiple solutions may exist. In such cases, we apply a  \textit{heuristic} by computing the average distance between each pair of objects in a solution and selecting the one with the minimum average distance.

\subsection{Pipeline Extensions}
\label{sec:csp_ext}

To showcase the flexibility and technical potential of CSVG, we provide two extensions to CSP formulation for addressing negation- and counting-based queries.\\

\noindent\textbf{Negation-Based Queries.} For queries with negative statements, we allow a CSP variable to be marked as ``negative'' by the LLM. In the CSP solver, we iteratively generate all possible solutions and validate each one individually. When encountering a negative variable $x_j$, we test each instance $i_k$ in the variable's domain $\mathcal{D}_j$ and discard the solution if any constraint on $x_j$ is satisfied when assigning $i_k$ to $x_j$. For instance, given the query \textit{``the nightstand without any trash can beside it''}, when testing all combinations of assignments, if a trash can instance is found near the nightstand instance, the solution is considered invalid, which aligns with the semantics of the query. Supporting negation-based queries only requires a minor modification to the CSP solver, along with a function to define negative variables. This is far more efficient than collecting dedicated data and retraining a model in supervised learning.


\noindent\textbf{Counting-Based Queries.} To handle counting-based queries, we simply add an example to the prompt for the LLM, as this scenario is already expressible with our proposed CSP formulation. Suppose we have a query \textit{``The third chair to the left of the table''}. This statement indicates that two additional chairs must exist to the right of the target. As shown in Fig.~\ref{fig:minmax_demo_and_counting_based_demo}~(b), we explicitly create a variable for each chair. The CSP solver ensures that each chair is assigned a unique instance, thereby identifying the third chair to the left of the table as the target, as specified in the query.
\section{Experiments and Results}
\label{sec:experiments}
 
\noindent\textbf{Datasets.} We evaluate our approach using the widely adopted Nr3D~\cite{referit3d} and ScanRefer~\cite{scanrefer} datasets. For Nr3D, we report scores on different partitions of the validation set; for ScanRefer, we report the accuracies on the entire validation set.

\noindent\textbf{Evaluation metrics.} On the Nr3D dataset, we use ground truth segmentations (so do methods we compare to), and the performance metric is ratio of queries for which the target is correctly predicted. On the ScanRefer dataset, we use the Acc@0.25 and Acc@0.5 metrics on the ScanRefer dataset, which calculate the percentages of predicted bounding boxes with an IoU (Intersection over Union) greater than 0.25 and 0.5, respectively, w.r.t. the ground truth.

\noindent\textbf{Baselines.} We compare our method with both supervised and zero-shot approaches from the current state of the art. For supervised methods, we include ScanRefer~\cite{scanrefer} and ReferIt3D~\cite{referit3d}, which propose both the datasets and models, and 3DVG-Transformer~\cite{3dvgtransformer}, BUTD-DETR~\cite{butddetr}, and EDA~\cite{eda}, all of which show notable improvements over previous works. For zero-shot methods, we include LLM-Grounder~\cite{llmgrounder}, ZSVG3D~\cite{visualprogramming}, and VLM-Grounder~\cite{vlmgrounder}, all of which have made their code publicly available.

\noindent\textbf{LLMs.} LLM-Grounder and ZSVG3D utilize GPT-4~\cite{gpt4} as their LLM agent, while VLM-Grounder employs GPT-4V~\cite{gpt4v}. In this paper, we deploy Mistral-Large-2407 (123B parameters)~\cite{mistral} locally using the SGLang framework~\cite{sglang}, which enables high inference throughput. We do not perform full evaluation of our system using any of the ChatGPT models due to potential high costs associated with their use, as our system has significantly larger input and output lengths compared to ZSVG3D. However, we argue that if our method works well with a smaller model, scaling it up to a larger, more powerful model would likely improve performance without introducing issues.

\subsection{Main Results}

\begin{table}
\begin{center}
\begin{NiceTabular}{@{}lccccccc@{}}
\toprule
\textbf{Method} & \textbf{Supervision} & \textbf{LLM} & \textbf{Easy} & \textbf{Hard} & \textbf{Dep.} & \textbf{Indep.} & \textbf{Overall} \\
\midrule 
ReferIt3D~\cite{referit3d} & fully & - & 43.6 & 27.9 & 32.5 & 37.1 & 35.6 \\
3DVG-Transformer~\cite{3dvgtransformer} & fully & - & 48.5 & 34.8 & 34.8 & 43.7 & 40.8 \\
BUTD-DETR~\cite{butddetr} & fully & - & \underline{60.7} & \underline{48.4} & 46.0& \underline{58.0} & \underline{54.6}\\
EDA~\cite{eda} & fully & - & 58.2& 46.1& \underline{50.2} & 53.1 & 52.1 \\
3D-VisTA~\cite{3dvista} & fully & - & \textbf{72.1} & \textbf{56.7} & \textbf{61.5} & \textbf{65.1} & \textbf{64.2} \\
\midrule
ZSVG3D~\cite{visualprogramming} & zero-shot & GPT-4 & 46.5 & 31.7 & 36.8 & 40.0 & 39.0 \\
VLM-Grounder~\cite{vlmgrounder} & zero-shot & GPT-4V & \underline{55.2} & \underline{39.5} & \underline{45.8} & \underline{49.4} & \underline{48.0} \\
Ours & zero-shot & Mistral-Large-2407 & \textbf{67.1}& \textbf{51.3}& \textbf{53.0}& \textbf{62.5}& \textbf{59.2}\\
\bottomrule
\end{NiceTabular}
\end{center}
\caption{3DVG results on the Nr3D \cite{referit3d} dataset. The best and second-best scores are bold and underlined, respectively.}
\label{tab:eval_nr3d}
\end{table}

\begin{table}
\fontsize{9.6}{9.6}\selectfont
\renewcommand{\arraystretch}{1.5}
\begin{center}
\resizebox{\textwidth}{!}{
\begin{NiceTabular}{@{}lcccccccc@{}}
\toprule 
\multirow{2}{*}[-\baselineskip]{\textbf{Method}} & \multirow{2}{*}[-\baselineskip]{\textbf{Supervision}} & \multirow{2}{*}[-\baselineskip]{\textbf{LLM}} & \multicolumn{2}{c}{ \textbf{Unique} } & \multicolumn{2}{c}{ \textbf{Multiple} } & \multicolumn{2}{c}{ \textbf{Overall} } \\
 & & & Acc@0.25 & Acc@0.5 & Acc@0.25 & Acc@0.5 & Acc@0.25 & Acc@0.5 \\
\hline ScanRefer~\cite{scanrefer} & fully  &--& 65.0 & 43.3 & 30.6 & 19.8 & 37.3 & 24.3 \\
3DVG-Transformer~\cite{3dvgtransformer} & fully  &--& 81.9 & 60.6 & 39.3 & 28.4 & 47.6 & 34.7 \\
BUTD-DETR~\cite{butddetr} & fully  &--& \underline{84.2} & \underline{66.3} & \underline{46.6} & \underline{35.1} & \underline{52.2} & \underline{39.8}\\
EDA~\cite{eda} & fully  &--& \textbf{85.8} & \textbf{68.6} & \textbf{49.1} & \textbf{37.6} & \textbf{54.6} & \textbf{42.3} \\
\hline
LLM-Grounder~\cite{llmgrounder} & zero-shot & GPT-4 & - & - & -  & - & 17.1 & 5.3 \\
ZSVG3D~\cite{visualprogramming} & zero-shot & GPT-4 & 63.8 & \underline{58.4} & 27.7 & 24.6 & 36.4 & 32.7 \\
VLM-Grounder~\cite{vlmgrounder}$^\dagger$ & zero-shot & GPT-4V & \underline{66.0} & 29.8 & \textbf{48.3}  & \textbf{33.5} & \textbf{51.6} & \underline{32.8} \\
Ours (Mask3D) & zero-shot & Mistral-Large-2407 & \textbf{68.8} & \textbf{61.2}& \underline{38.4}& \underline{27.3}& \underline{49.6}& \textbf{39.8}\\
\hline
Ours (GT Seg.) & zero-shot & Mistral-Large-2407 & 94.1& 93.8& 50.2& 42.8& 66.3& 61.6\\
\bottomrule
\end{NiceTabular}
}
\end{center}
\caption{3DVG results on the ScanRefer~\cite{scanrefer} validation set. The best and second best scores are embolded and underlined, repsectively.}
\label{tab:eval_scanrefer}
\end{table}


\noindent\textbf{Quantitative Results.} As shown in Table~\ref{tab:eval_nr3d}, on the Nr3D dataset, our proposed CSVG achieves the best overall score of $62.5$ on Acc@0.25 and $59.2$ on Acc@0.5 among zero-shot methods, and even surpassing several supervised methods, including BUTD-DETR and EDA. On the ScanRefer dataset (Table~\ref{tab:eval_scanrefer}), CSVG achieves higher Acc@0.5 score than any other zero-shot methods, and is comparable to BUTD-DETR, while outperforming 3DVG-Transformer, both of which are supervised methods.

\begin{figure*}[t]
\begin{center}
\includegraphics[width=1.0\linewidth]{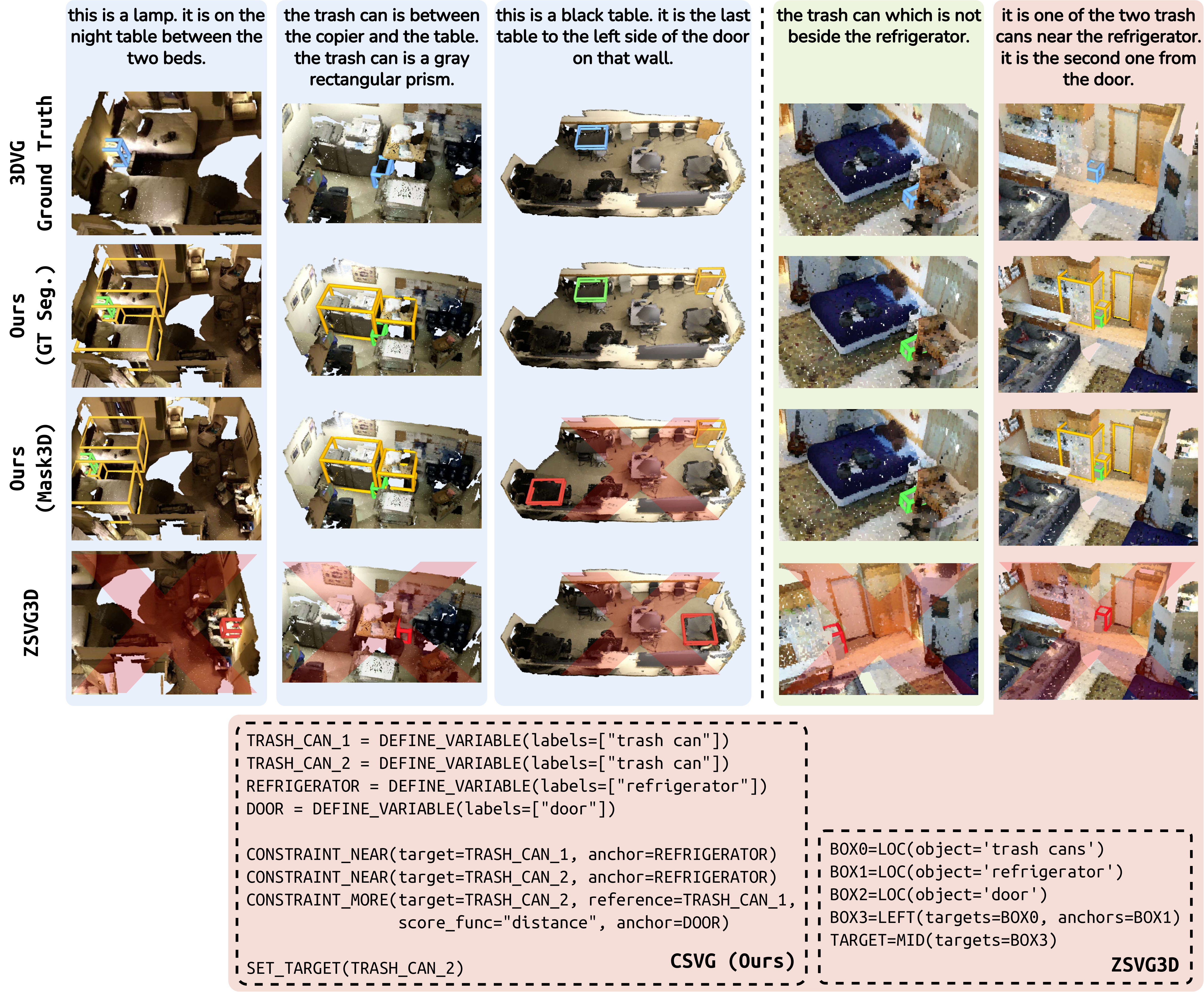}
\end{center}
\caption{Qualitative evaluation results given five queries. We denote ground truth with \textcolor[rgb]{0.4,0.7,1.0}{blue} bounding boxes. For correctly and incorrectly identified grounding targets, we use \textcolor[rgb]{0.3,1.0,0.3}{green} and \textcolor[rgb]{1.0,0.2,0.2}{red} boxes, respectively. \textcolor[rgb]{1.0,0.7,0.0}{Orange} boxes indicate anchor objects recognized by CSVG.}
\label{fig:quali_eval}
\end{figure*}

\noindent\textbf{Qualitative Results} We showcase our proposed framework with five queries as illustrated in Fig.~\ref{fig:quali_eval}. We compare our results using both Mask3D and ground truth segmentations to ZSVG3D.
For the first two queries, our system correctly identified both the target (green box) and anchor objects (blue box), while ZSVG3D failed.
However, for the third query, both our Mask3D and ZSVG3D system failed. This failure occurred because Mask3D missed the target (the upper table) during instance segmentation; in contrast, our method produced the correct result when using the ground truth segmentation. In the last two columns of Fig.~\ref{fig:quali_eval}, we demonstrate the effectiveness of extending CSVG to handle negation- and counting-based queries. In the fourth query, CSVG successfully located the trash can near the table, whereas ZSVG3D incorrectly predicted a trash can near the refrigerator as the target. For the last query, CSVG accurately grounded the three anchor objects -- the refrigerator, the door, and the first trash can from the door. On the other hand, ZSVG3D mistakenly identified the first trash can as the target. We also provide the generated programs, showing that CSVG precisely follows the intended strategy (Fig.~\ref{fig:minmax_demo_and_counting_based_demo}~(b)), while the program generated by ZSVG3D is semantically obscure.

\subsection{Ablation Study}
\label{sec:abl}
\noindent\textbf{CSP Components.} In accordance to the introduction in Sec.~\ref{sec:csp_form}, we ablate the three main CSVG components, i.e., the global satisfaction of spatial constraints, min/max constraints, and the heuristic for solution selection, using 1000 random samples from the ScanRefer validation set. For ablating the global satisfaction component, we only consider a single spatial relation at a time, similar to ZSVG3D. As shown in Table~\ref{tab:abl_comp}~(a), removing all three components leads to inferior grounding accuracies compared to ZSVG3D evaluated on the whole ScanRefer validation set. Adding our CSP solver with global constraint satisfaction improves the Acc@0.5 score to $33.2$, outperforming ZSVG3D ($32.7$). Incorporating the min/max constraints provides a further modest boost in both scores, because queries requiring these constraints are relatively rare, the overall impact remains limited. In contrast, applying the solution selection heuristic yields a more notable improvement in accuracy. Since we are using only 1000 samples, potential bias in this subset of queries could affect the results, which contributes to lower scores observed here compared to our full evaluation results in Table~\ref{tab:eval_scanrefer}.

\begin{table}
\begin{center}
\begin{tabular}{ccc}
\makecell{\bmvaHangBox{
\resizebox{0.5\textwidth}{!}{
\begin{tabular}{@{}ccc|cc@{}}
\toprule
\multicolumn{3}{c|}{Components} & \multicolumn{2}{c}{Accuracy} \\
\midrule
Global Satis.  & Min/Max & Heuristic & Acc@0.25 & Acc@0.5 \\
\midrule
\ding{56} & \ding{56} &\ding{56}          & 39.3 & 28.9 \\
\ding{52} & \ding{56} &\ding{56}          & 44.5 & 33.2 \\
\ding{52} & \ding{52} &\ding{56}          & 45.7 & 34.2 \\
\ding{52} & \ding{56} &\ding{52}          & 48.1 & 36.8 \\
\ding{52} & \ding{52} &\ding{52}          & 48.9 & 37.3 \\
\bottomrule
\end{tabular}
}
}}&
\makecell{\bmvaHangBox{
\resizebox{0.45\textwidth}{!}{
\begin{tabular}{@{}c|cc@{}}
\toprule
Heuristic & Acc@0.25 & Acc@0.5 \\
\midrule
Random                               & 45.9 & 35.6 \\
Maximum Average Distance             & 39.8 & 30.1 \\
Minimum Average Distance$^\dagger$ & 49.6 & 39.8 \\
\bottomrule
\end{tabular}
}
}}\\
(a)&(b)
\end{tabular}
\end{center}
\caption{\textbf{(a)} Ablation on the three main components of CSVG framework. The evaluation is performed on 1000 queries sampled from the ScanRefer validation set. \textbf{(b)} Ablation of different heuristics for solution selection on the entire ScanRefer validation set. $^\dagger$Default setting in CSVG.}
\label{tab:abl_comp}
\end{table}

\noindent\textbf{Heuristics for Solution Selection.} CSPs generated by the LLM are often under-constrained, leading to multiple possible solutions. We now investigate three heuristics for solution selection. Besides the strategy of random selection, we compute the average distance of all pairs of instances in the solutions and take the one with either minimum or maximum score. As shown in  Table~\ref{tab:abl_comp}~(b), the ``Minimum Average Distance'' strategy used in our CSVG performs the best. The ``Random'' heuristic produces lower accuracy, and ``Maximum Average Distance'' delivers the worst results. These differences in grounding accuracy stem from the fact that, in most queries from the ScanRefer dataset, the objects mentioned in a query are typically very close to each other; only in rare cases is this not the case.




\section{Conclusion}
\label{sec:conclusion}

In this paper, we propose CSVG, a novel zero-shot 3D visual grounding framework based on CSP. Our method can be directly applied to diverse scenarios without the need of training, utilizing the CSP formulation to deliver globally valid solutions that account for all spatial constraints simultaneously. This addresses the limitations of existing LLM-driven methods which rely on local reasoning of pairwise relations. The proposed CSVG framework can further handle complex scenes and queries involving negation and counting with minor modifications. We perform extensive experiments and ablation studies on two 3DVG datasets. Results have evidently shown the effectiveness of our proposed framework, with substantial improvements over previous methods.

\noindent\textbf{Future Work.} Following this direction, we can still explore how to integrate appearance information, e.g., color and shape, into the grounding pipeline. Also, instead of using a set of predefined functions, letting the LLM generate relation checking functions on-the-fly can greatly increase the flexibility of the system. Besides constraints, it could be beneficial to also let the LLM choose the heuristic to use for under-constrained CSPs.

\section*{Acknowledgement}


The authors would like to thank Computer Vision for Human-Computer Interaction Lab (cv:hci), led by Prof. Dr.-Ing. Rainer Stiefelhagen, at Karlsruhe Institute of Technology, for providing GPUs and other support to our work.

\bibliography{main}

\clearpage
\setcounter{page}{1}
\setcounter{section}{0}
\renewcommand{\thesection}{\Alph{section}}

\section*{Appendix}

The code of this paper is available \href{https://github.com/sunsleaf/CSVG}{here}.
The website is available here: \href{https://asig-x.github.io/csvg_web}{here}.

\newcommand{\code}[1]{{\ttfamily \footnotesize{#1}}}

\definecolor{codegreen}{rgb}{0,0.6,0}
\definecolor{codegray}{rgb}{0.5,0.5,0.5}
\definecolor{codepurple}{rgb}{0.58,0,0.82}
\definecolor{backcolour}{rgb}{0.95,0.95,0.92}

\lstdefinestyle{mystyle}{
    backgroundcolor=\color{backcolour},   
    commentstyle=\color{codegreen},
    keywordstyle=\color{magenta},
    stringstyle=\color{codepurple},
    basicstyle=\ttfamily\linespread{1.15}\scriptsize, 
    breakatwhitespace=false,         
    breaklines=true,                 
    captionpos=b,                    
    keepspaces=true,                 
    numbers=none,                    
    showspaces=false,                
    showstringspaces=false,
    showtabs=false,                  
    tabsize=2,
    captionpos=b,
}
\lstset{style=mystyle}

\section{Prompts for LLM}
We now introduce the LLM prompts for generating the Python programs for formulating CSPs. Listing \ref{lst:system_msg} shows the system message, and Listing~\ref{lst:in_context_example_1}-\ref{lst:in_context_example_11} provides the eleven in-context examples given to the LLM. Regarding the system message, we replace the two placeholders, i.e., \code{<[REGISTERED\_FUNCTIONS\_PLACEHOLDER]>} and \code{<[REGISTERED\_SCORE\_FUNCTIONS\_PLACEHOLDER]>}, with a list of predefined function signatures (see Sec.~\ref{sec:predef_funcs}) and a list of score functions (see Sec.~\ref{sec:score_funcs}), respectively. Moreover, we parse the \code{<[SYSTEM]>}, \code{<[USER]>} and \code{<[ASSISTANT]>} symbols and associate their corresponding message contents with the roles of ``system'', ``user'', ``assistant'', respectively. Each role will be substituted by a special token pretrained for the LLM. This can regulate the LLM's output to have a structure similar to the assistant responses given in the in-context examples, i.e., the text following \code{<[ASSISTANT]>}.

\begin{lstlisting}[float=h, language=Python, caption={Predefined functions for CSP building.}, label={lst:predef_funcs}]
CONSTRAINT_ABOVE(target: CSPVar, anchor: CSPVar) -> CSPConstraint
CONSTRAINT_BELOW(target: CSPVar, anchor: CSPVar) -> CSPConstraint
CONSTRAINT_ON(target: CSPVar, anchor: CSPVar) -> CSPConstraint
CONSTRAINT_UNDER(target: CSPVar, anchor: CSPVar) -> CSPConstraint
CONSTRAINT_FAR(target: CSPVar, anchor: CSPVar) -> CSPConstraint
CONSTRAINT_AWAY(target: CSPVar, anchor: CSPVar) -> CSPConstraint
CONSTRAINT_ACROSS(target: CSPVar, anchor: CSPVar) -> CSPConstraint
CONSTRAINT_OPPOSITE(target: CSPVar, anchor: CSPVar) -> CSPConstraint
CONSTRAINT_NEAR(target: CSPVar, anchor: CSPVar) -> CSPConstraint
CONSTRAINT_BESIDE(target: CSPVar, anchor: CSPVar) -> CSPConstraint
CONSTRAINT_CLOSE(target: CSPVar, anchor: CSPVar) -> CSPConstraint
CONSTRAINT_LEFT(target: CSPVar, anchor: CSPVar) -> CSPConstraint
CONSTRAINT_RIGHT(target: CSPVar, anchor: CSPVar) -> CSPConstraint
CONSTRAINT_FRONT(target: CSPVar, anchor: CSPVar) -> CSPConstraint
CONSTRAINT_BEHIND(target: CSPVar, anchor: CSPVar) -> CSPConstraint
CONSTRAINT_CENTER(target: CSPVar, anchor: CSPVar) -> CSPConstraint
CONSTRAINT_MIDDLE(target: CSPVar, anchor: CSPVar) -> CSPConstraint
CONSTRAINT_IN(target: CSPVar, anchor: CSPVar) -> CSPConstraint
CONSTRAINT_INSIDE(target: CSPVar, anchor: CSPVar) -> CSPConstraint
CONSTRAINT_BETWEEN(target: CSPVar, anchors: set[CSPVar]) -> CSPConstraint
CONSTRAINT_LESS(target: CSPVar, reference: CSPVar, score_func: str, anchor: CSPVar | None = None) -> CSPConstraint
CONSTRAINT_MORE(target: CSPVar, reference: CSPVar, score_func: str, anchor: CSPVar | None = None) -> CSPConstraint
CONSTRAINT_MAX_OF(target: CSPVar, score_func: str, anchor: CSPVar | None = None) -> CSPConstraint
CONSTRAINT_MIN_OF(target: CSPVar, score_func: str, anchor: CSPVar | None = None) -> CSPConstraint
DEFINE_NEGATIVE_VARIABLE(labels: list[str]) -> CSPVar
DEFINE_VARIABLE(labels: list[str]) -> CSPVar
SET_TARGET(obj: CSPVar) -> None
\end{lstlisting}

\section{Building CSPs}
\subsection{Predefined Functions}
\label{sec:predef_funcs}

We show in Listing~\ref{lst:predef_funcs} the signatures of predefined functions available to the LLM, which are used to build CSPs based on the queries. Here, the class \code{CSPVar} stores information (e.g.,, labels) of a CSP variable. The descriptions of these functions are as follows.
\begin{itemize}
\item \code{\textbf{CONSTRAINT\_ABOVE}}
creates a constraint requiring that the instance assigned to the \code{target} variable should be above the instance assigned to the \code{anchor} variable.
It checks two conditions:
\begin{itemize}
\item The z-coordinate of the center of the target instance is larger than that of the anchor instance.
\item The horizontal distance between the centers of the target and anchor instances should be within a threshold (\code{ABOVE\_BELOW\_HORIZONTAL\_DISTANCE}).    
\end{itemize}

\item \code{\textbf{CONSTRAINT\_BELOW}}
creates a constraint that requires the instance assigned to the \code{target} variable to be below the \code{anchor} instance.
It checks two conditions:
\begin{itemize}
\item The z-coordinate of the center of the target instance is smaller than that of the anchor instance.
\item The horizontal distance between the centers of the target and anchor instances should be within a threshold (\code{ABOVE\_BELOW\_HORIZONTAL\_DISTANCE}). 
\end{itemize}

\item \code{\textbf{CONSTRAINT\_ON}}
applies the same strategy as in \code{CONSTRAINT\_ABOVE}. One may also require the target and anchor instances to be closer to each other than the threshold in \code{CONSTRAINT\_ABOVE}.

\item \code{\textbf{CONSTRAINT\_UNDER}}
applies the same strategy as in \code{CONSTRAINT\_BELOW}.

\item \code{\textbf{CONSTRAINT\_FAR}}
creates a constraint requiring that the instance assigned to the \code{target} variable should be far from the instance assigned to the \code{anchor} variable.
It checks the following condition:
\begin{itemize}
\item The distance between the centers of the target and anchor instances should be above a threshold (\code{FAR\_DISTANCE}).
\end{itemize}

\item \code{\textbf{CONSTRAINT\_AWAY}}, \code{\textbf{CONSTRAINT\_OPPOSITE}} and \code{\textbf{CONSTRAINT\_ACROSS}} apply the same strategy as in \code{CONSTRAINT\_FAR}.

\item \code{\textbf{CONSTRAINT\_NEAR}}
creates a constraint requiring that the instance assigned to the \code{target} variable should be near the \code{anchor} instance. It checks the following condition:
\begin{itemize}
\item The distance between the centers of the target and anchor instances should be within a threshold (\code{NEAR\_DISTANCE}).
\end{itemize}

\item \code{\textbf{CONSTRAINT\_BESIDE}} and \code{\textbf{CONSTRAINT\_CLOSE}}
apply the same strategy as \code{CONSTRAINT\_NEAR}.

\item \code{\textbf{CONSTRAINT\_LEFT}}
creates a constraint requiring that the instance assigned to the \code{target} variable should be to the left of the \code{anchor} instance assuming that the viewpoint is at the center of the scene. It converts the center coordinates of the target instance into a viewer's frame at the anchor instance towards the center of the scene, with x-axis going to the right, y-axis upward, and z-axis backward. This function then checks if the x-coordinate of the target instance's center is positive in the viewer's frame.

\item \code{\textbf{CONSTRAINT\_RIGHT}}
applies a strategy similar to \code{CONSTRAINT\_LEFT}, except that it requires the x-coordinate of the target instance's center to be negative in the viewer's frame.

\item \code{\textbf{CONSTRAINT\_FRONT}} and \code{\textbf{CONSTRAINT\_BEHIND}}
apply the same strategy as in \code{CONSTRAINT\_NEAR}. We have noticed that this type of spatial relation is highly view-dependent and cannot be easily determined systematically.

\item \code{\textbf{CONSTRAINT\_CENTER}}
creates a constraint requiring that the instance assigned to the \code{target} variable should be close to the center of the instance assigned to the \code{anchor} variable.
It checks whether the center of the target instance is inside the bounding box of the anchor instance.

\item \code{\textbf{CONSTRAINT\_MIDDLE}}, \code{\textbf{CONSTRAINT\_IN}} and \code{\textbf{CONSTRAINT\_INSIDE}} apply the same strategy as \code{CONSTRAINT\_CENTER}.

\item \code{\textbf{CONSTRAINT\_BETWEEN}}
creates a constraint requiring the instance assigned to the \code{target} variable to be between the instances assigned to the \code{anchor} variables. Concretely, it checks whether the distance between the center of the target instance and the center of all anchor instances is within a threshold (\code{BETWEEN\_DISTANCE}).

\item \code{\textbf{CONSTRAINT\_LESS}}
requires that the instance assigned to the \code{target} variable must have a lower score than the instance assigned to the \code{reference} variable, where the scores are computed by a score function \code{score\_func}. Some score functions require an \code{anchor} for computing the score, e.g.,, \code{``distance''}.

\item \code{\textbf{CONSTRAINT\_MORE}}
applies a strategy similar to \code{CONSTRAINT\_LESS}, except that the instance assigned to the \code{target} variable should have a higher score than that of the \code{reference} variable.

\item \code{\textbf{CONSTRAINT\_MAX\_OF}}
creates a constraint requiring that the instance assigned to the \code{target} variable should have the maximum score among all instances with the same label. The semantics of the \code{score\_func} and \code{anchor} parameters are similar to those in \code{CONSTRAINT\_LESS} and \code{CONSTRAINT\_MORE}.

\item \code{\textbf{CONSTRAINT\_MIN\_OF}}
is similar to \code{CONSTRAINT\_MAX\_OF}, except that we require the instance assigned to the \code{target} variable to have the minimum score.

\item \code{\textbf{DEFINE\_VARIABLE}}
defines a CSP variable with a set of labels (which should refer to similar things, e.g.,, ``dishwasher'' and ``dish washing machine'').

\item \code{\textbf{DEFINE\_NEGATIVE\_VARIABLE}}
defines a CSP variable marked as negative, indicating that no instance in the domain of this variable should satisfy any constraints involving this variable.

\item \code{\textbf{SET\_TARGET}}
marks the given CSP variable as the target; the grounding results are the instances assigned to this variable.
\end{itemize}

Note that some threshold values (e.g., \code{NEAR\_DISTANCE}) are used in these predefined functions. These parameters are adjusted in accordance to the given datasets or scenarios. For ScanReferi.e. and Nr3Di.e. datasets used in our evaluation, we have noticed that setting these thresholds to large values produces the best results. In this case, our system tends to focus more on determining the type of a spatial relation and less on the distances between objects. For other datasets or experiments in real-world scenarios, these parameters may need further adjustments.

\subsection{Score Functions}
\label{sec:score_funcs}

As mentioned earlier, several predefined functions require the \code{score\_func} parameter to compare instances. We now provide a detailed explanation of associated score functions.
\\
\code{\textbf{``distance''}}
computes the distance between the target instance and the anchor instance as the score.
\\
\code{\textbf{``size-x''}}
computes the size of the bounding box of an instance along x-axis as the score.
\\
\code{\textbf{``size-y''}}
computes the size of the bounding box of an instance along y-axis as the score.
\\
\code{\textbf{``size-z''}}
computes the size of the bounding box of an instance along z-axis as the score.
\\
\code{\textbf{``size''}}
computes the size of the bounding box of an instance along x-, y-, and z-axis and takes the maximum of them as the score.
\\
\code{\textbf{``position-z''}}
computes the z-coordinate of the center of an instance.
\\
\code{\textbf{``left''}}
converts the center coordinates of the target instance in the viewer's frame in the same fashion as we describe \code{CONSTRAINT\_LEFT}, and a larger x-value leads to a higher score, indicating that the target instance is more ``to the left''.
\\
\code{\textbf{``right''}}
converts the center coordinates of the target instance in the viewer's frame in the same fashion as we describe \code{CONSTRAINT\_LEFT}, and a lower x-value leads to a higher score, indicating that the target instance is more ``to the right''.
\\
\code{\textbf{``front''}}
converts the center coordinates of the target instance in the viewer's frame in the same fashion as we describe \code{CONSTRAINT\_LEFT}, and a larger z-value leads to a higher score, indicating that the target instance is more ``to the front''. Here, the direction away from the center of the scene is considered as the ``front''.
\\
\code{\textbf{``distance-to-center''}}
computes the distance from the center of an instance to the center of the scene.
\\
\code{\textbf{``distance-to-middle''}}
is an alias for the score function \code{``distance-to-center''}.

\section{CSP Solver}
\noindent\textbf{Solving CSP with Backtracking.} 
We now describe the algorithm for solving CSPs generated by the LLM as shown in Alg.~\ref{alg:csp_solver}. It handles spatial constraints, min/max constraints and negative variables. The algorithm takes the CSP built from the Python program generated by LLM as input and predicts a bounding box to ground the target. We first initialize some auxiliary variables (line 1-4) and filter solutions satisfying all spatial constraints (line 5-26). Concretely, for each combination of assignments to normal variables (line 5), we iterate through all constraints (line 6) and check their satisfiability. We first check if the constraint has a negative variable (line 7). If this is the case, we traverse over all instances in the domain of negative variable and skip the solution if any assignment of the negative variable satisfies the constraint (line 14-20). If the constraint does not involve any negative variables, we simply check its satisfiability with the current solution (line 9-12). Finally, solutions that satisfy all constraints are recorded for further processing (line 23-25). Following the evaluation of spatial constraints, we proceed to verify the min/max constraints (line 28-38). For each min/max constraint, we first group all recorded solutions w.r.t.\ the assignment to the anchor variables (line 31-32). Within each group, we keep the solution that produces the largest/smallest score function value with the instance assigned to the target variable (line 33-36). In this fashion, we obtain a smaller set of solutions, to which the heuristic described in the main paper is applied (line 39) to determine the final solution. Given this solution, a bounding box is produced for grounding the target (line 40). 

\noindent\textbf{Grounding through Local Constraints.} 
Besides the CSP solver described previously, we have also implemented a separate solver that only considers a single constraint at a time to facilitate our ablation study (``Global Satis.''). As shown in Alg.~\ref{alg:non_csp_solver}, we initialize the candidate instances of each CSP variable with all the instances in its domain (line 2-4). We then search for an unprocessed constraint with anchor variables not used as target variables in other unprocessed constraints (line 7-14). If this constraint exists, we filter the candidate instances of its target variable with the instances available to the anchor variables (line 15-16). The procedure is repeated until all constraints are processed. If some constraints persist but none are eligible for processing, the CSP remains unsolved. The target instance is the first one in the set of candidate instances associated with the target variable (line 22-23). Here, we treat the CSP as a graph with directed edges pointing from anchors to targets. Only variables with no incoming edges can be filtered. Once processed, the variable and the associated edges (constraints) are removed, enabling further processing of other variables.

\section{Qualitative Results}
We hereby demonstrate the proposed CSVG with more qualitative results using the ScanRefer dataset. In Fig.~\ref{fig:more_example_csvg_vs_zsvg_1} and~\ref{fig:more_example_csvg_vs_zsvg_2}, we depict the grounding results from CSVG in comparison with ZSVG3D. Our method delivers superior performance with both target and anchor objects located. In Fig.~\ref{fig:more_example_csvg_no_minmax}, we provide further results from CSVG of basic configurations (only spatial constraints) together with the LLM-generated Python programs. The programs clearly demonstrate logical structures that effectively capture the essence of the input query. Our system successfully identifies the correct target among the distractors (objects of the same label as the target) given the spatial relation from the query. In Fig.~\ref{fig:more_example_csvg_with_minmax}, we demonstrate CSVG with min/max constraints. Together with the score functions, the generated Python programs can express relatively complex logical structures using these constraints. In addition, we also demonstrate the effectiveness of the extensions mentioned in the paper. Since ScanRefer dataset does not contain negative or very complex counting-based queries, we take the 3D scenes and use customized queries. The results are shown in Fig.~\ref{fig:more_example_csvg_with_neg} and~\ref{fig:more_example_csvg_with_counting}, where we extend CSVG to handle negation- and counting-based queries. Particularly, in the last example, our proposed CSVG shows the capability of handling a very complex query, where we mix spatial constraints, min/max constraints and counting-based description.

\newcommand{\queryrow}[1]{\multicolumn{4}{c}{\parbox{0.95\linewidth}{\centering \footnotesize{\textbf{{#1} \vspace{10pt}}}}}}
\begin{figure*}
\setlength\tabcolsep{2pt}
\renewcommand{\arraystretch}{1.2}
\centering
\begin{tabular}{cccc}
 \textbf{3DVG Ground Truth} &
 \textbf{Ours (GT Seg.)} &
 \textbf{Ours (Mask3D)} &
 \textbf{ZSVG3D} \\
 \includegraphics[width=0.24\linewidth]{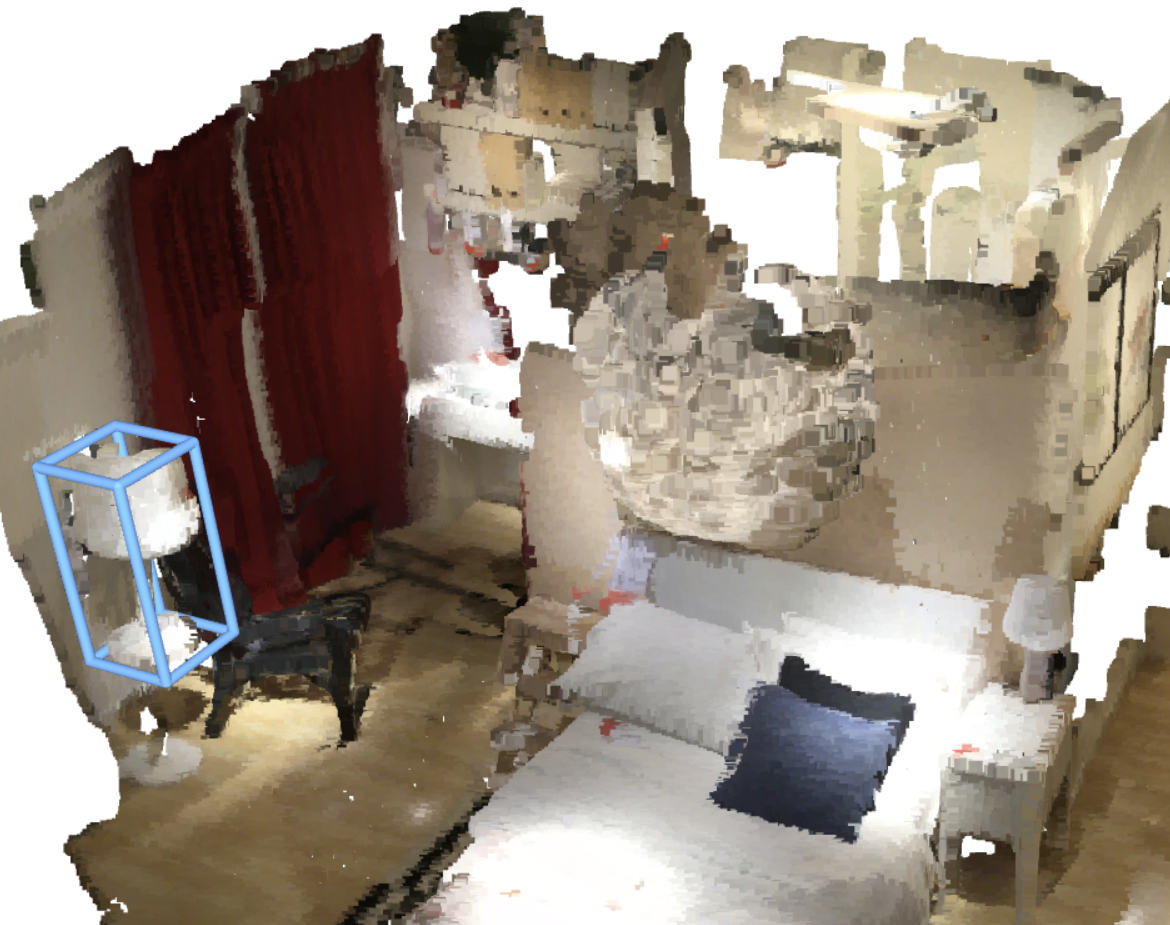} &
 \includegraphics[width=0.24\linewidth]{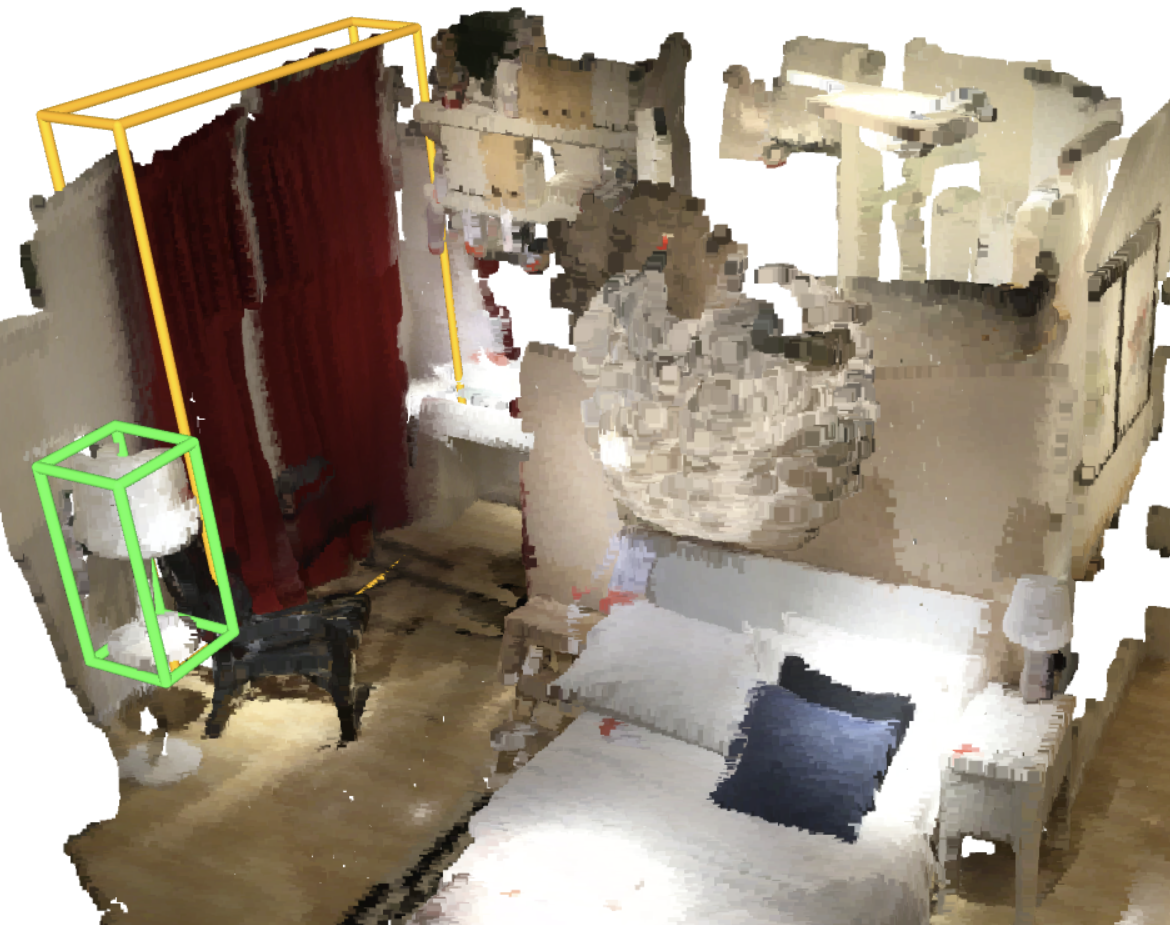} &
 \includegraphics[width=0.24\linewidth]{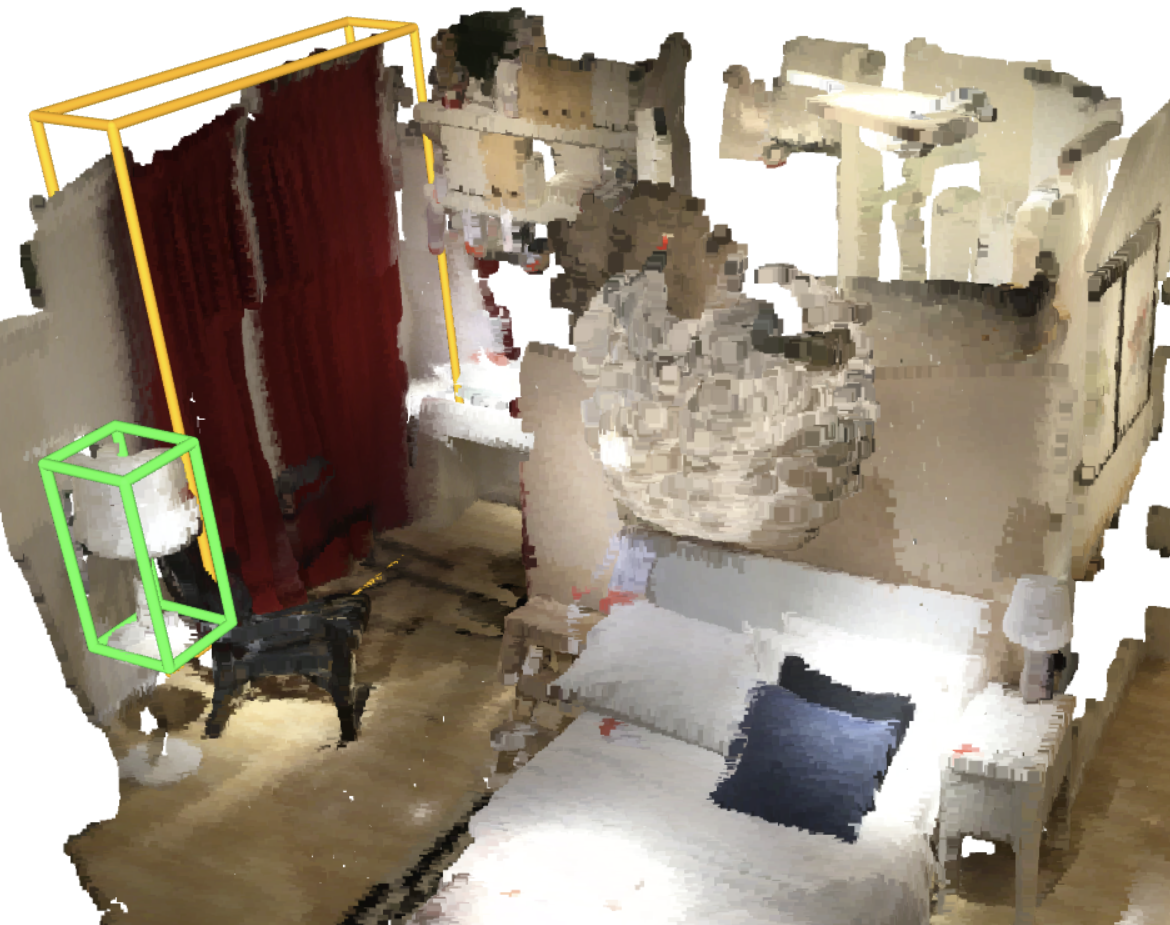} &
 \includegraphics[width=0.24\linewidth]{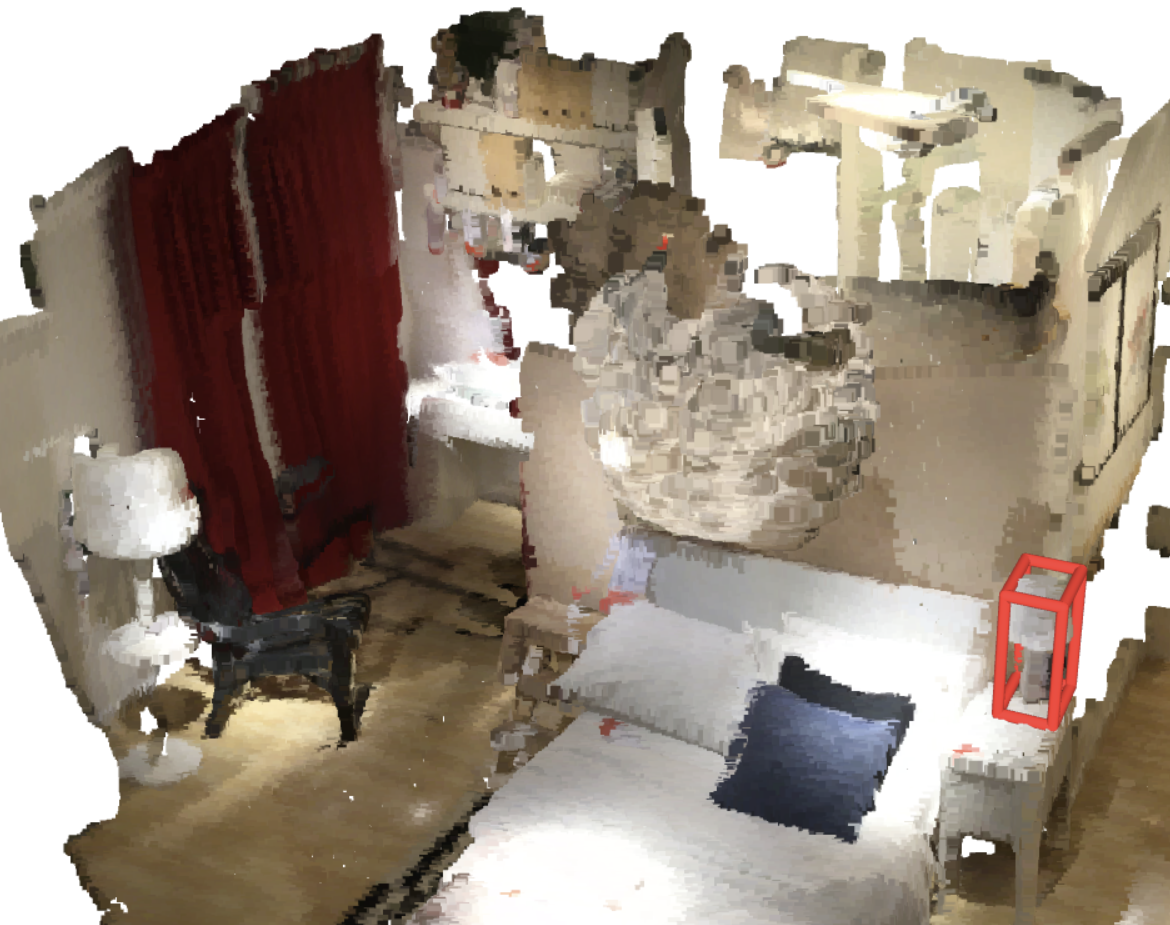} \\
 \queryrow{this is a lamp. its white in color to the left of burgundy drapes.} \\
 \includegraphics[width=0.24\linewidth]{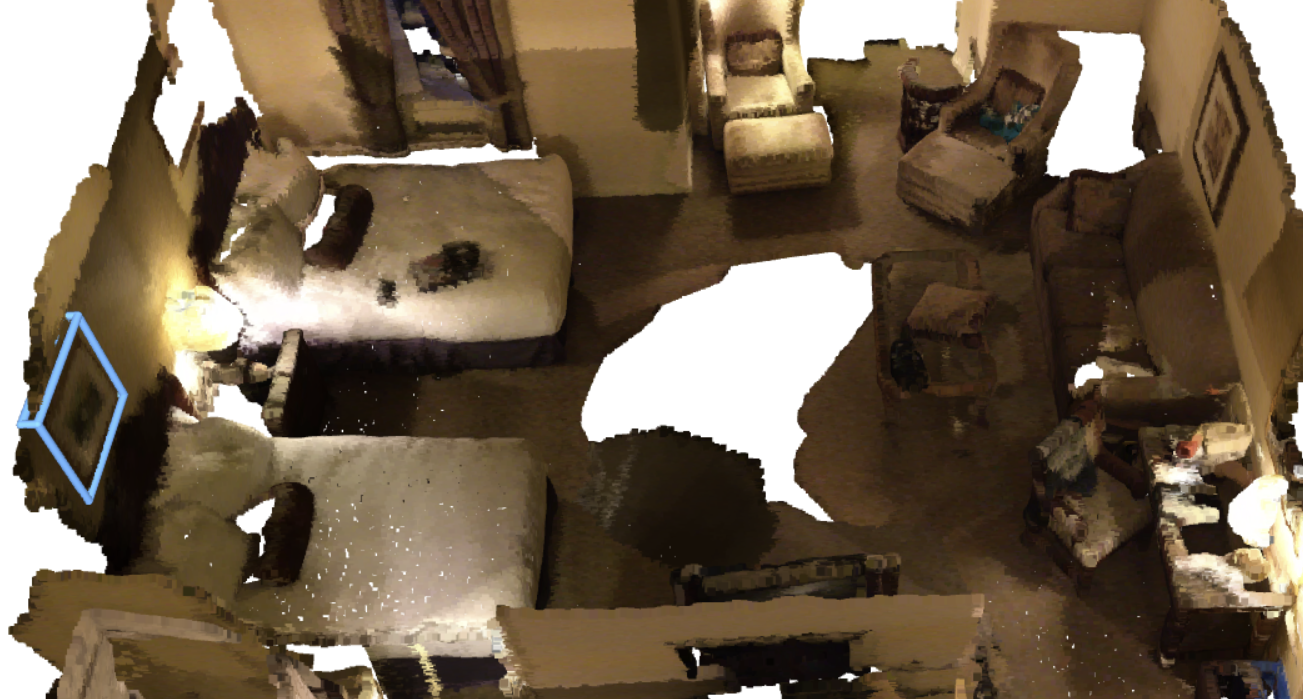} &
 \includegraphics[width=0.24\linewidth]{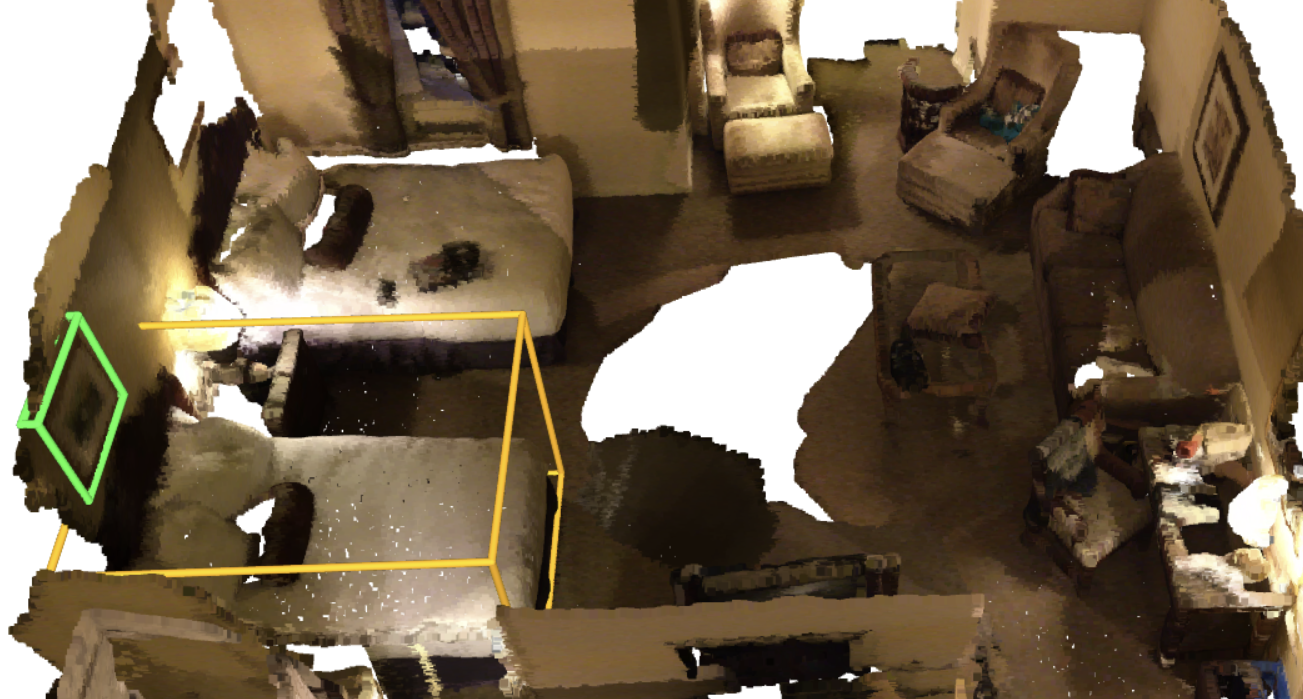} &
 \includegraphics[width=0.24\linewidth]{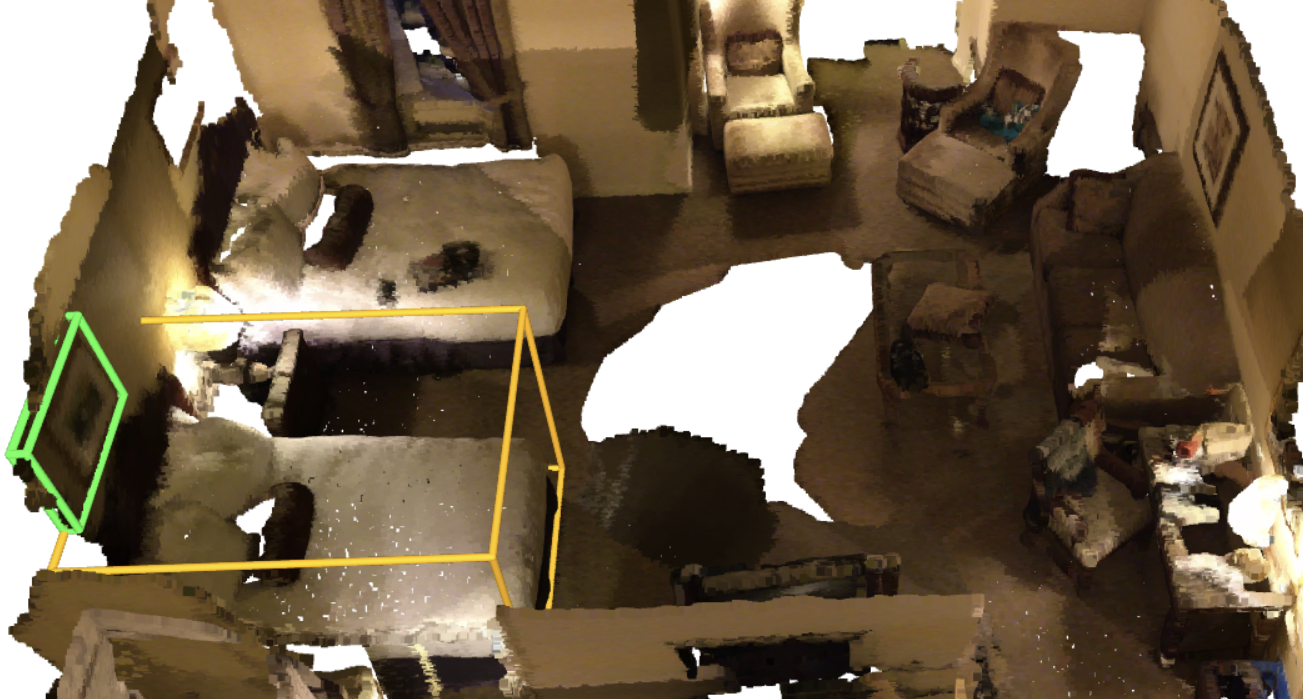} &
 \includegraphics[width=0.24\linewidth]{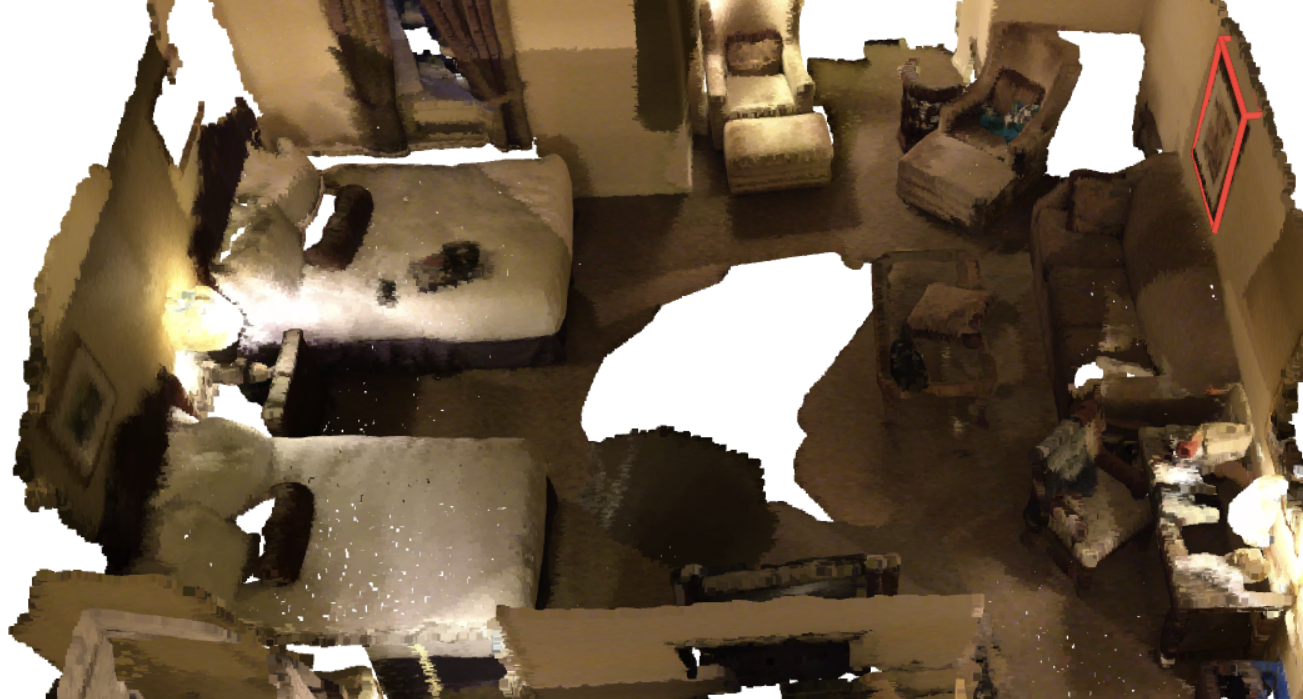} \\
 \queryrow{this is a square picture. it is above the bed.} \\
 \includegraphics[width=0.24\linewidth]{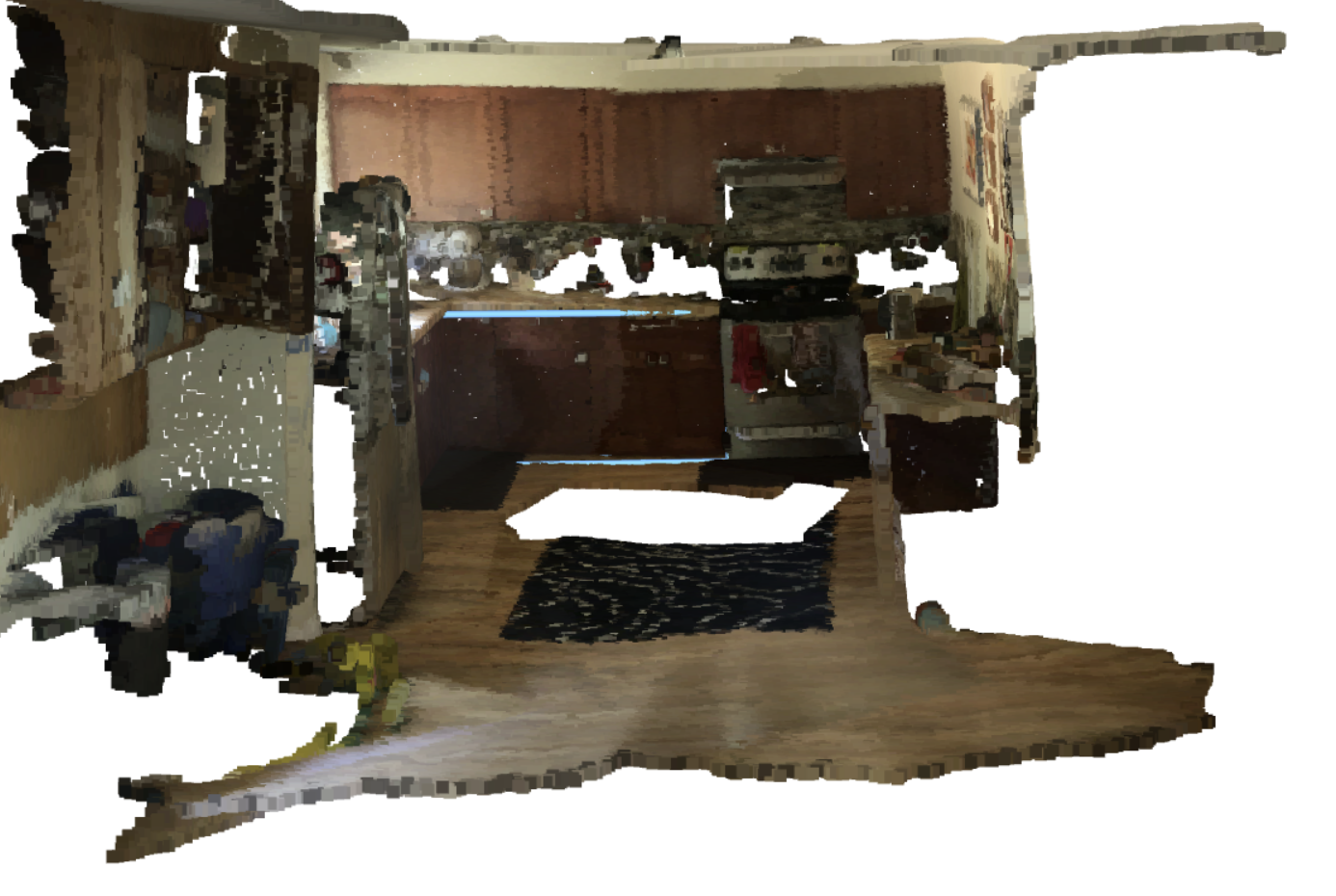} &
 \includegraphics[width=0.24\linewidth]{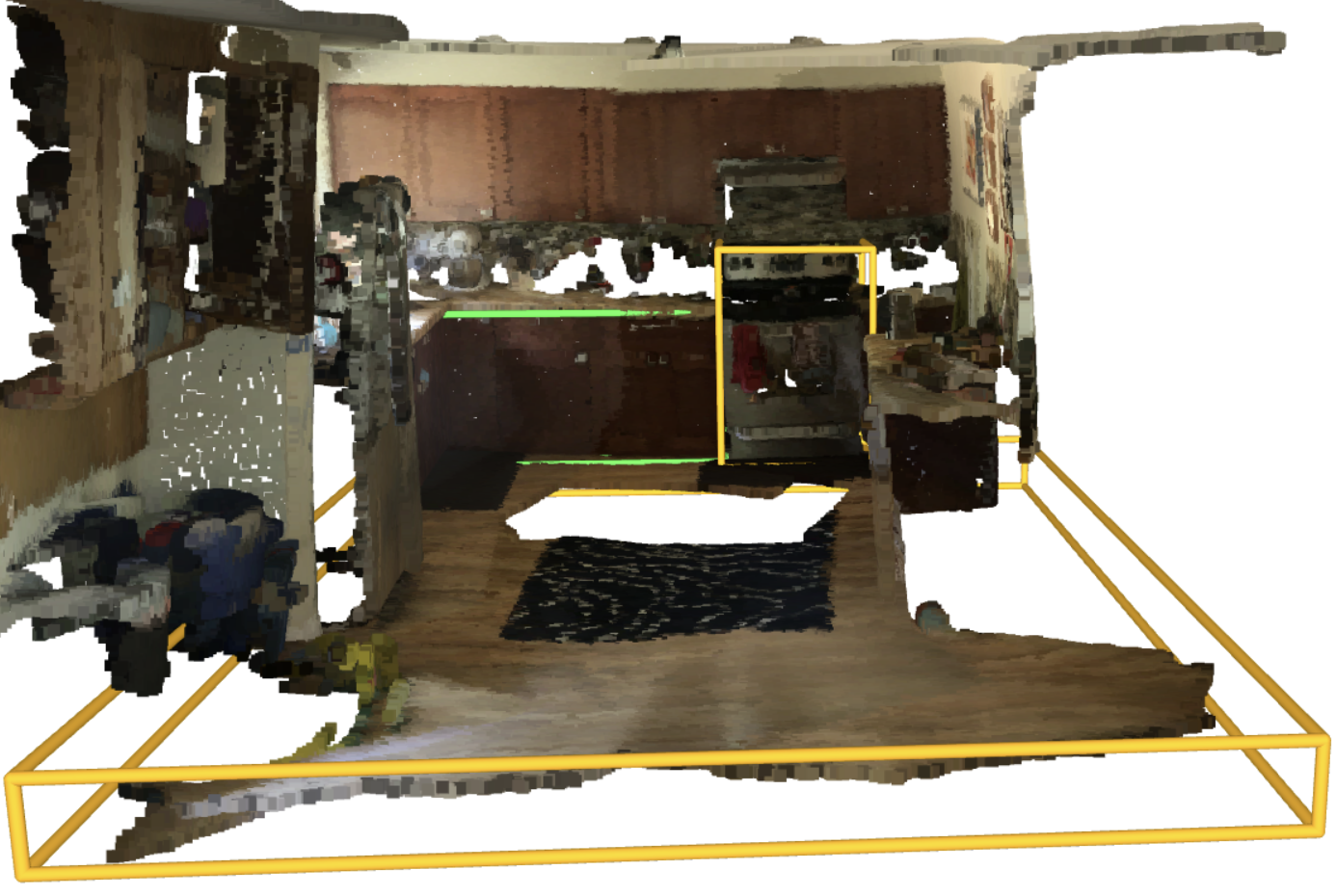} &
 \includegraphics[width=0.24\linewidth]{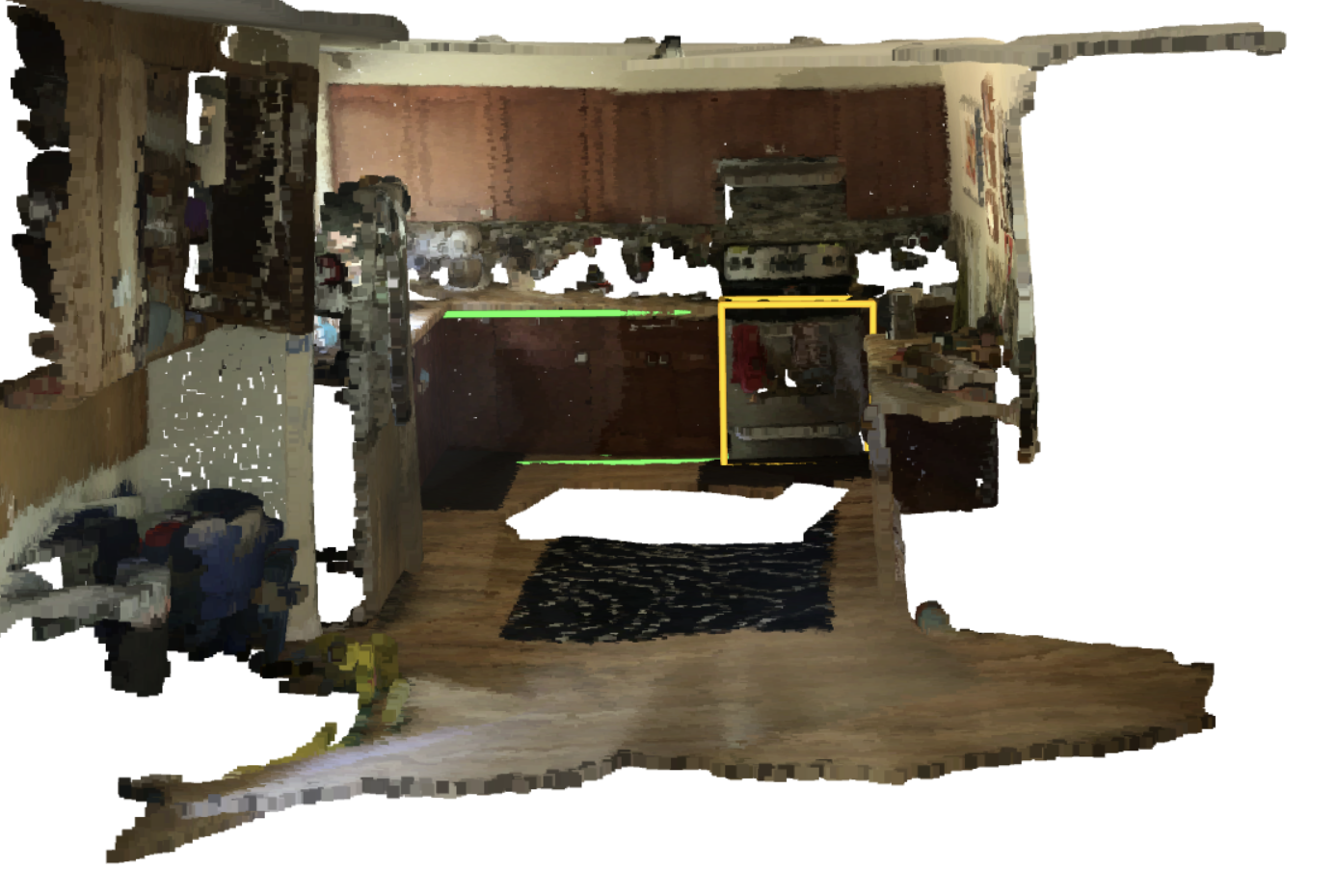} &
 \includegraphics[width=0.24\linewidth]{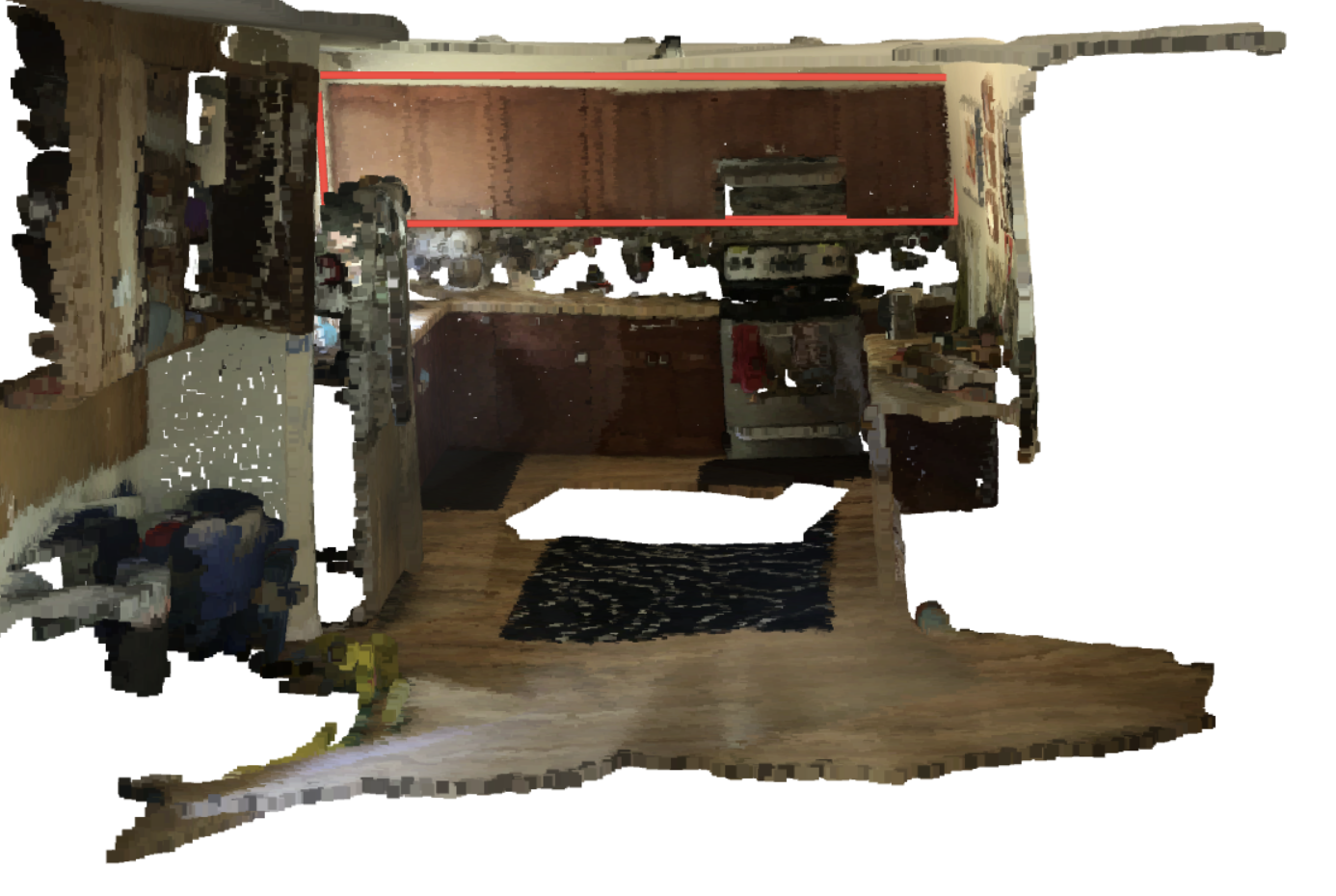} \\
 \queryrow{there is a set of brown kitchen cabinets on the floor. they are to the left of the oven.} \\
 \includegraphics[width=0.24\linewidth]{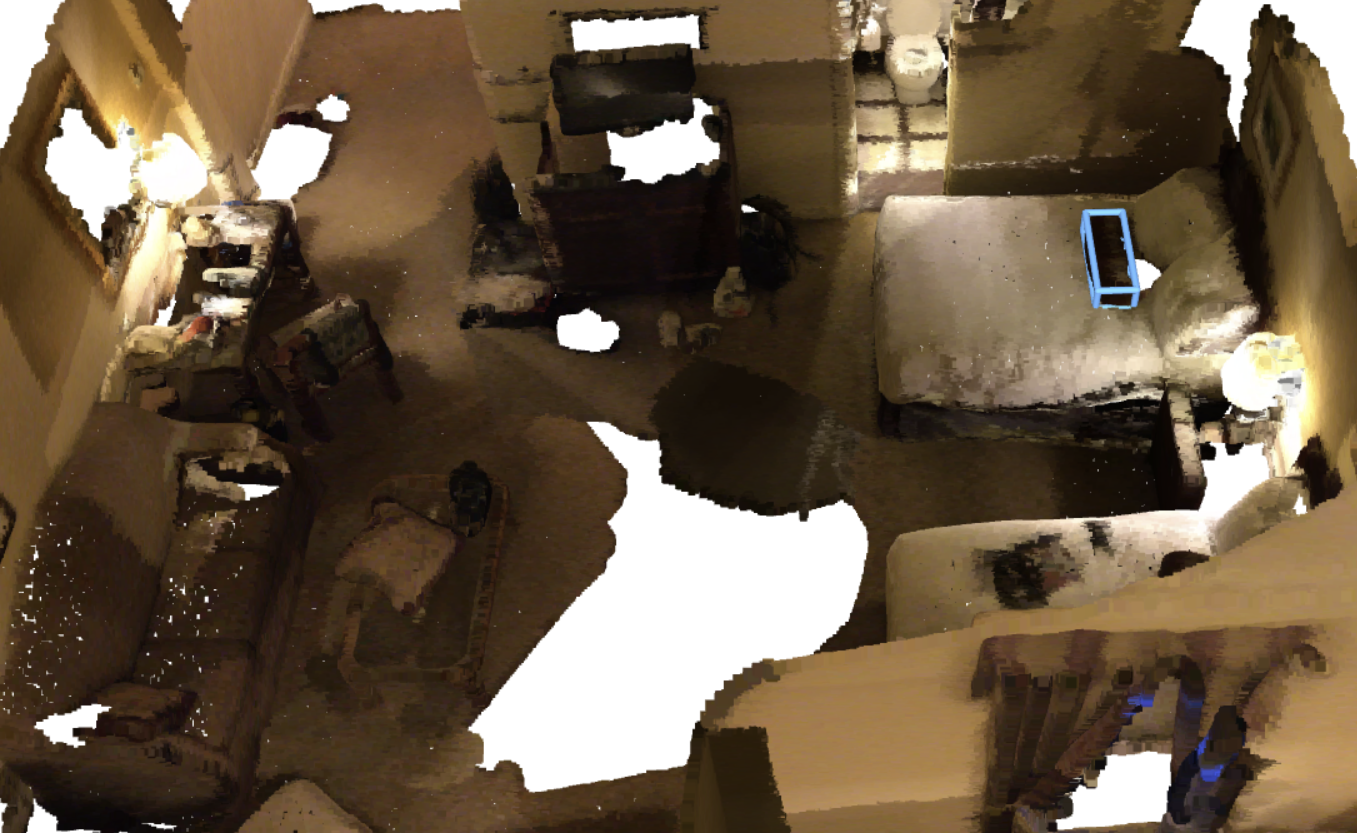} &
 \includegraphics[width=0.24\linewidth]{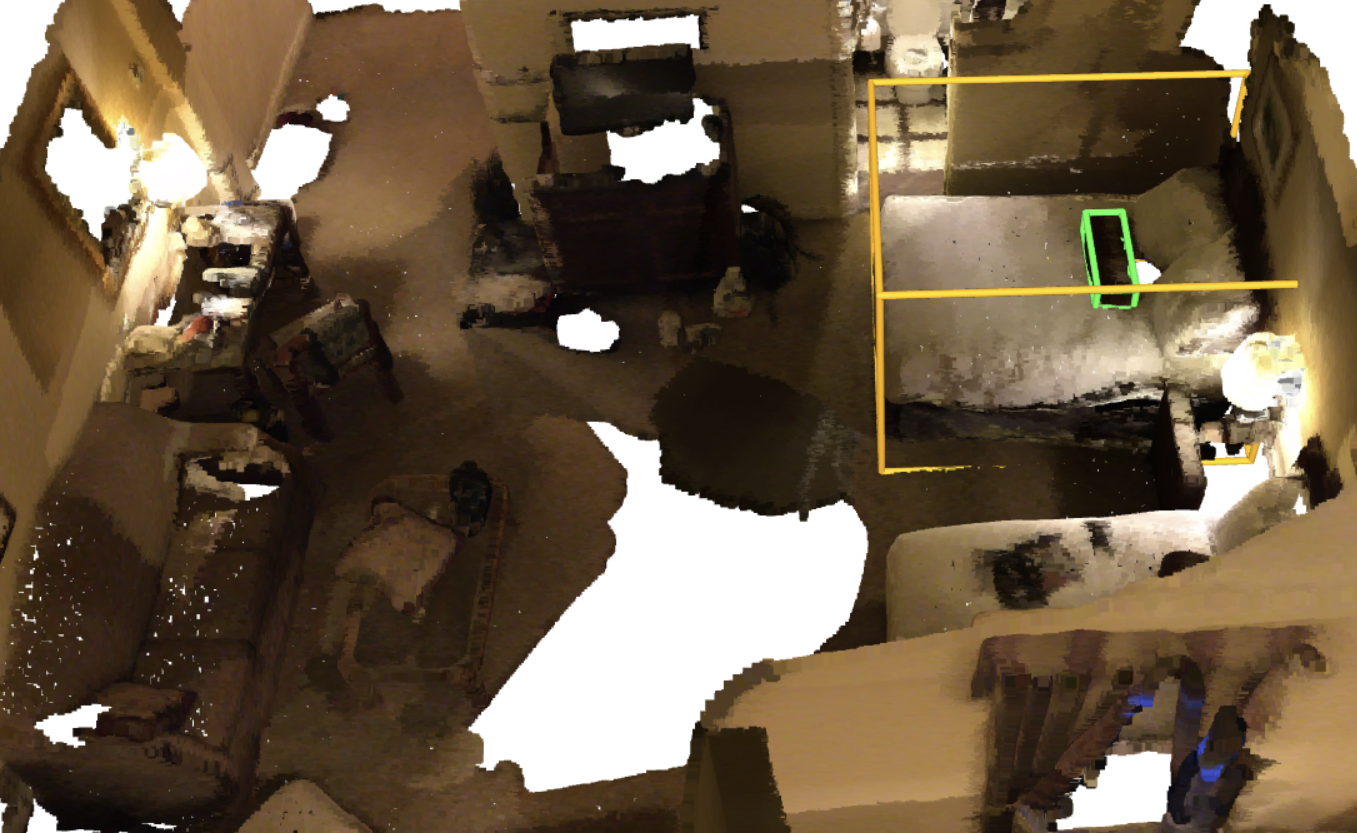} &
 \includegraphics[width=0.24\linewidth]{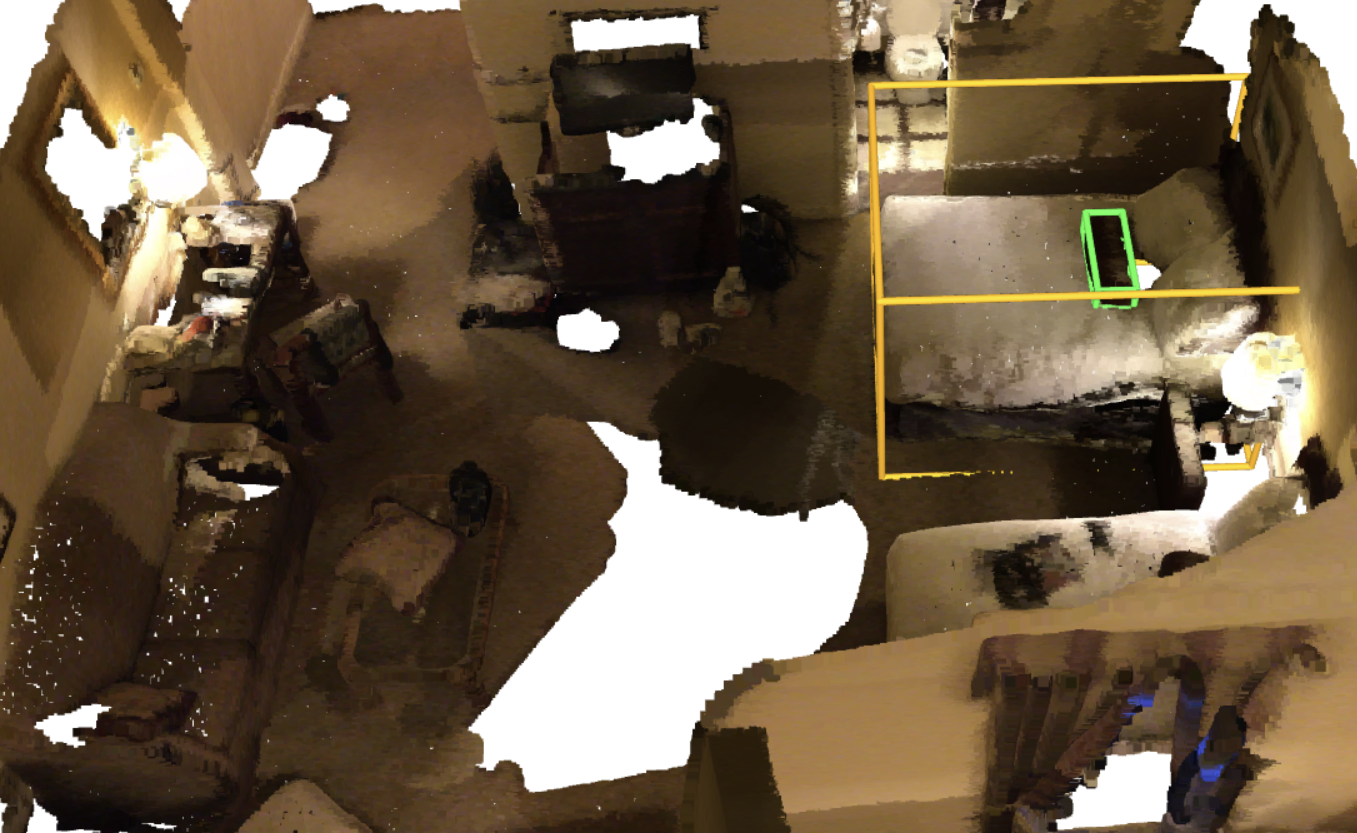} &
 \includegraphics[width=0.24\linewidth]{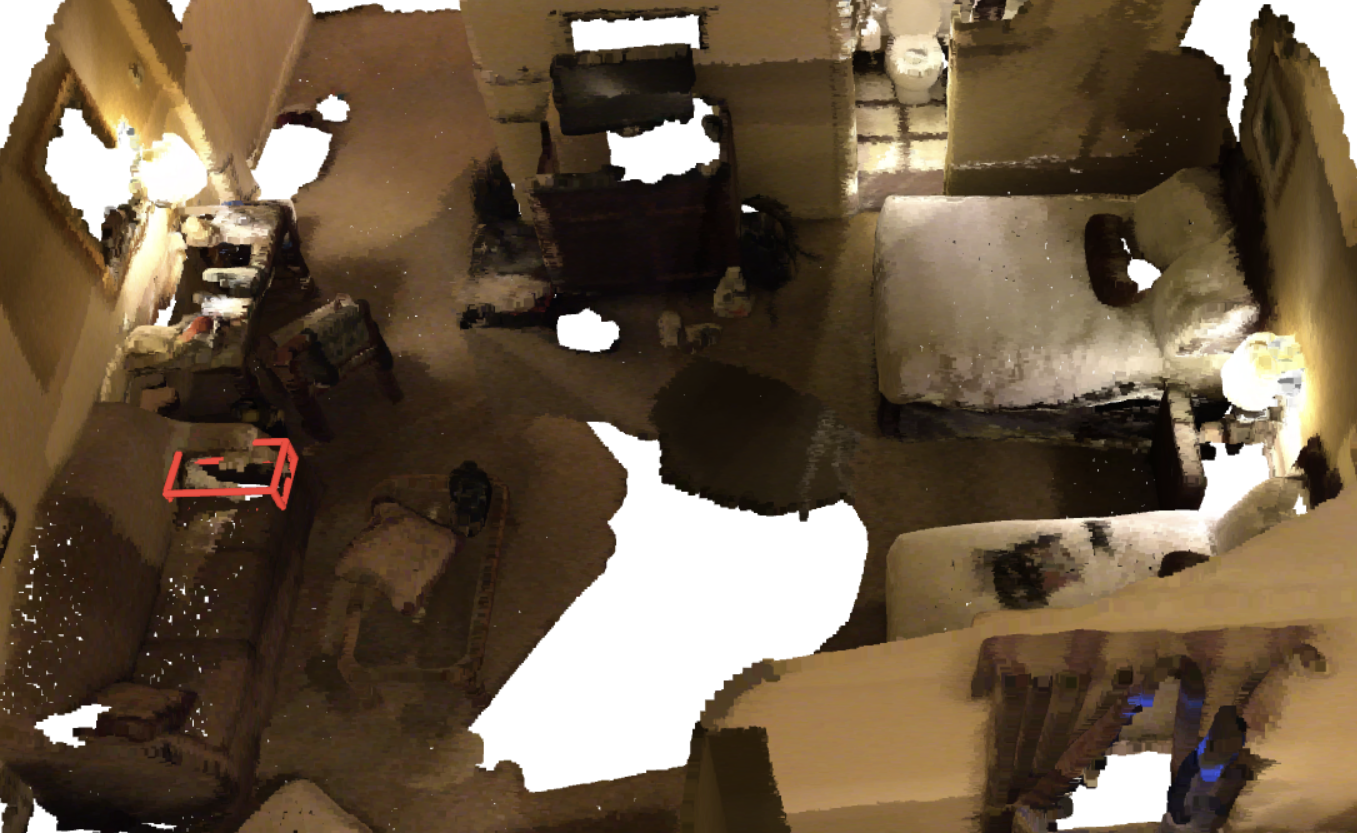} \\
 \queryrow{this pillow is on the bed. it is brown.} \\
 \includegraphics[width=0.24\linewidth]{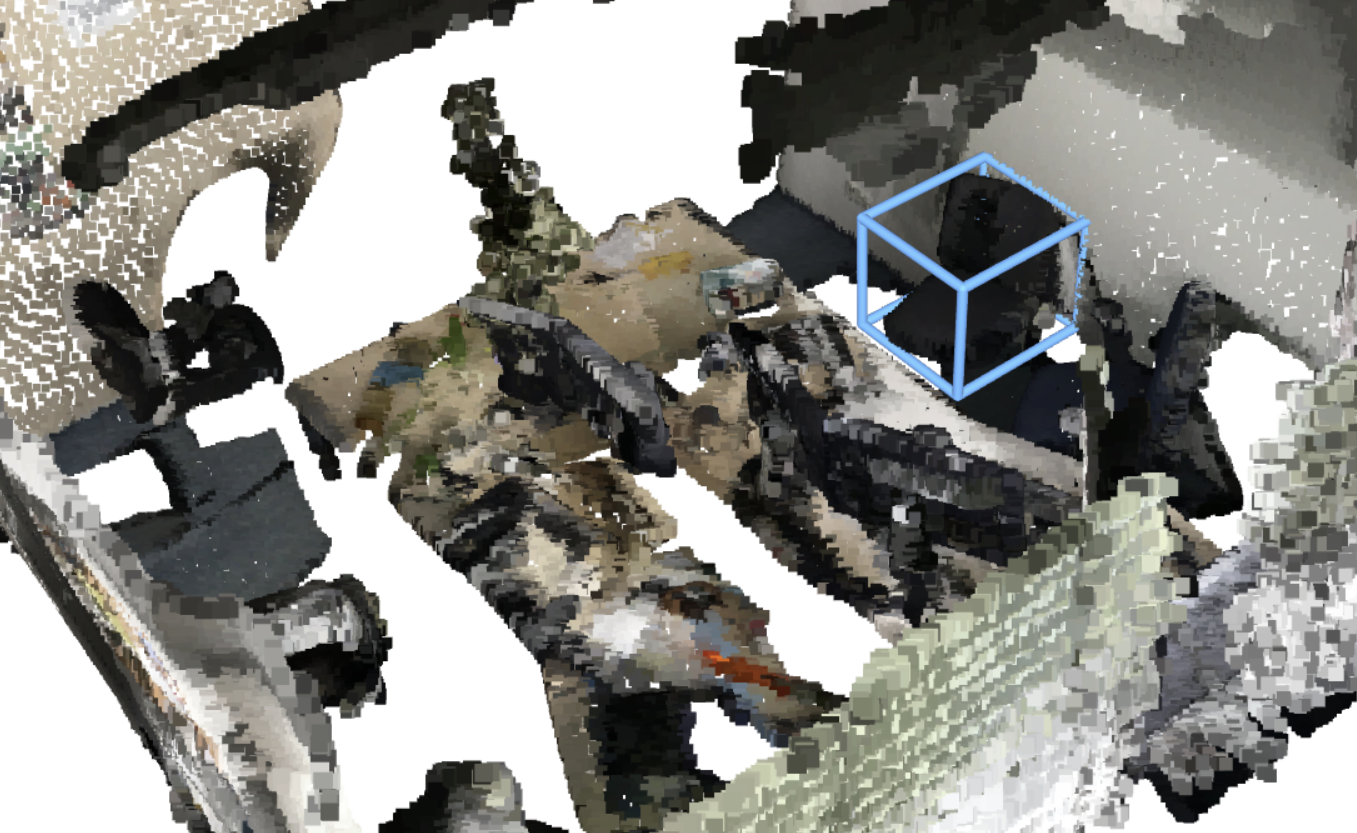} &
 \includegraphics[width=0.24\linewidth]{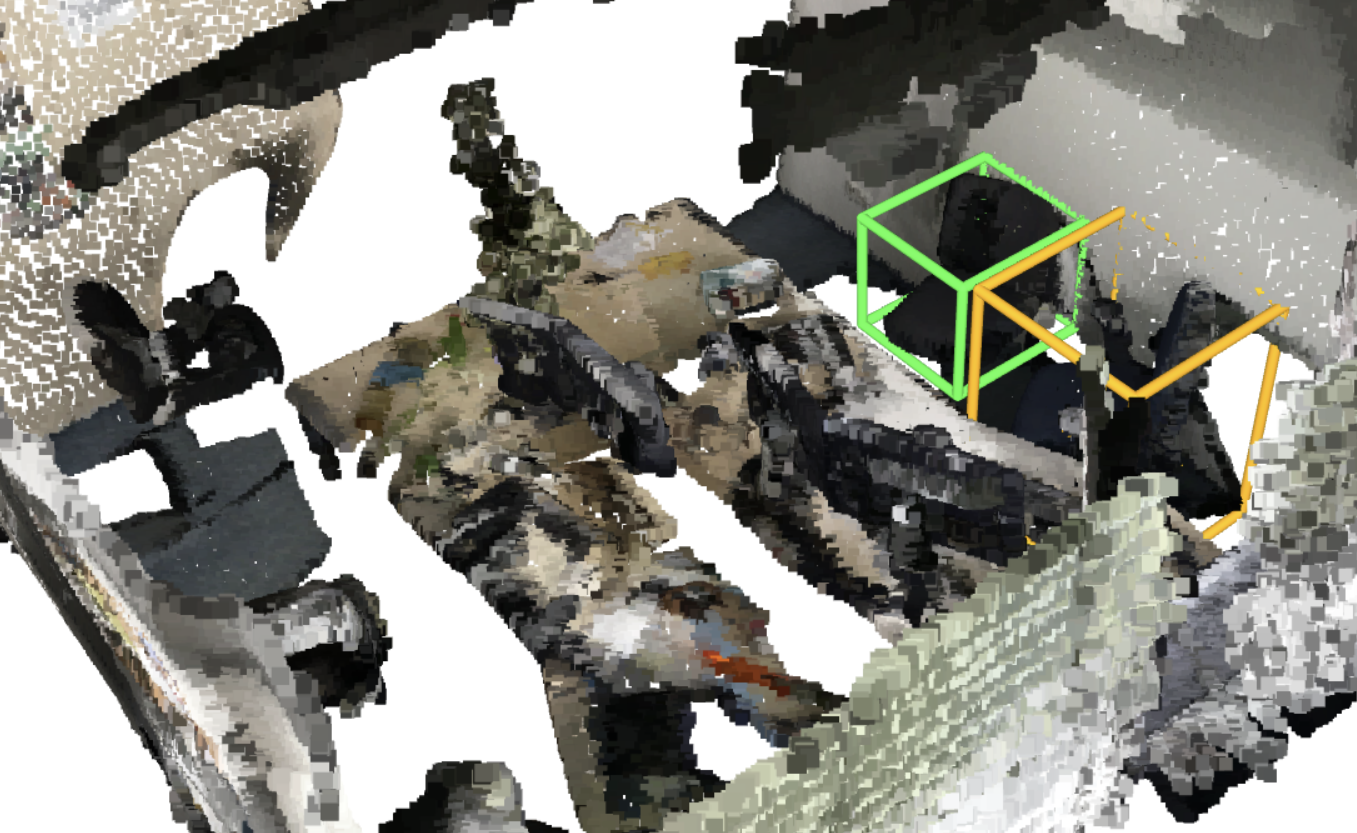} &
 \includegraphics[width=0.24\linewidth]{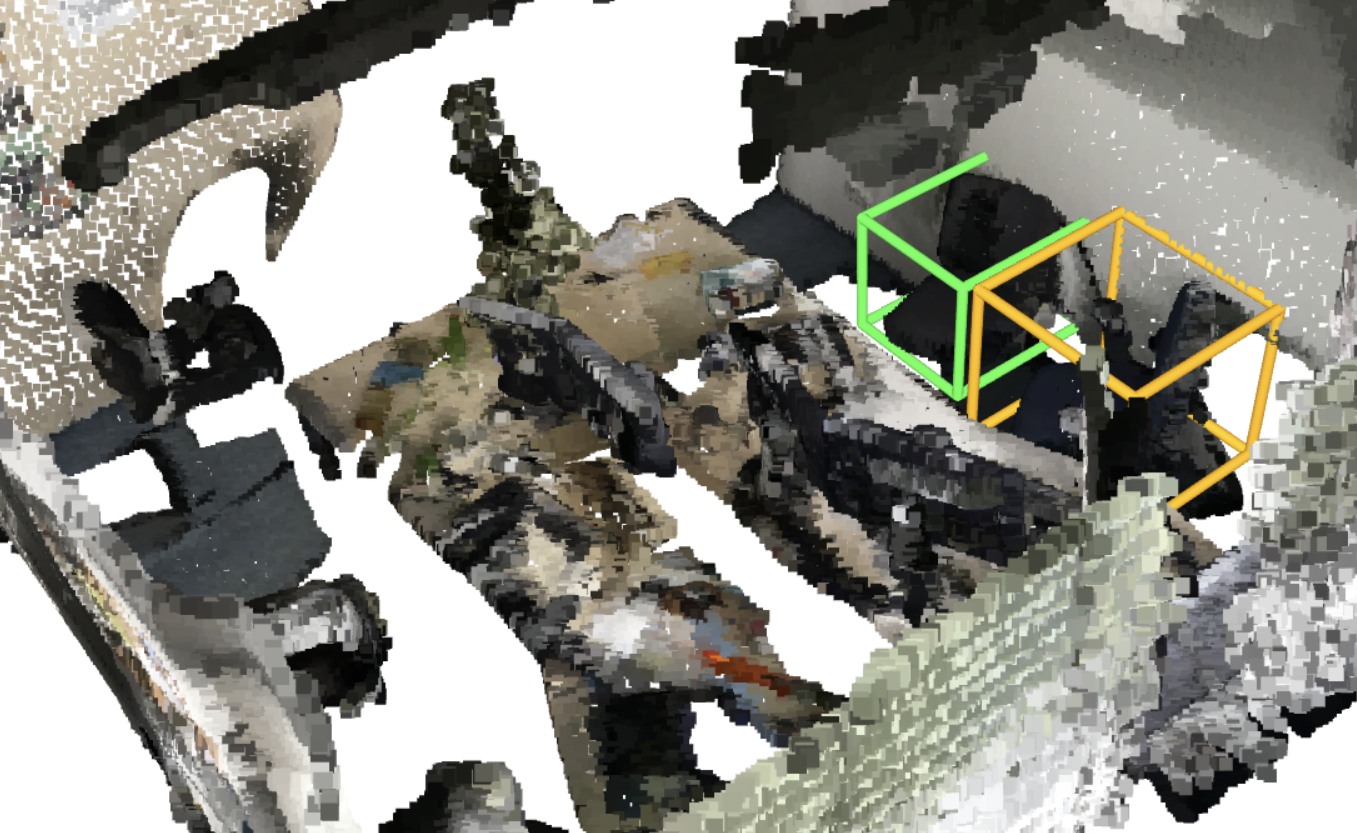} &
 \includegraphics[width=0.24\linewidth]{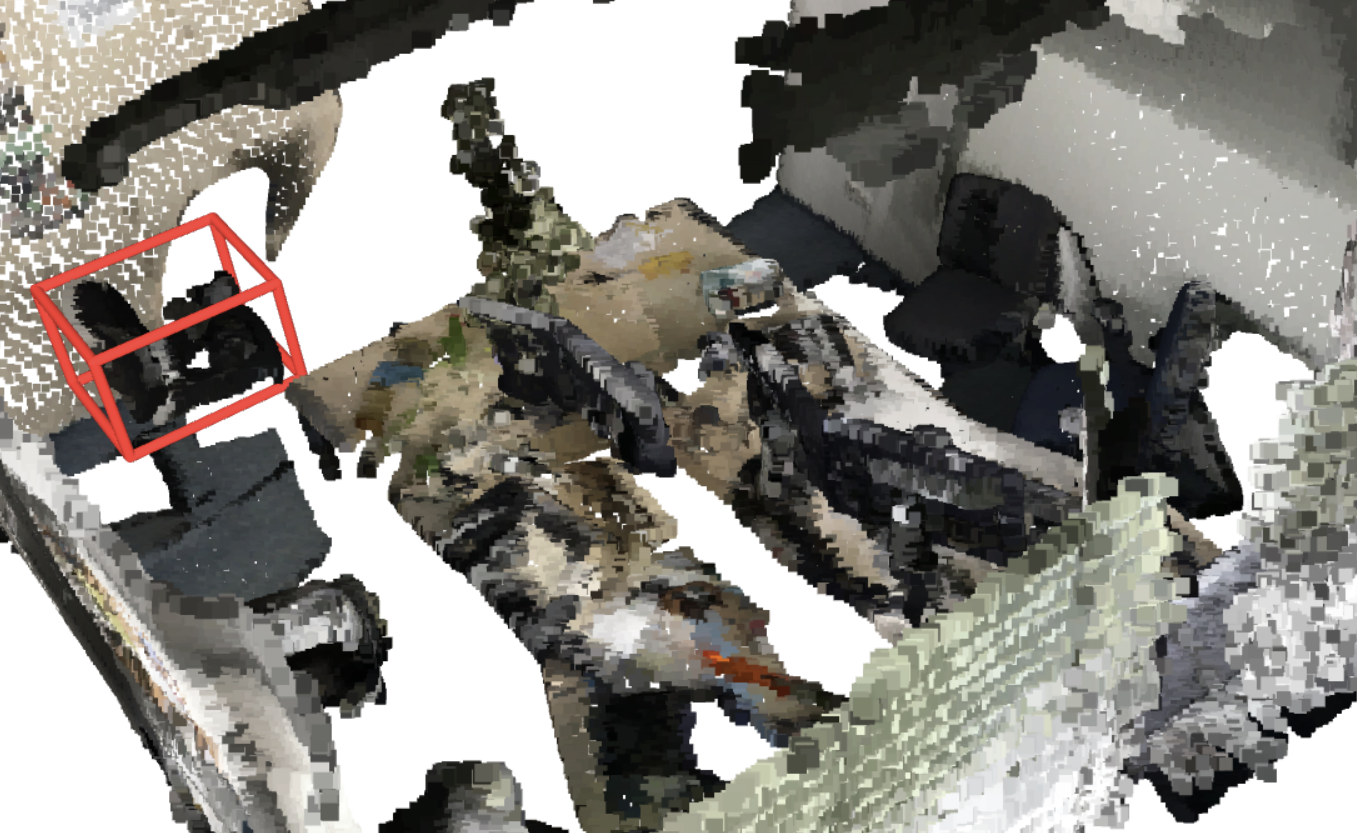} \\
 \queryrow{there is a black chair. placed on the left to the arm chair.} \\
\end{tabular}
\caption{Demonstrations of our proposed CSVG in comparison with ZSVG3Di.e.. Ground truth target: \textcolor[rgb]{0.4,0.7,1.0}{blue box}. Correct prediction: \textcolor[rgb]{0.3,1.0,0.3}{green box}. Incorrect prediction: \textcolor[rgb]{1.0,0.2,0.2}{red box}. Anchor objects discovered by CSVG: \textcolor[rgb]{1.0,0.7,0.0}{orange box}.}
\label{fig:more_example_csvg_vs_zsvg_1}
\end{figure*}

\begin{figure*}
\setlength\tabcolsep{2pt}
\renewcommand{\arraystretch}{1.2}
\centering
\begin{tabular}{cccc}
 \textbf{3DVG Ground Truth} &
 \textbf{Ours (GT Seg.)} &
 \textbf{Ours (Mask3D)} &
 \textbf{ZSVG3Di.e.} \\
 \includegraphics[width=0.24\linewidth]{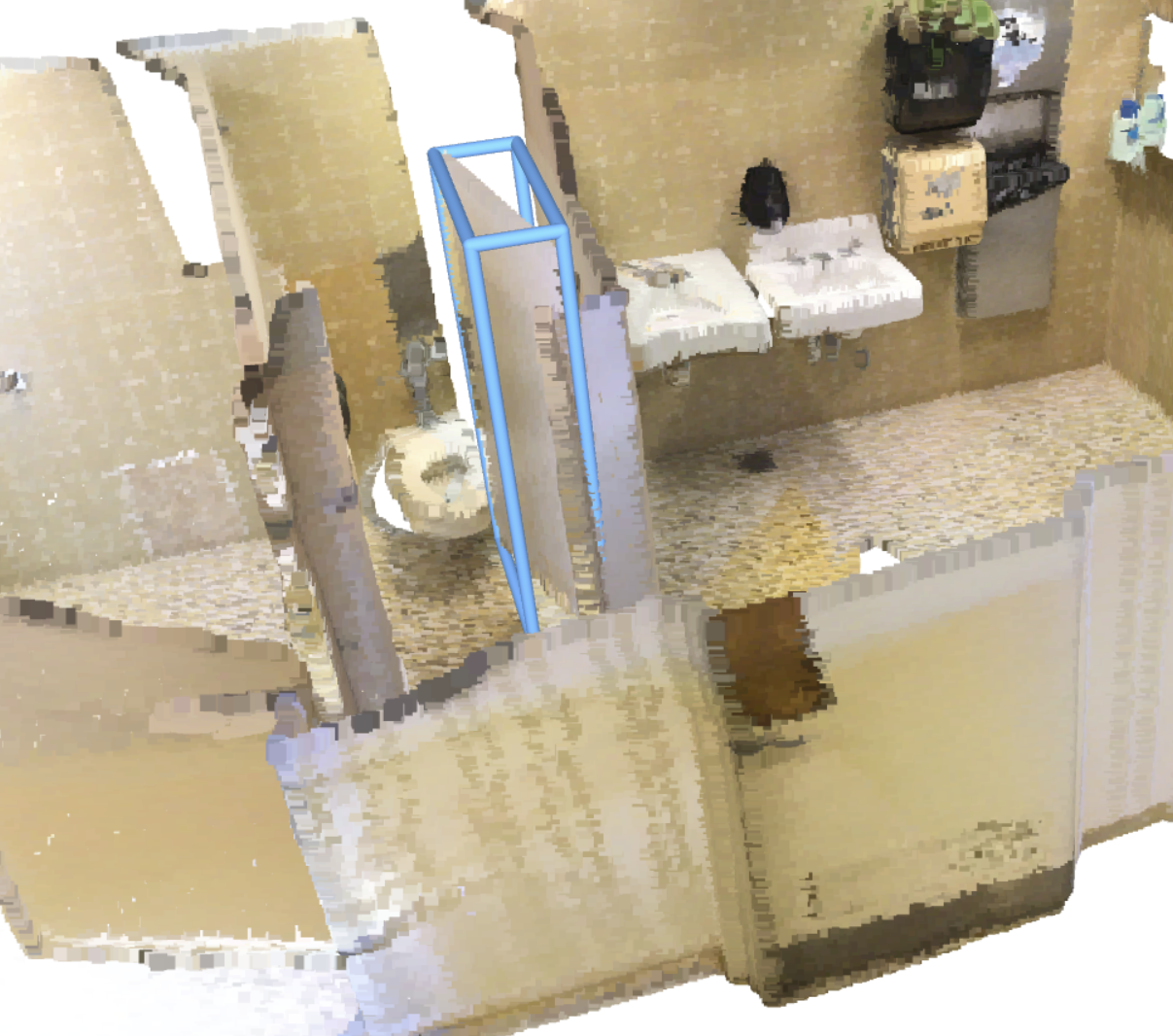} &
 \includegraphics[width=0.24\linewidth]{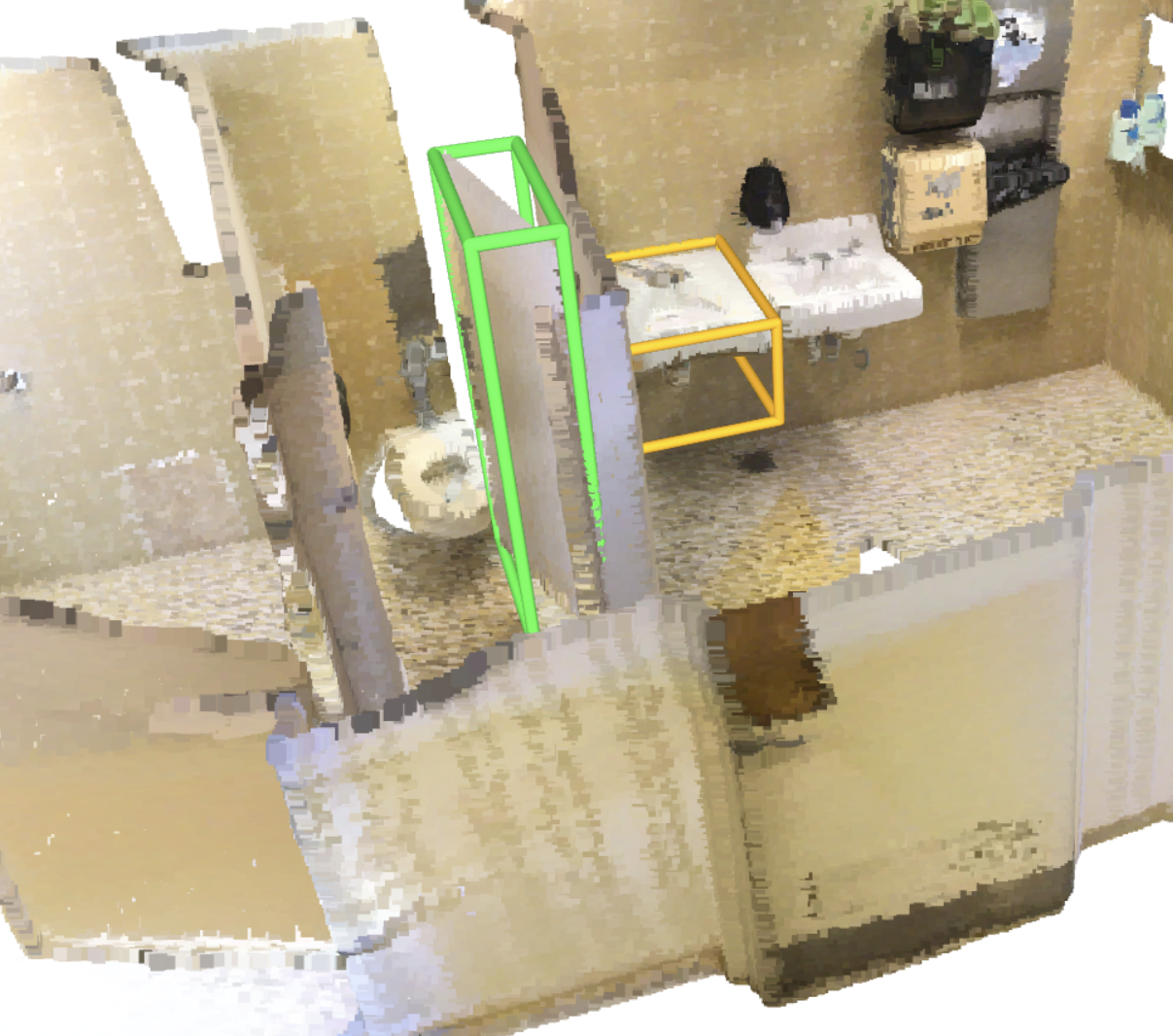} &
 \includegraphics[width=0.24\linewidth]{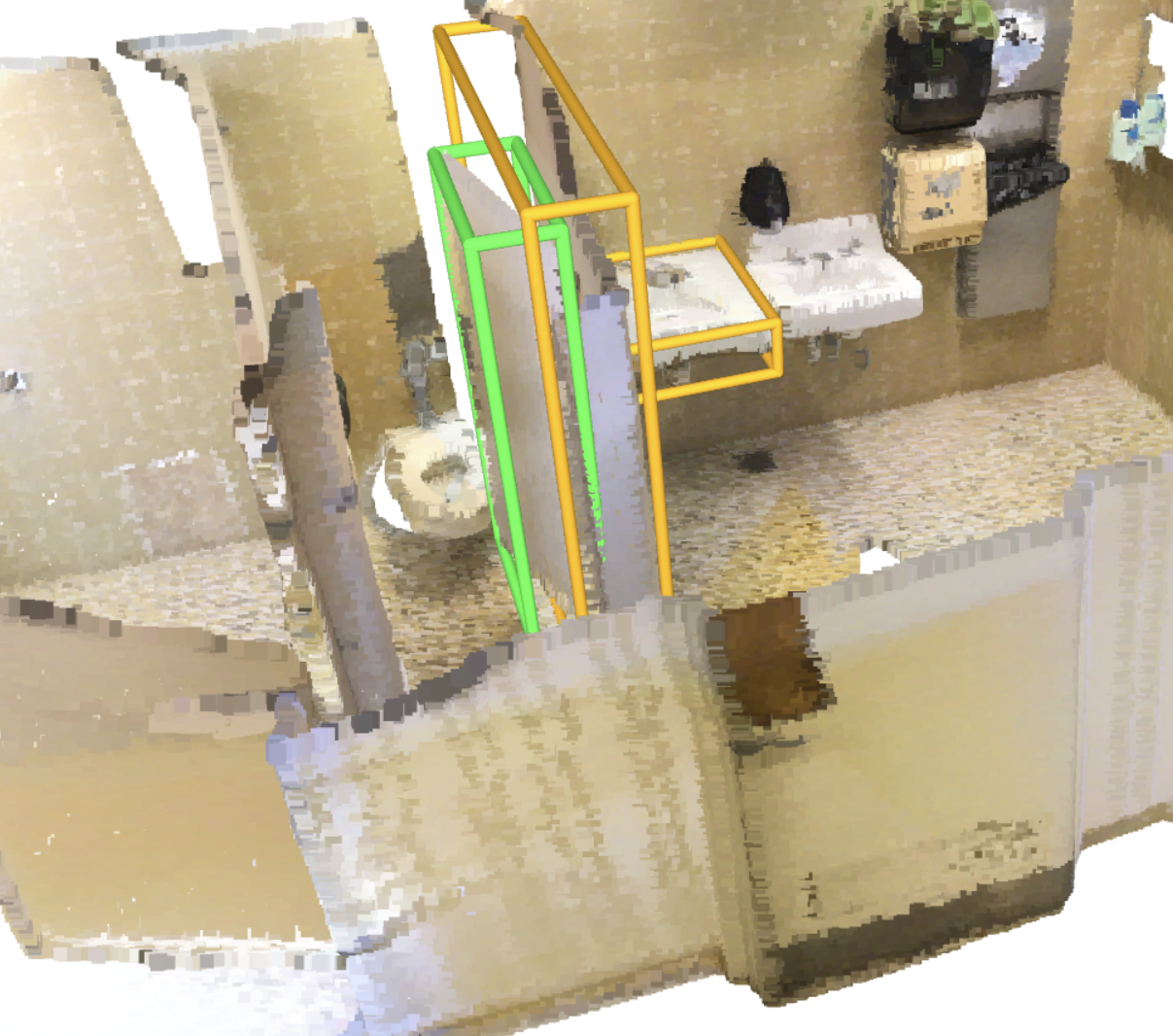} &
 \includegraphics[width=0.24\linewidth]{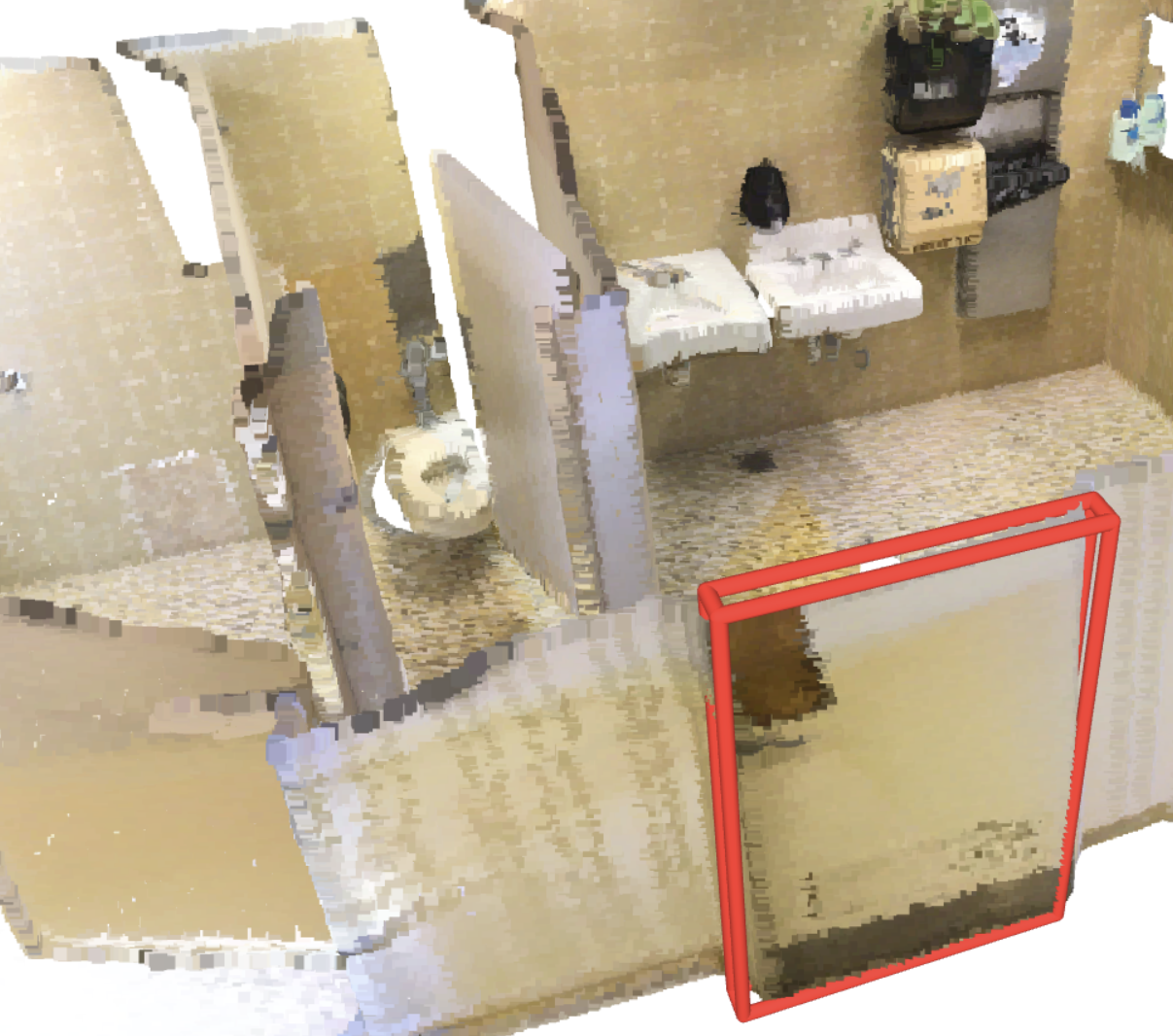} \\
 \queryrow{this is a tan bathroom stall door. this bathroom stall door is on the first stall to the left of the sinks.} \\
 \includegraphics[width=0.24\linewidth]{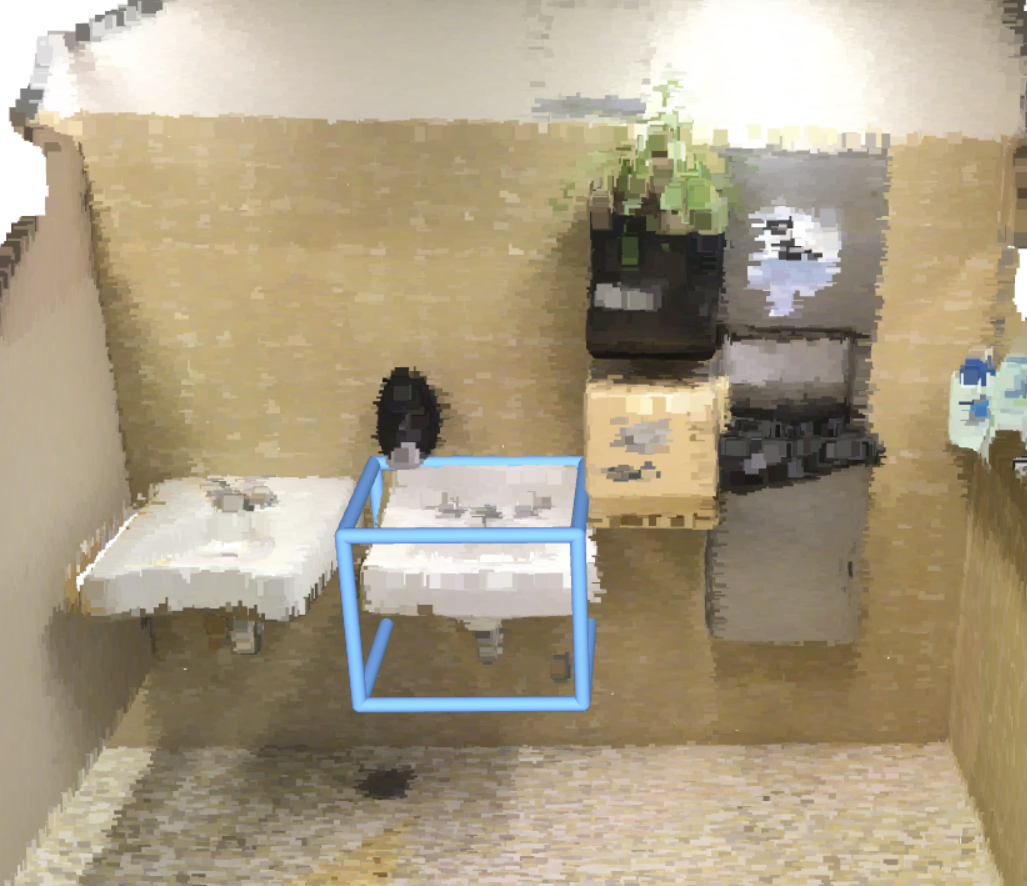} &
 \includegraphics[width=0.24\linewidth]{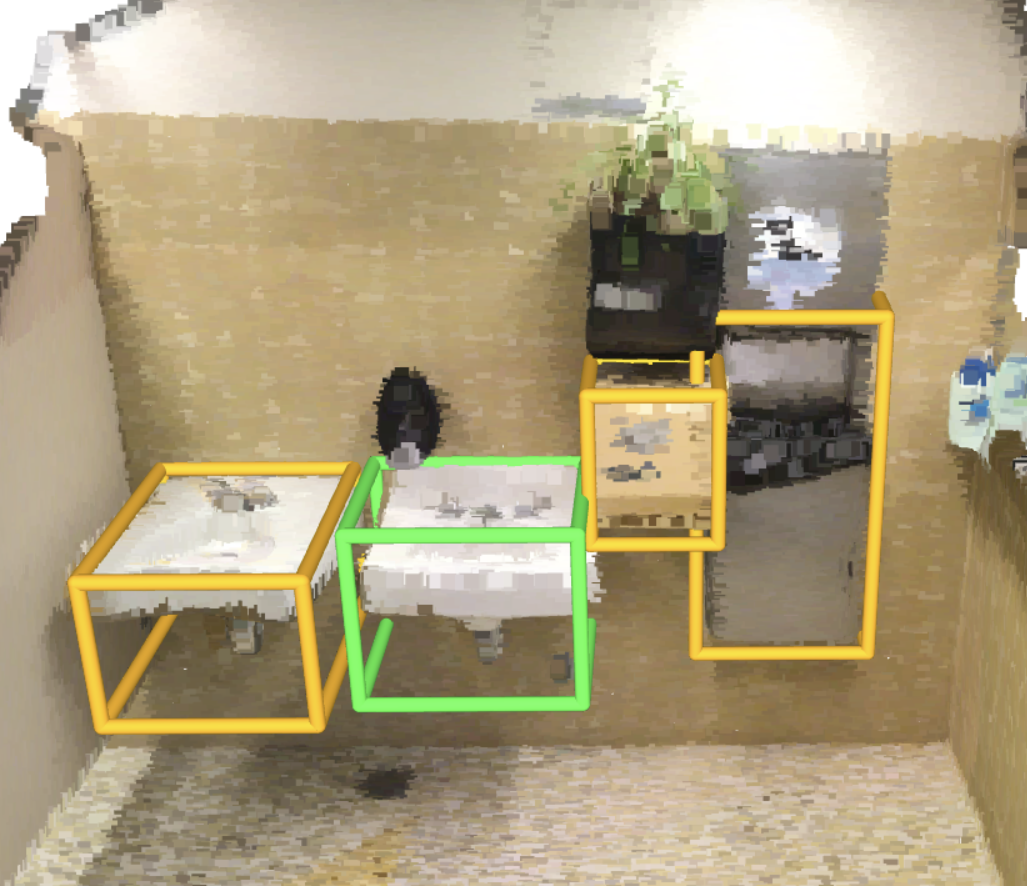} &
 \includegraphics[width=0.24\linewidth]{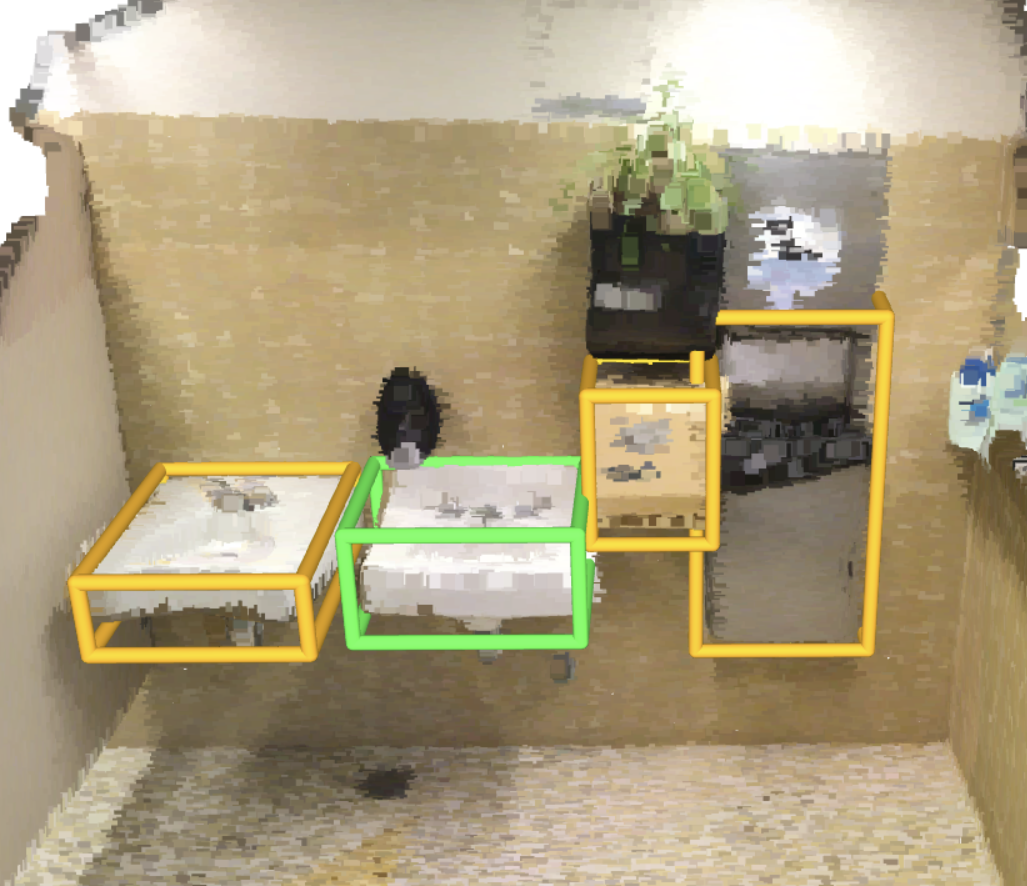} &
 \includegraphics[width=0.24\linewidth]{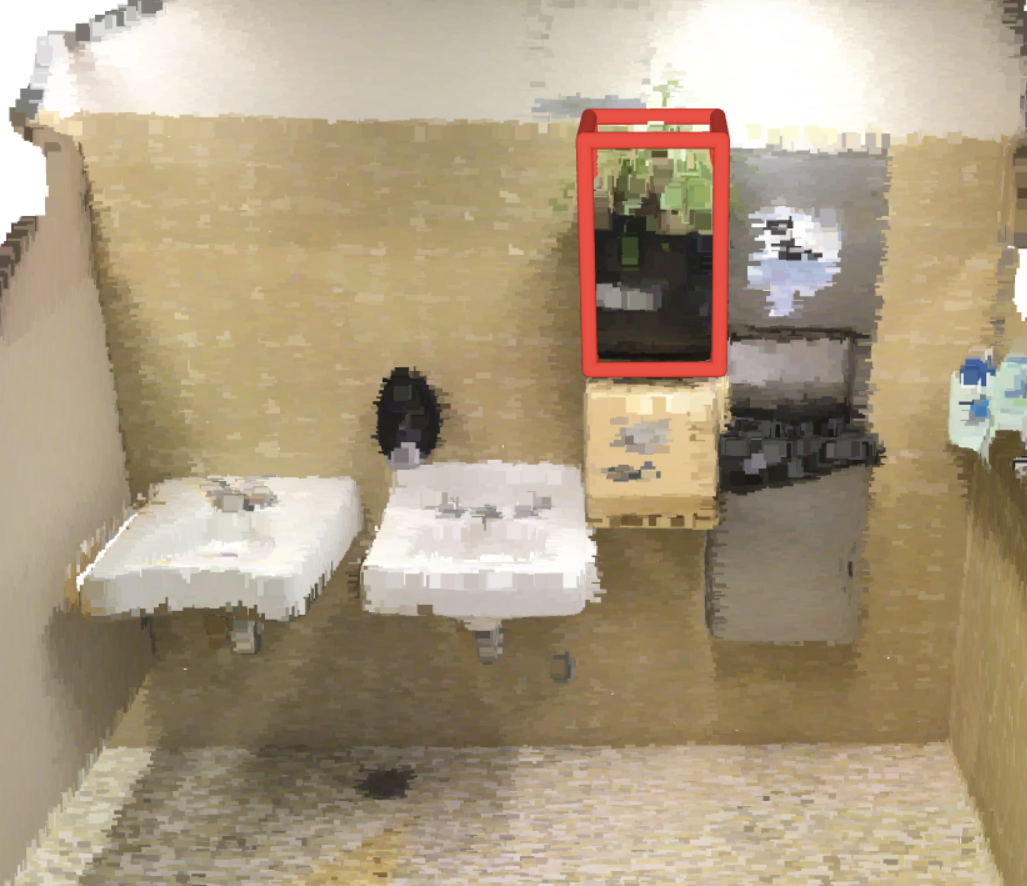} \\
 \queryrow{this sink is between another sink and a paper towel dispenser. a wall mounted trash can can also be seen besides the paper towel dispenser.} \\
 \includegraphics[width=0.24\linewidth]{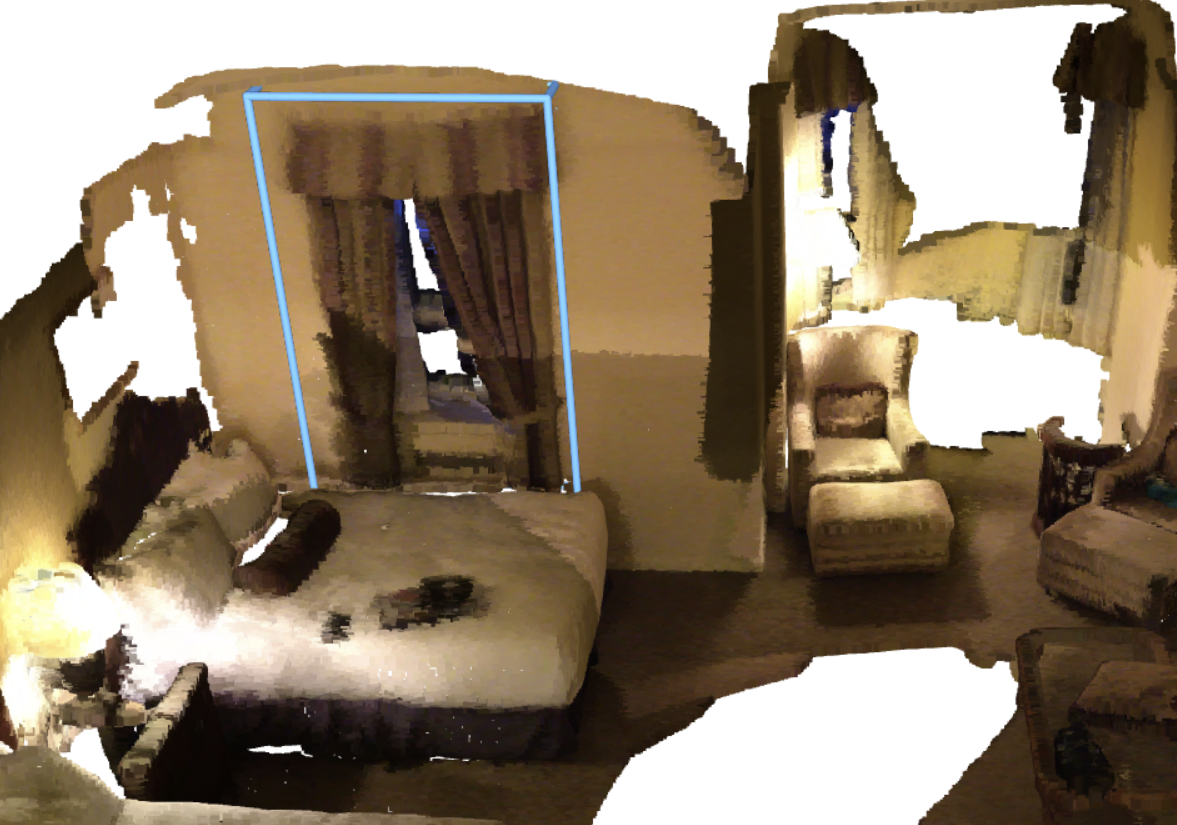} &
 \includegraphics[width=0.24\linewidth]{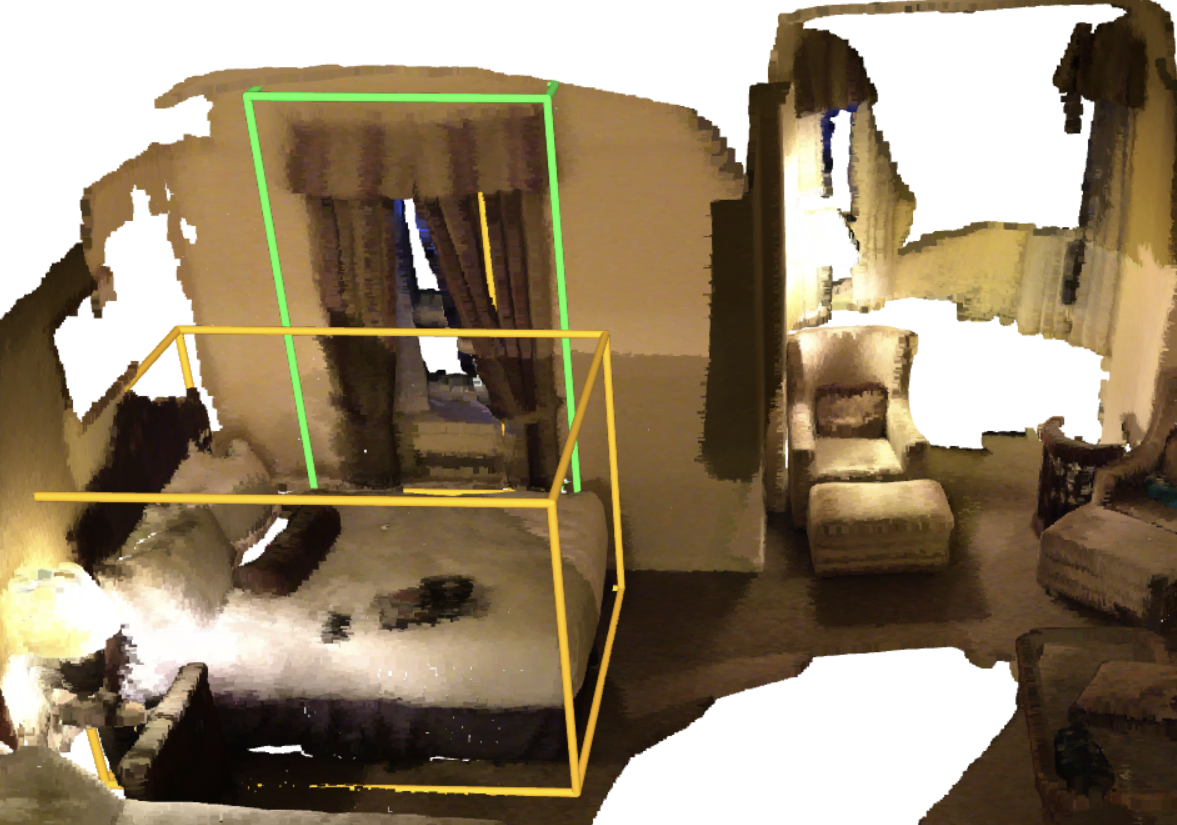} &
 \includegraphics[width=0.24\linewidth]{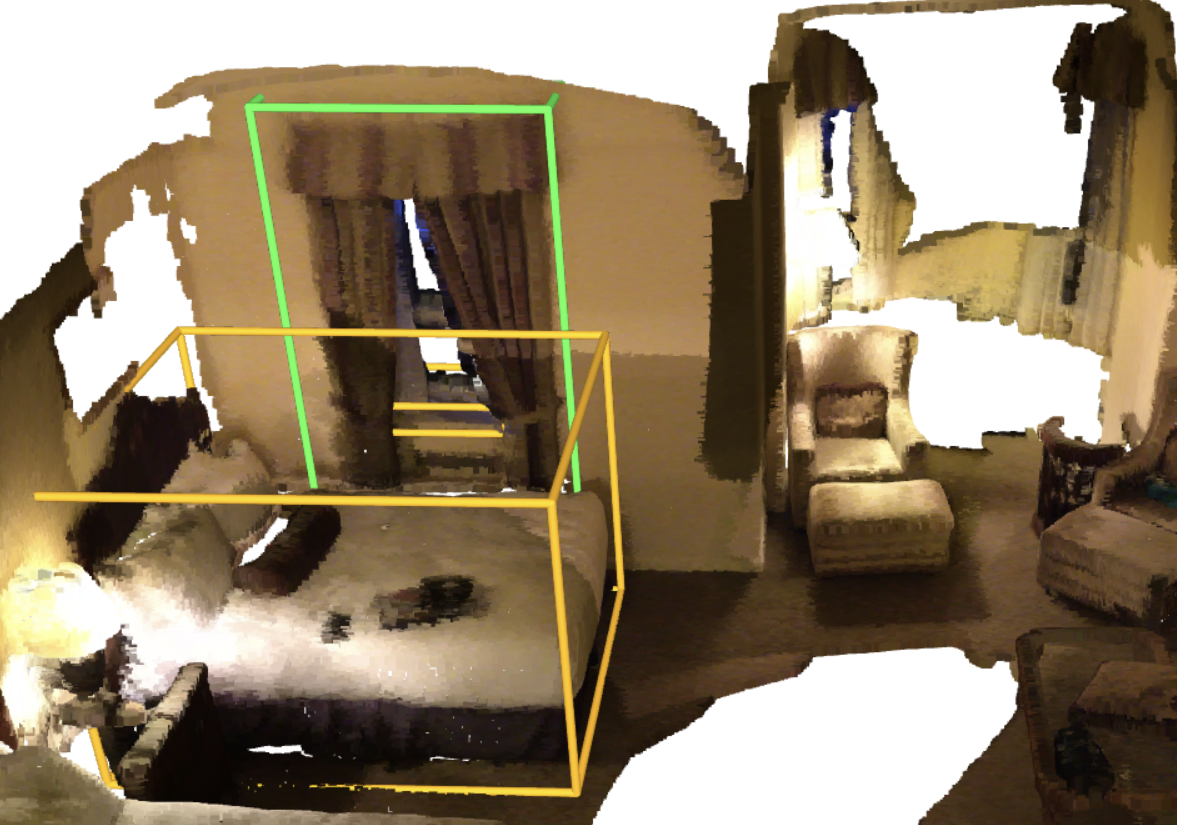} &
 \includegraphics[width=0.24\linewidth]{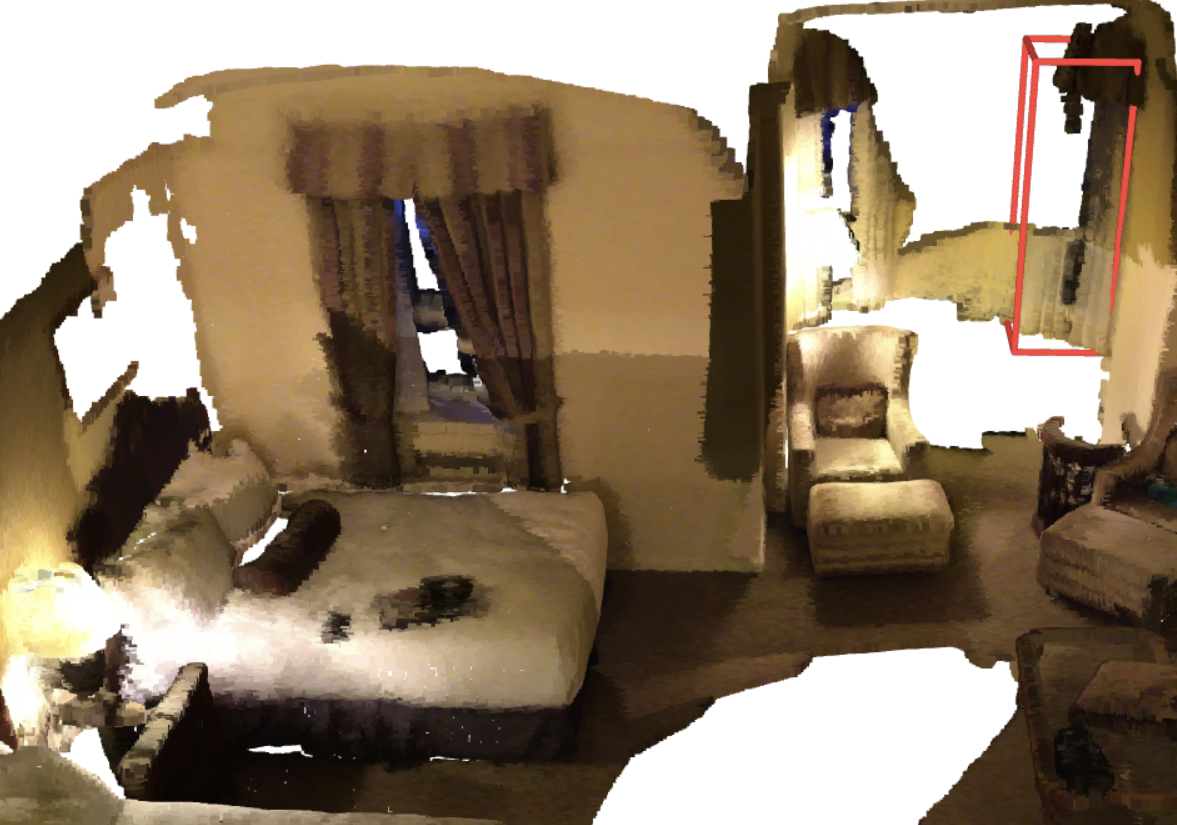} \\
 \queryrow{this is a brown curtain. it is drawn open in front of a window to the right of the bed.} \\
 \includegraphics[width=0.24\linewidth]{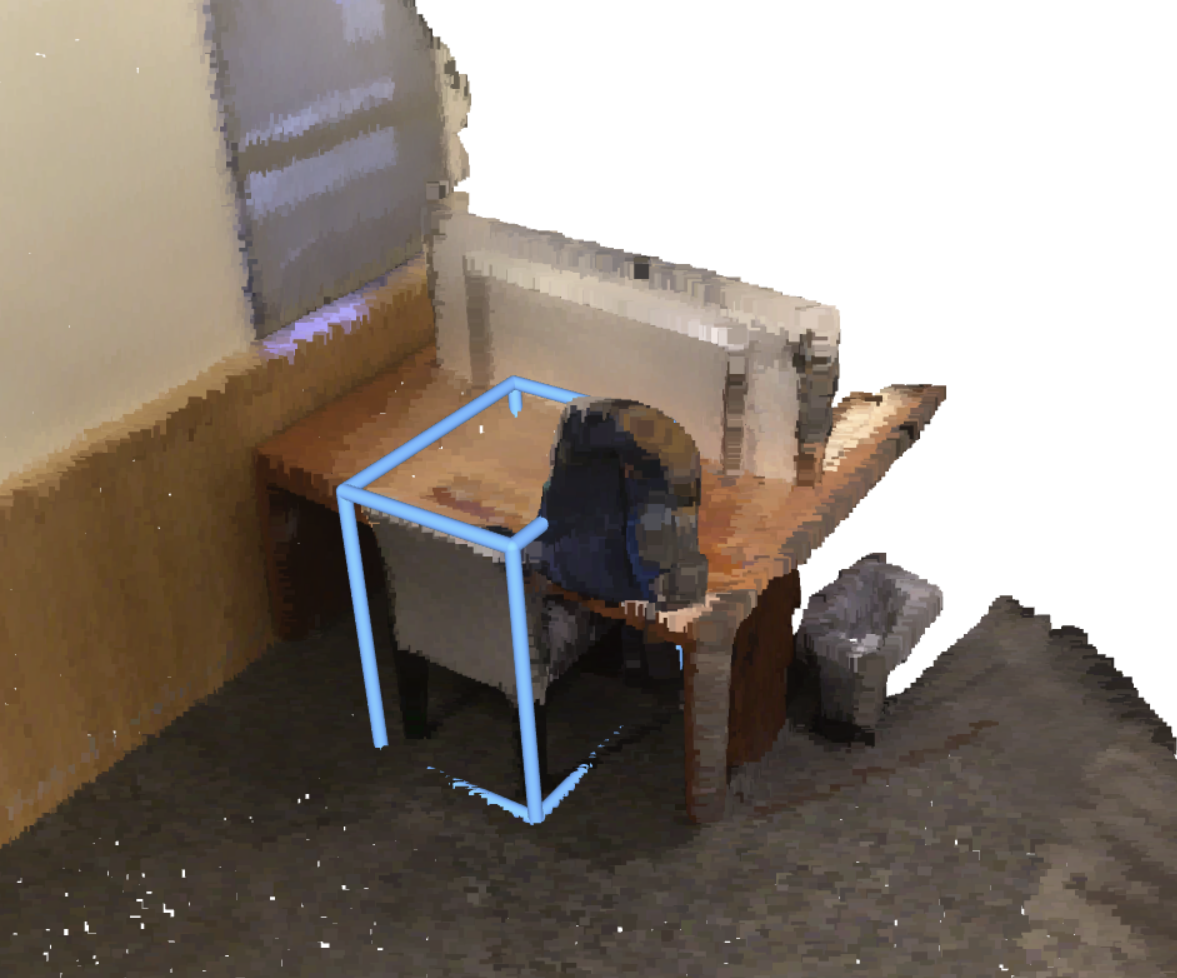} &
 \includegraphics[width=0.24\linewidth]{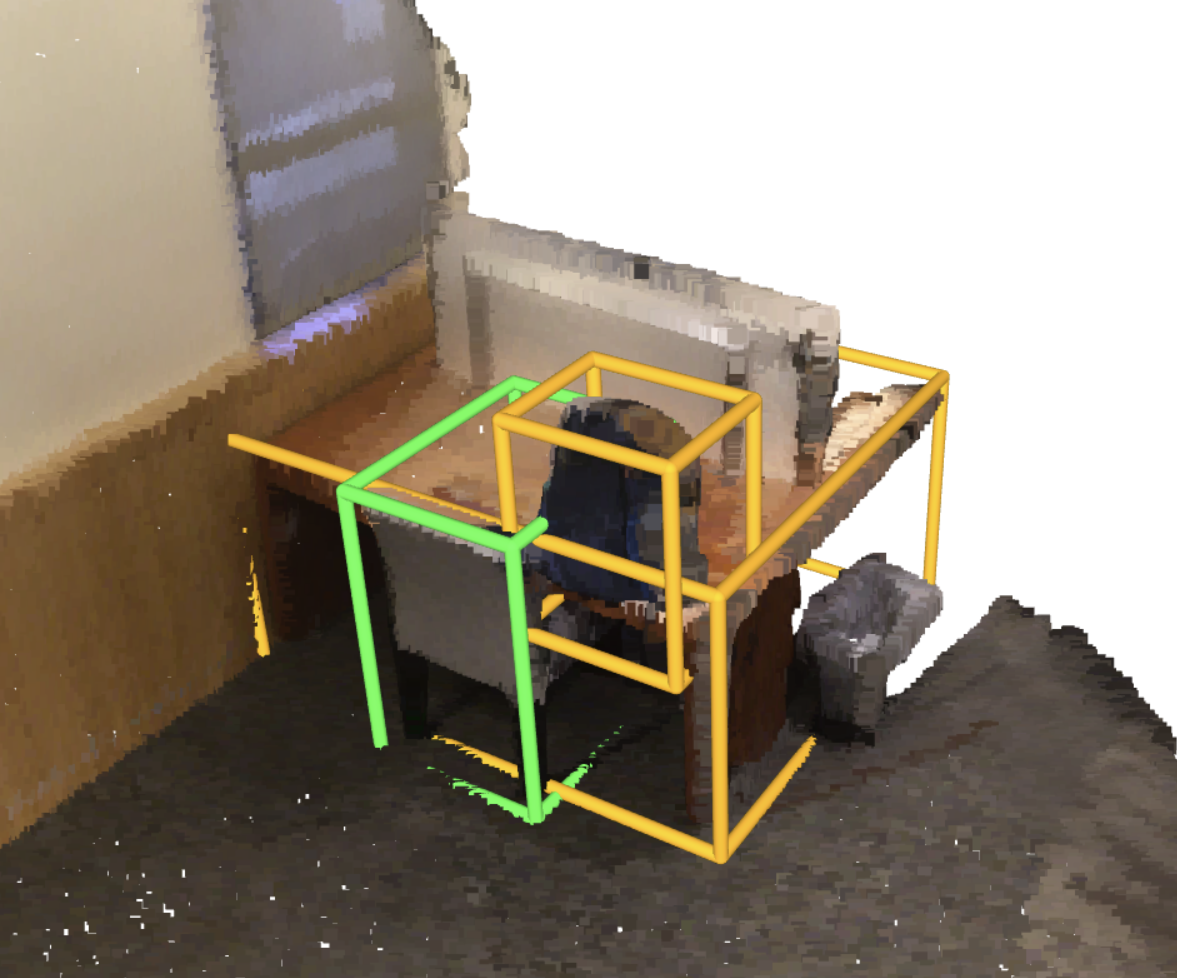} &
 \includegraphics[width=0.24\linewidth]{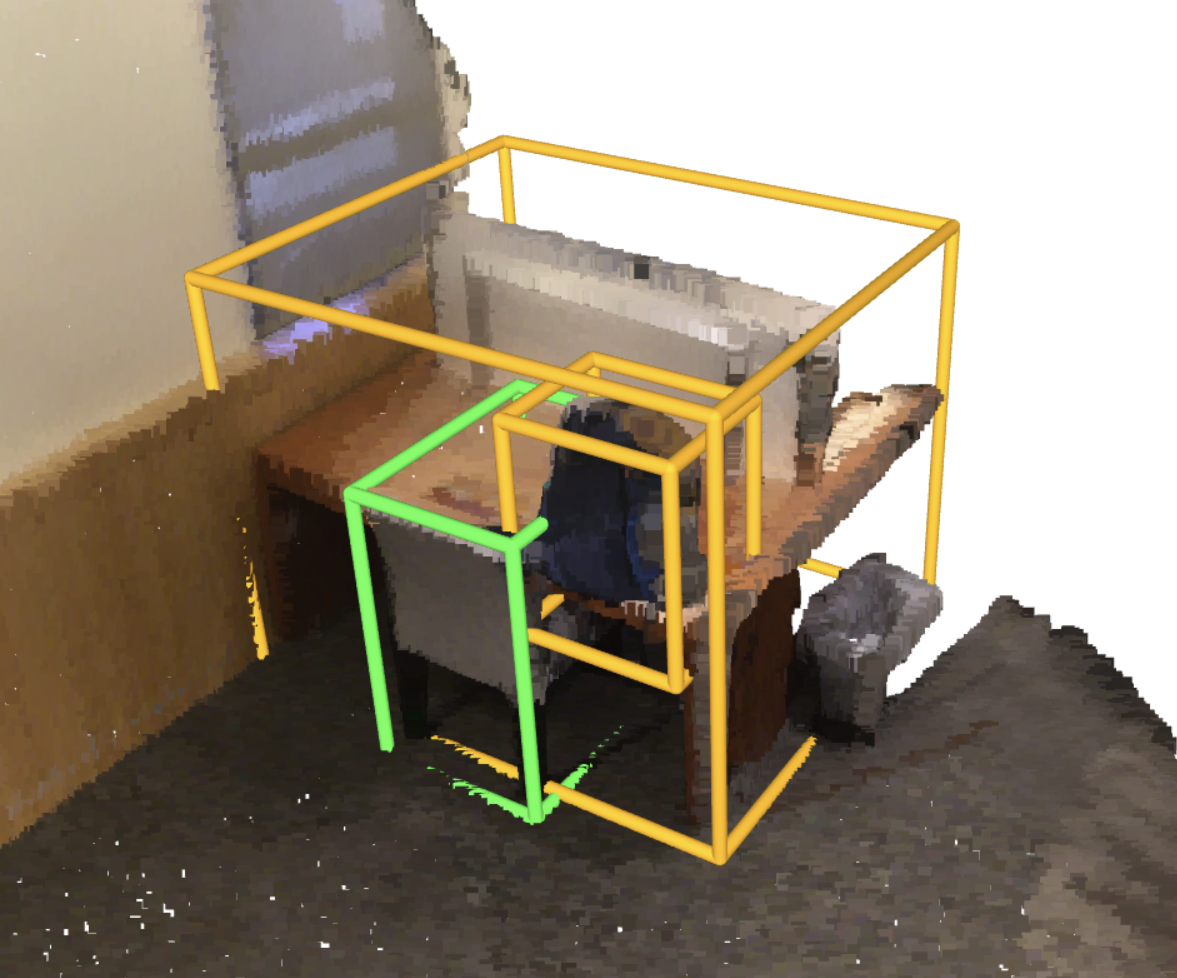} &
 \includegraphics[width=0.24\linewidth]{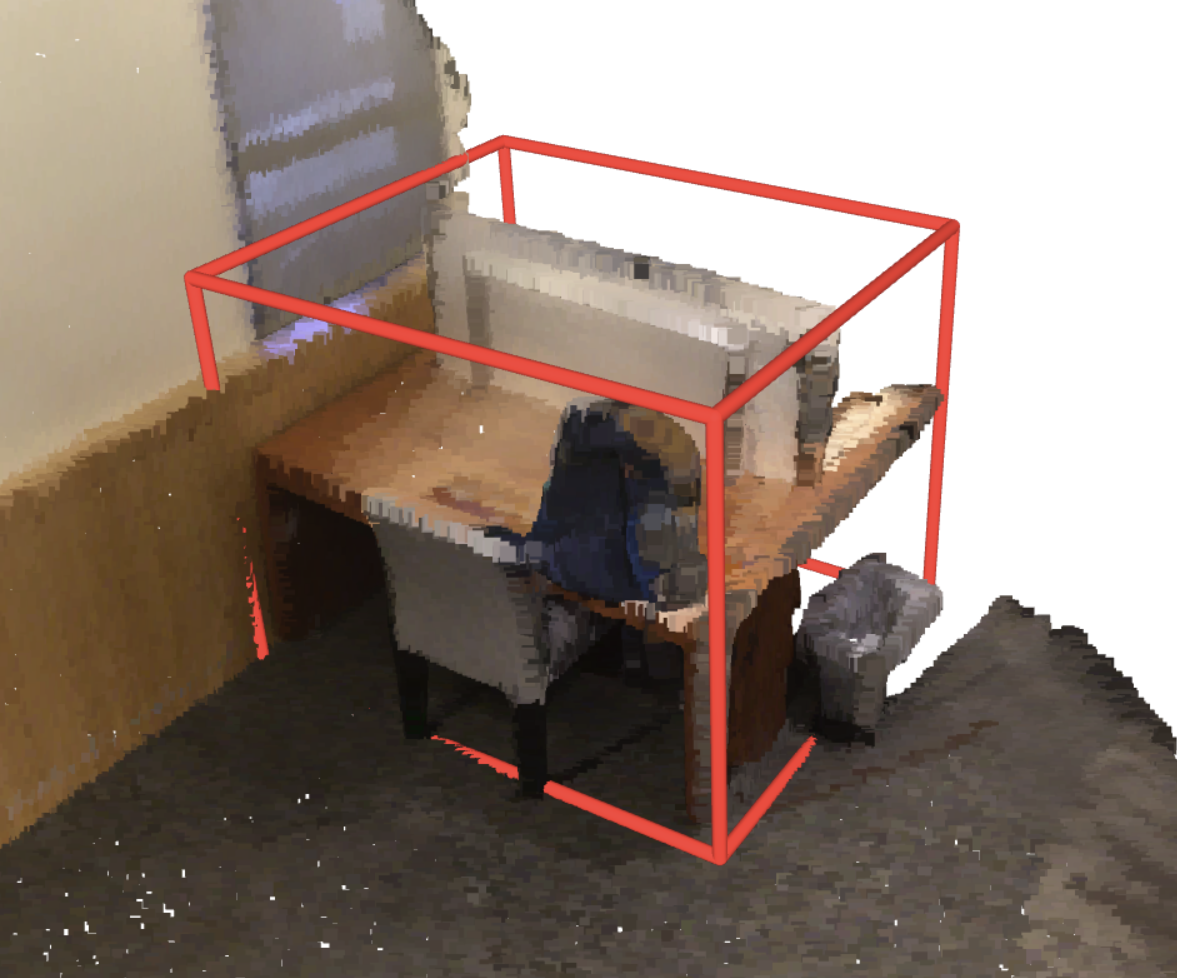} \\
 \queryrow{there is a rectangular gray chair. it is at the desk with a backpack on it.} \\
\end{tabular}
\caption{Demonstration of our proposed CSVG in comparison with ZSVG3Di.e.. Ground truth target: \textcolor[rgb]{0.4,0.7,1.0}{blue box}. Correct prediction: \textcolor[rgb]{0.3,1.0,0.3}{green box}. Incorrect prediction: \textcolor[rgb]{1.0,0.2,0.2}{red box}. Anchor objects discovered by CSVG: \textcolor[rgb]{1.0,0.7,0.0}{orange box}.}
\label{fig:more_example_csvg_vs_zsvg_2}
\end{figure*}

\newcommand{\querycell}[1]{{\parbox{1.0\linewidth}{\centering \footnotesize{\textbf{{#1} \vspace{10pt}}}}}}

\lstdefinestyle{mystyle2}{
    backgroundcolor=\color{backcolour},   
    commentstyle=\color{codegreen},
    keywordstyle=\color{magenta},
    stringstyle=\color{codepurple},
    basicstyle=\ttfamily\linespread{1.15}\tiny, 
    breakatwhitespace=false,         
    breaklines=true,                 
    captionpos=b,                    
    keepspaces=true,                 
    showspaces=false,                
    showstringspaces=false,
    showtabs=false,                  
    tabsize=2,
    captionpos=b,
    numbers=none,
    showlines=true,
}
\lstset{style=mystyle2}

\begin{figure*}
\setlength\tabcolsep{2pt}
\centering
\begin{tabular}{p{0.3\linewidth}p{0.3\linewidth}p{0.3\linewidth}}
\includegraphics[width=1.0\linewidth]{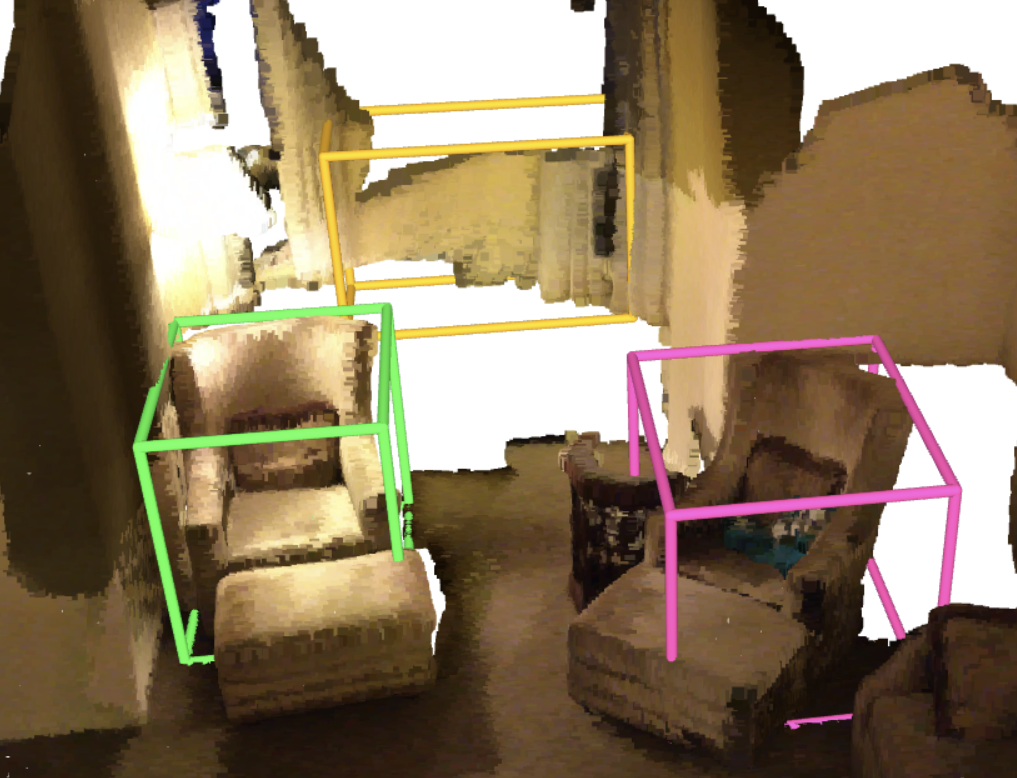} &
\includegraphics[width=1.0\linewidth]{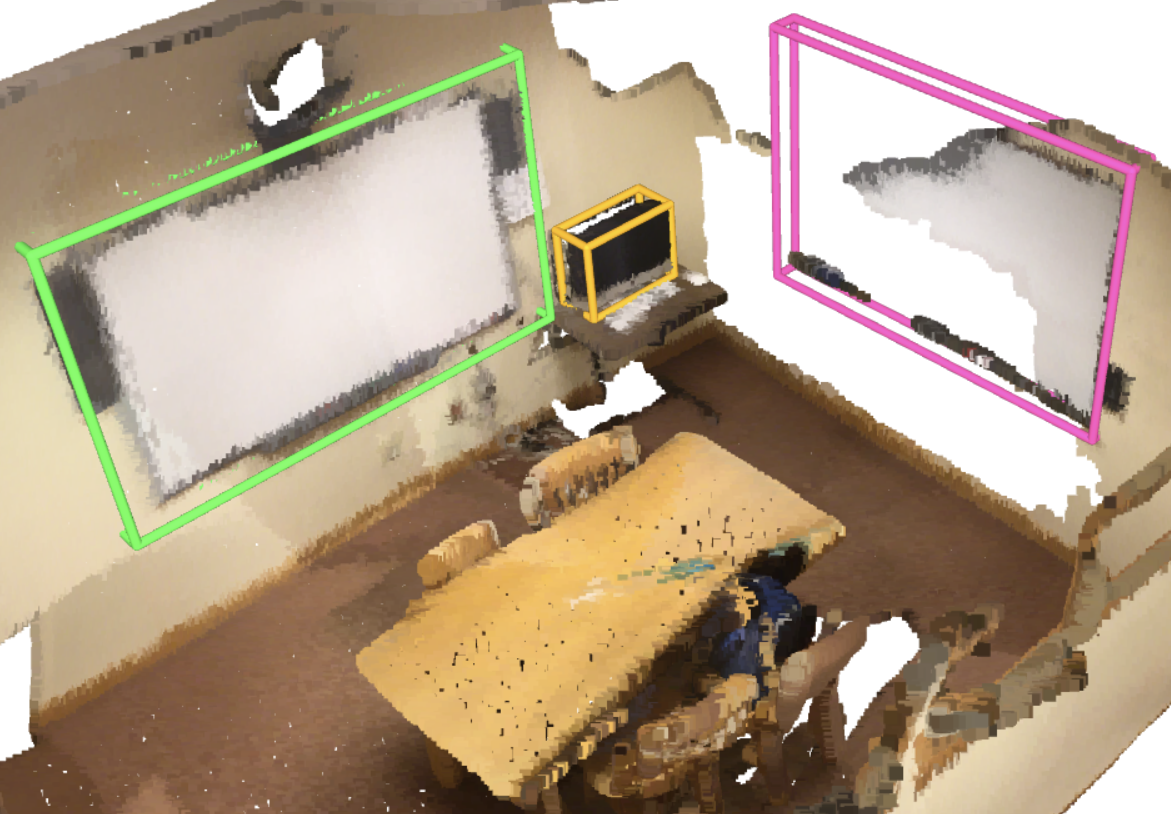} &
\includegraphics[width=1.0\linewidth]{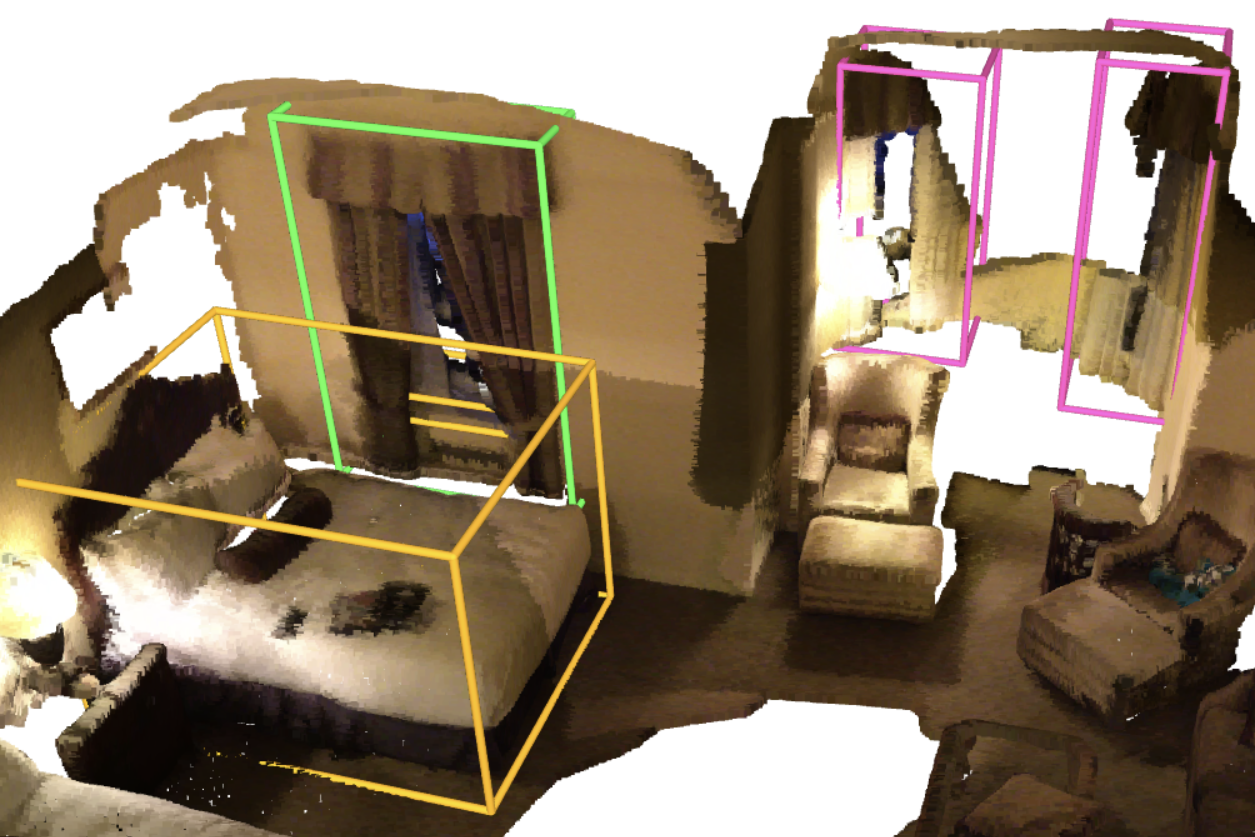} \\
\begin{minipage}{\linewidth}
\begin{lstlisting}
ARMCHAIR_0 = DEF_VAR(labels=["armchair"])
WINDOW_0 = DEF_VAR(labels=["window"])
LEFT(target=ARMCHAIR_0, anchor=WINDOW_0)
SET_TARGET(ARMCHAIR_0)


\end{lstlisting}
\end{minipage} &
\begin{minipage}{\linewidth}
\begin{lstlisting}
WHITEBOARD = DEF_VAR(labels=["whiteboard"])
MONITOR = DEF_VAR(labels=["monitor"])
LEFT(target=WHITEBOARD, anchor=MONITOR)
SET_TARGET(WHITEBOARD)


\end{lstlisting}
\end{minipage} &
\begin{minipage}{\linewidth}
\begin{lstlisting}
CURTAINS = DEF_VAR(labels=["curtain"])
WINDOWS = DEF_VAR(labels=["window"])
BED = DEF_VAR(labels=["bed"])
ABOVE(target=CURTAINS, anchor=WINDOWS)
BEHIND(target=CURTAINS, anchor=BED)
SET_TARGET(CURTAINS)
\end{lstlisting}
\end{minipage} \\
\querycell{the armchair is beige colored. it is to the left of the window.} &
\querycell{there is a rectangular whiteboard. it is on the wall and to the left of a monitor.} &
\querycell{these are the curtains over the windows. they are just behind the bed.} \\
\\
\includegraphics[width=1.0\linewidth]{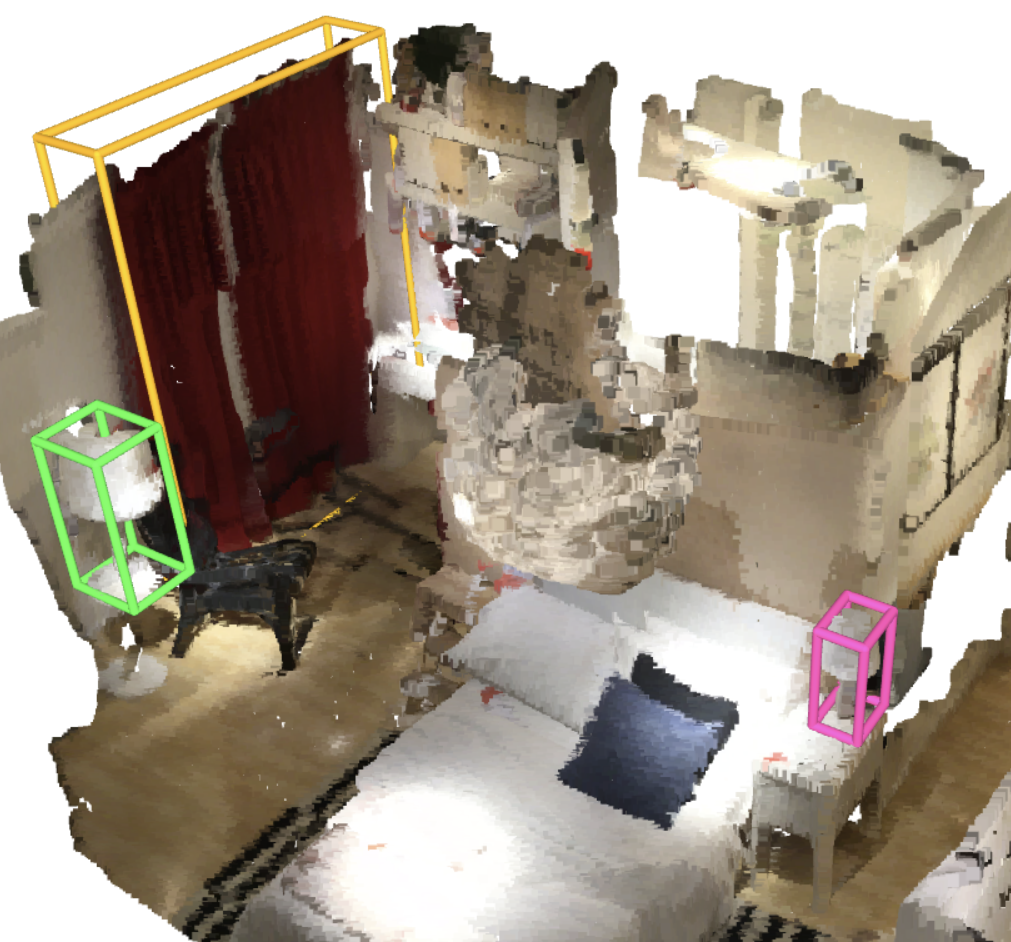} &
\includegraphics[width=1.0\linewidth]{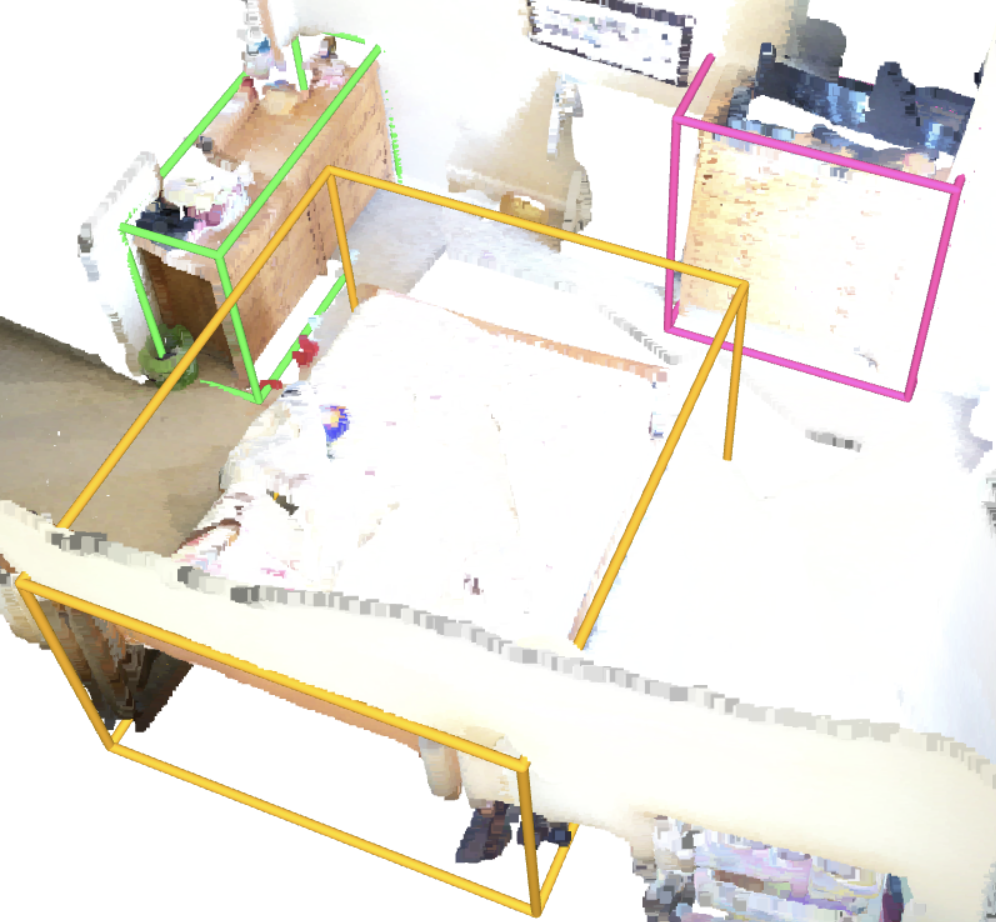} &
\includegraphics[width=1.0\linewidth]{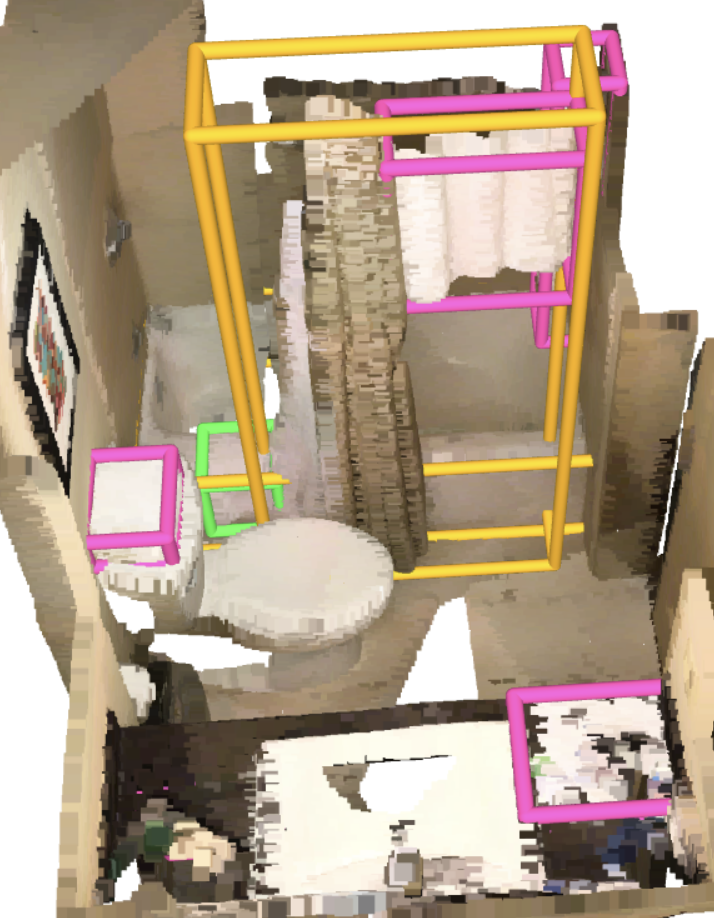} \\
\begin{minipage}{\linewidth}
\begin{lstlisting}
LAMP_0 = DEF_VAR(labels=["lamp"])
CURTAIN_0 = DEF_VAR(labels=["curtain"])
LEFT(target=LAMP_0, anchor=CURTAIN_0)
SET_TARGET(LAMP_0)


\end{lstlisting}
\end{minipage} &
\begin{minipage}{\linewidth}
\begin{lstlisting}
DRESSER_0 = DEF_VAR(labels=["dresser"])
BED_0 = DEF_VAR(labels=["bed"])
NEAR(target=DRESSER_0, anchor=BED_0)
RIGHT(target=DRESSER_0, anchor=BED_0)
SET_TARGET(DRESSER_0)

\end{lstlisting}
\end{minipage} &
\begin{minipage}{\linewidth}
\begin{lstlisting}
TOWEL_0 = DEF_VAR(labels=["towel"])
BATHTUB_0 = DEF_VAR(labels=["bathtub"])
CURTAIN_0 = DEF_VAR(labels=["shower curtain"])
ON(target=TOWEL_0, anchor=BATHTUB_0)
LEFT(target=TOWEL_0, anchor=CURTAIN_0)
SET_TARGET(TOWEL_0)
\end{lstlisting}
\end{minipage} \\
\querycell{this is a lamp. it is white in color to the left of burgundy drapes.} &
\querycell{this is a dresser in the bedroom. it is near the right foot of the bed.} &
\querycell{there is a towel hanging on the edge of the bathtub. it is on the left side of the shower curtain.} \\
\end{tabular}
\caption{Demonstration of the proposed CSVG with only spatial constraints. Distractors (objects with the same label as the target): \textcolor[rgb]{1.0,0.3,0.7}{magenta box}. Correct prediction: \textcolor[rgb]{0.3,1.0,0.3}{green box}. Anchor objects discovered by CSVG: \textcolor[rgb]{1.0,0.7,0.0}{orange box}. In the generated programs, the {\ttfamily DEFINE\_VARIABLE} function is abbreviated as {\ttfamily DEF\_VAR}, and we remove the {\ttfamily CONSTRAINT\_} prefix in constraints for simplicity.}
\label{fig:more_example_csvg_no_minmax}
\end{figure*}

\begin{figure*}
\setlength\tabcolsep{2pt}
\centering
\begin{tabular}{p{0.3\linewidth}p{0.3\linewidth}p{0.3\linewidth}}
\includegraphics[width=1.0\linewidth]{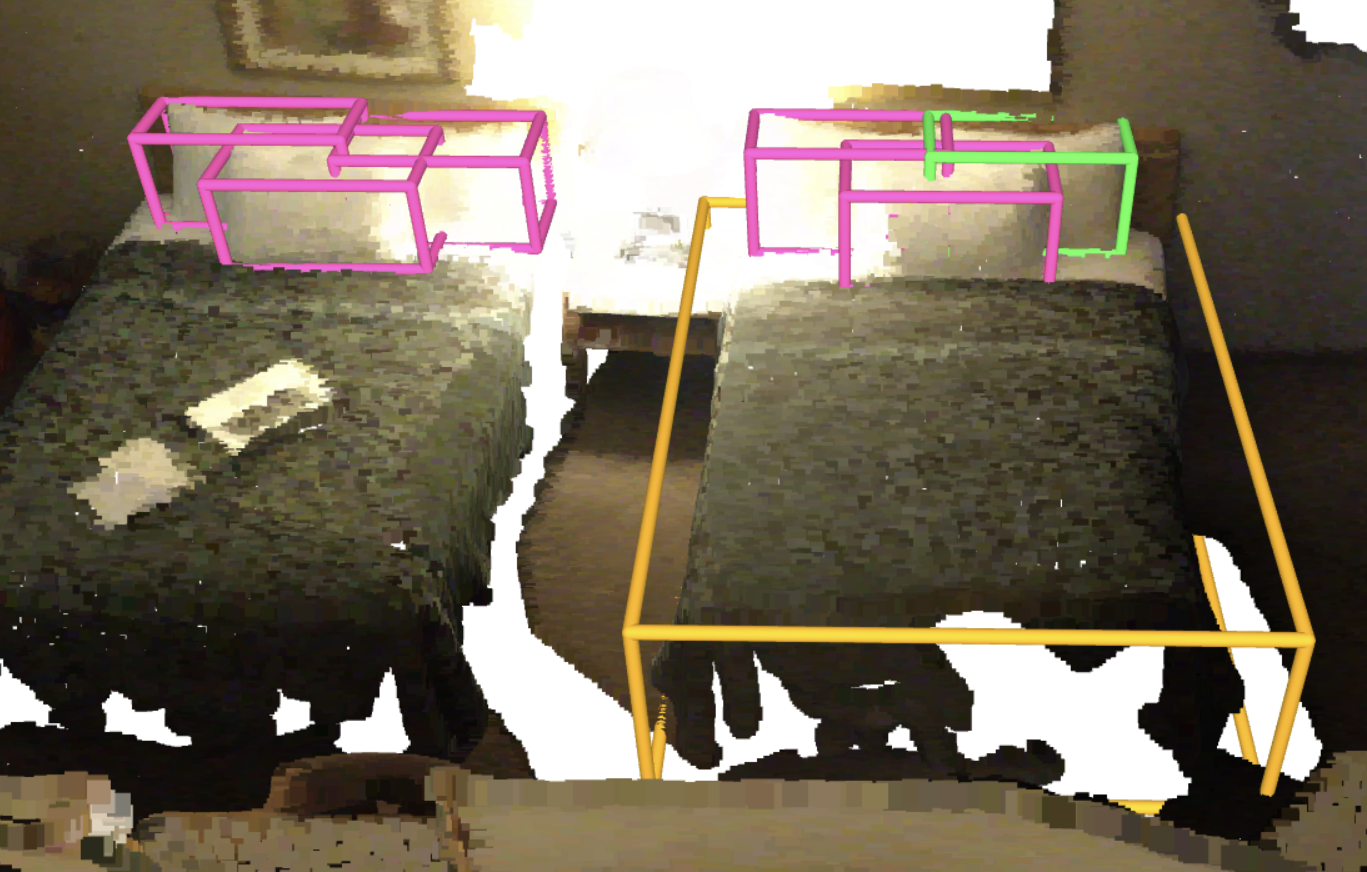} &
\includegraphics[width=1.0\linewidth]{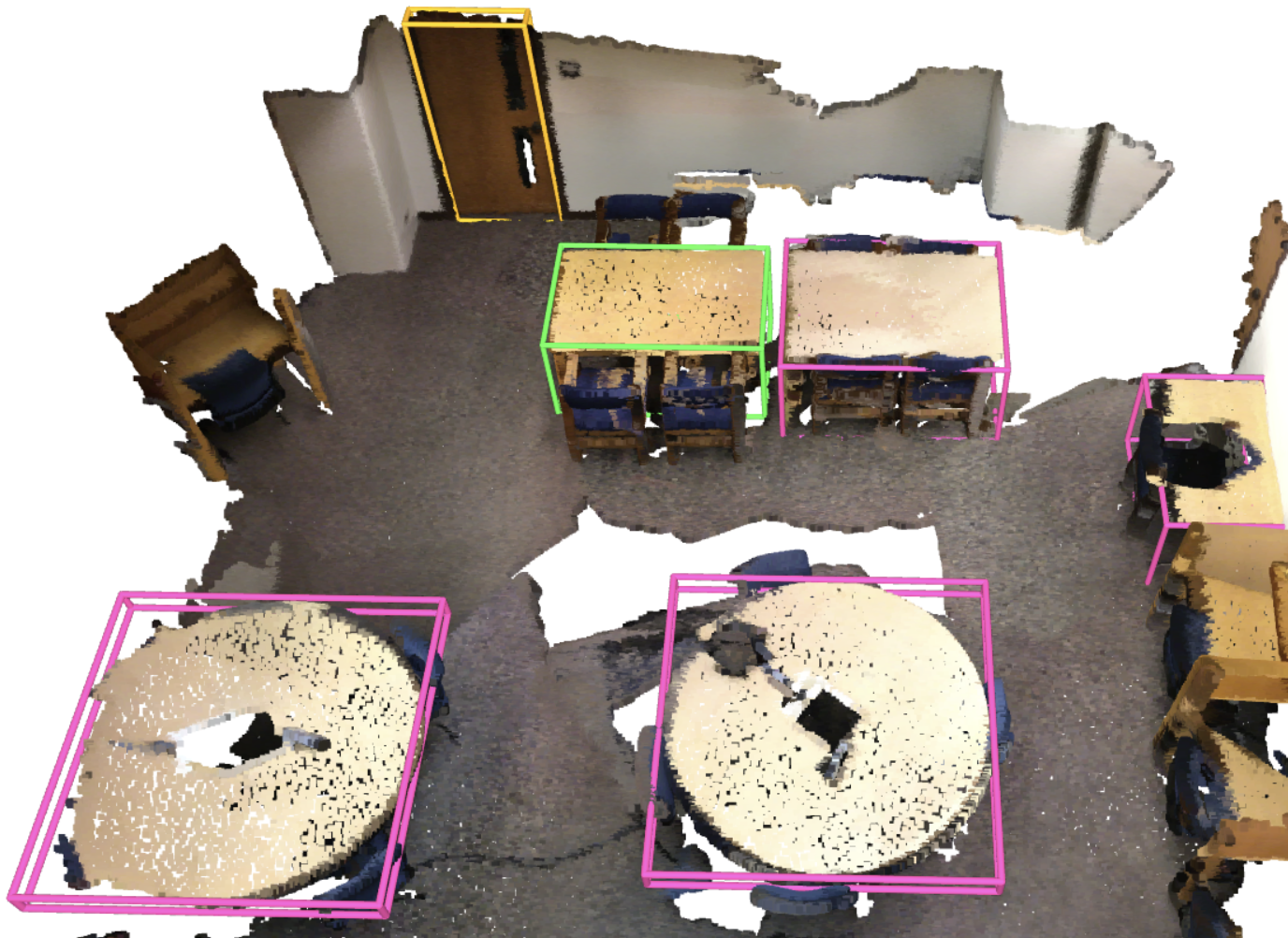} &
\includegraphics[width=1.0\linewidth]{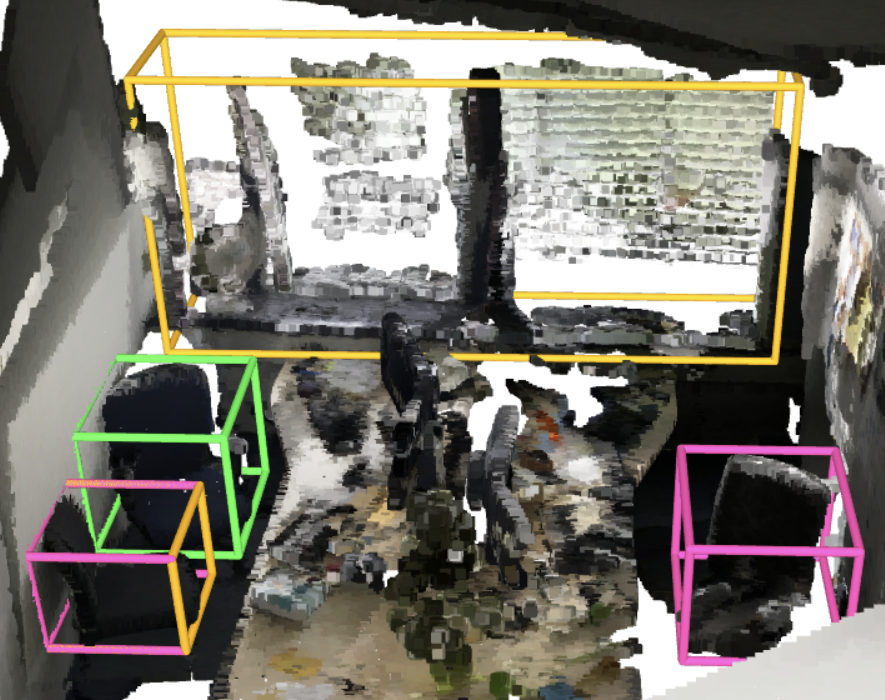} \\
\begin{minipage}{\linewidth}
\begin{lstlisting}
PILLOW_0 = DEF_VAR(labels=["pillow"])
BED_0 = DEF_VAR(labels=["bed"])
ON(target=PILLOW_0, anchor=BED_0)
MAX_OF(target=PILLOW_0, score_func="right")
SET_TARGET(PILLOW_0)



\end{lstlisting}
\end{minipage} &
\begin{minipage}{\linewidth}
\begin{lstlisting}
TABLE_0 = DEF_VAR(labels=["table", "desk"])
DOOR_0 = DEF_VAR(labels=["door"])
MIN_OF(target=TABLE_0, 
       score_func="distance", 
       anchor=DOOR_0)
SET_TARGET(TABLE_0)


\end{lstlisting}
\end{minipage} &
\begin{minipage}{\linewidth}
\begin{lstlisting}
SWIVEL_CHAIR = DEF_VAR(labels=["chair"])
STATIONARY_CHAIR = DEF_VAR(labels=["chair"])
WINDOW = DEF_VAR(labels=["window"])
BESIDE(target=SWIVEL_CHAIR, 
       anchor=STATIONARY_CHAIR)
MIN_OF(target=SWIVEL_CHAIR, 
       score_func="distance", anchor=WINDOW)
SET_TARGET(SWIVEL_CHAIR)
\end{lstlisting}
\end{minipage} \\
\querycell{the pillow is the rightmost one that is on top of the bed. the pillow is a caramel white color and rectangular.} &
\querycell{the table is the closest one to the entrance door. the table is light brown and square.} &
\querycell{a swivel chair sits sandwhiched between a stationary chair and a fan. it's the closest chair to the window.} \\
\\
\includegraphics[width=1.0\linewidth]{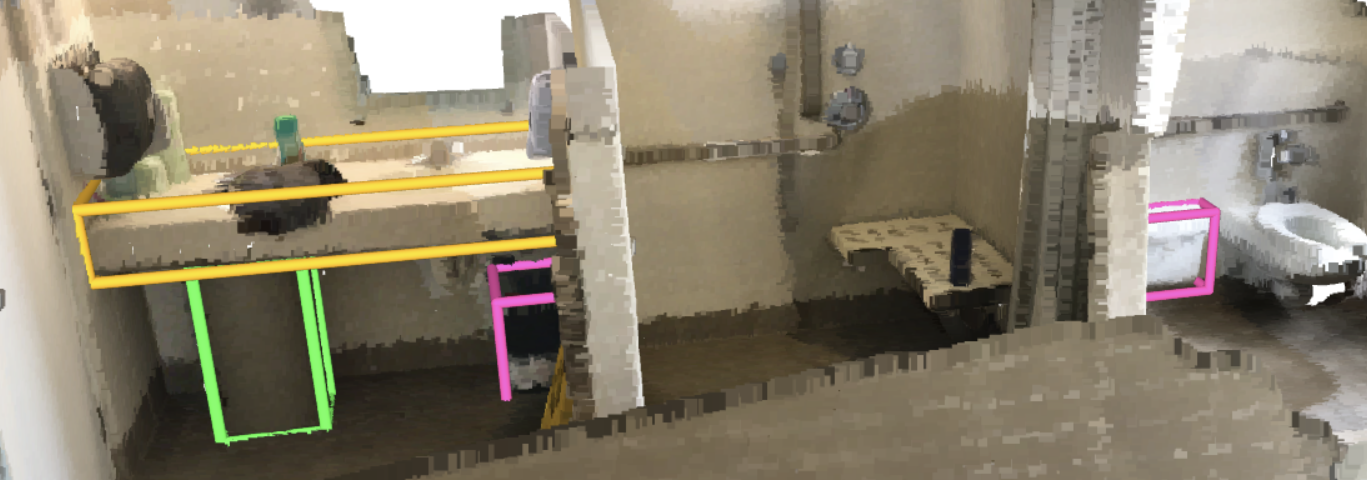} &
\includegraphics[width=1.0\linewidth]{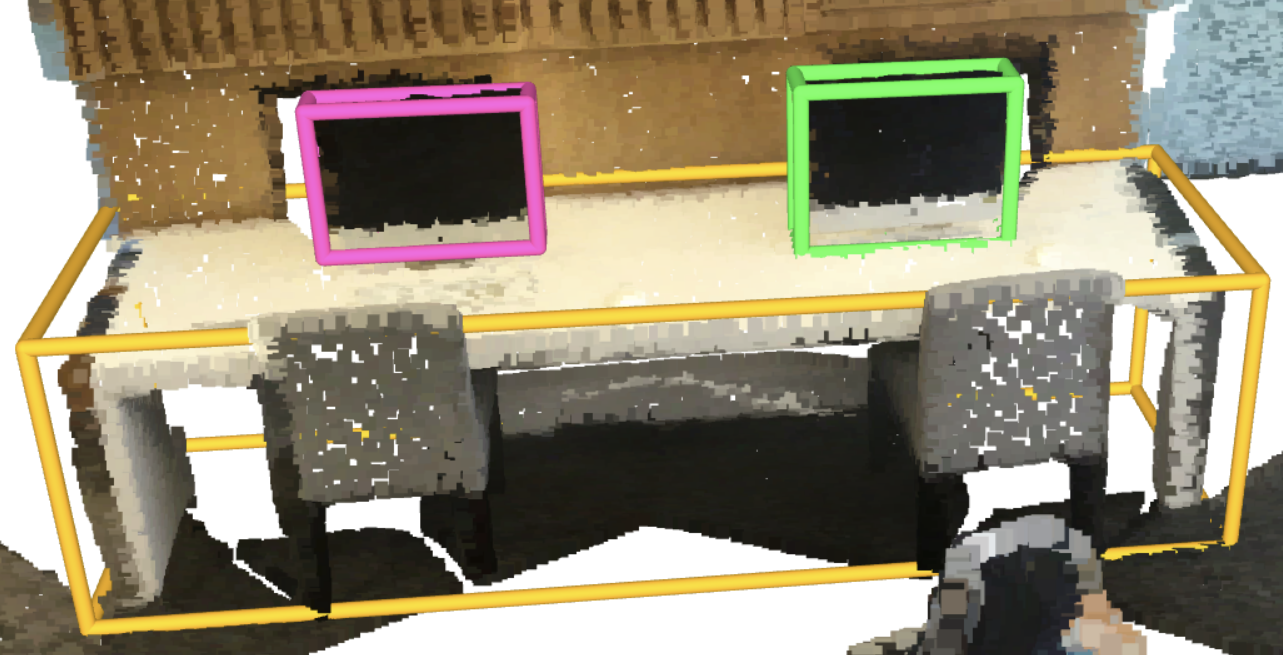} &
\includegraphics[width=1.0\linewidth]{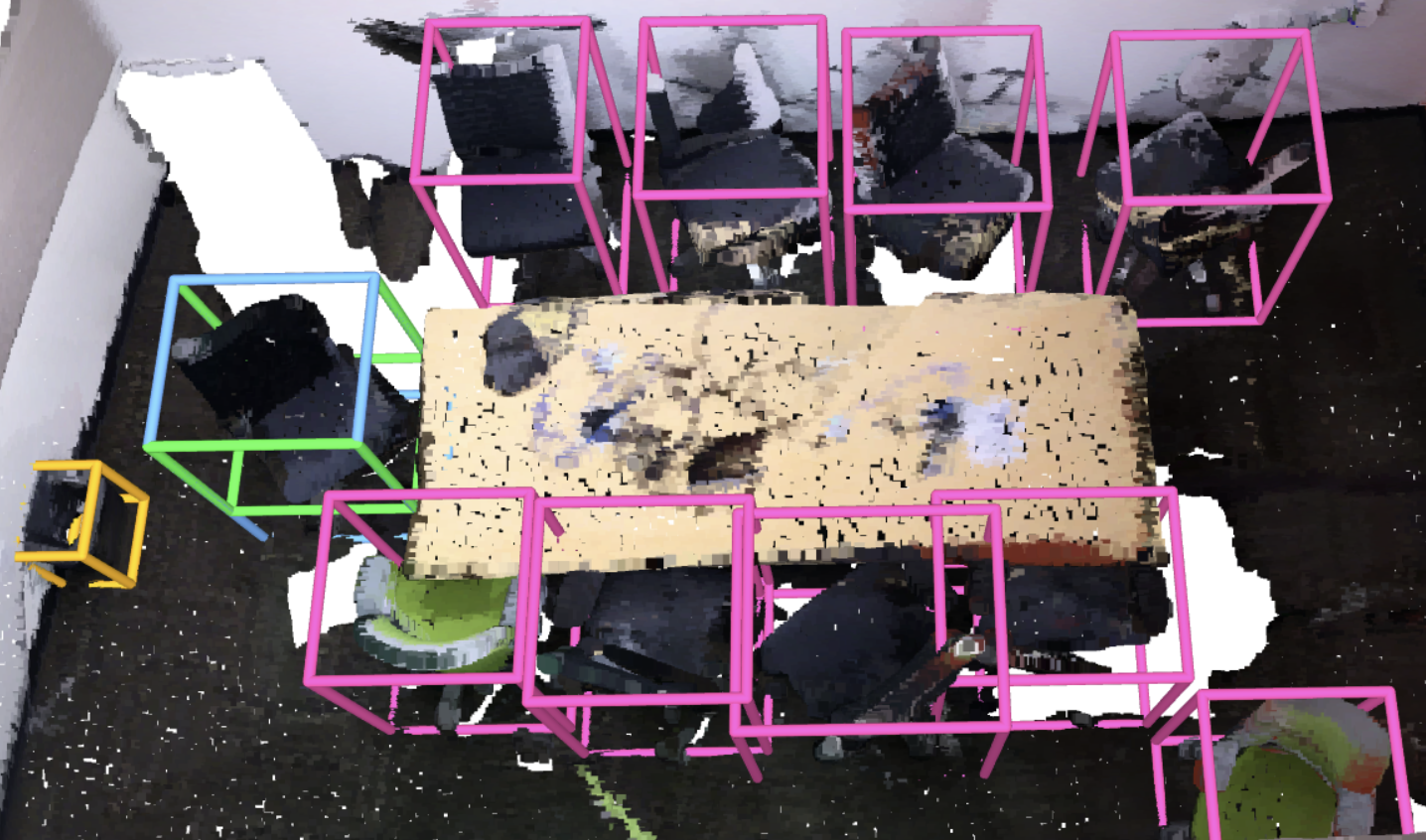} \\
\begin{minipage}{\linewidth}
\begin{lstlisting}
TRASH_CAN_0 = 
  DEF_VAR(labels=["trash can", "trash bin"])
SINK_COUNTER_0 = 
  DEF_VAR(labels=["counter", "sink"])
UNDER(target=TRASH_CAN_0, anchor=SINK_COUNTER_0)
MAX_OF(target=TRASH_CAN_0, score_func="size-z")
SET_TARGET(TRASH_CAN_0)
\end{lstlisting}
\end{minipage} &
\begin{minipage}{\linewidth}
\begin{lstlisting}
MONITOR_0 = DEF_VAR(labels=["monitor"])
TABLE_0 = DEF_VAR(labels=["table"])
ON(target=MONITOR_0, anchor=TABLE_0)
MAX_OF(target=MONITOR_0, score_func="right")
SET_TARGET(MONITOR_0)


\end{lstlisting}
\end{minipage} &
\begin{minipage}{\linewidth}
\begin{lstlisting}
DESK_CHAIR_0 = DEF_VAR(labels=["office chair"])
TRASH_CAN_0 = DEF_VAR(labels=["trash can"])
MIN_OF(target=DESK_CHAIR_0, 
       score_func="distance", 
       anchor=TRASH_CAN_0)
SET_TARGET(DESK_CHAIR_0)

\end{lstlisting}
\end{minipage} \\
\querycell{the tall trash can. the trash can is under the sink counter.} &
\querycell{it is a black computer monitor. the black computer monitor sits on the far right of the desk.} &
\querycell{a black desk chair. it is the one closest to the black bin.} \\
\end{tabular}
\caption{Demonstration of the proposed CSVG with min/max constraints. Distractors (object with the same label as the target): \textcolor[rgb]{1.0,0.3,0.7}{magenta box}. Correct prediction: \textcolor[rgb]{0.3,1.0,0.3}{green box}. Anchor objects discovered by CSVG: \textcolor[rgb]{1.0,0.7,0.0}{orange box}. In the generated programs, the {\ttfamily DEFINE\_VARIABLE} function is abbreviated as {\ttfamily DEF\_VAR}, and we remove the {\ttfamily CONSTRAINT\_} prefix in constraints for simplicity.}
\label{fig:more_example_csvg_with_minmax}
\end{figure*}

\lstdefinestyle{mystyle3}{
    backgroundcolor=\color{backcolour},   
    commentstyle=\color{codegreen},
    keywordstyle=\color{magenta},
    stringstyle=\color{codepurple},
    basicstyle=\ttfamily\linespread{1.15}\scriptsize, 
    breakatwhitespace=false,         
    breaklines=true,                 
    captionpos=b,                    
    keepspaces=true,                 
    showspaces=false,                
    showstringspaces=false,
    showtabs=false,                  
    tabsize=2,
    captionpos=b,
    numbers=none,
}
\lstset{style=mystyle3}

\begin{figure*}
\setlength\tabcolsep{2pt}
\centering
\begin{tabular}{p{0.45\linewidth}p{0.45\linewidth}}
\includegraphics[width=1.0\linewidth]{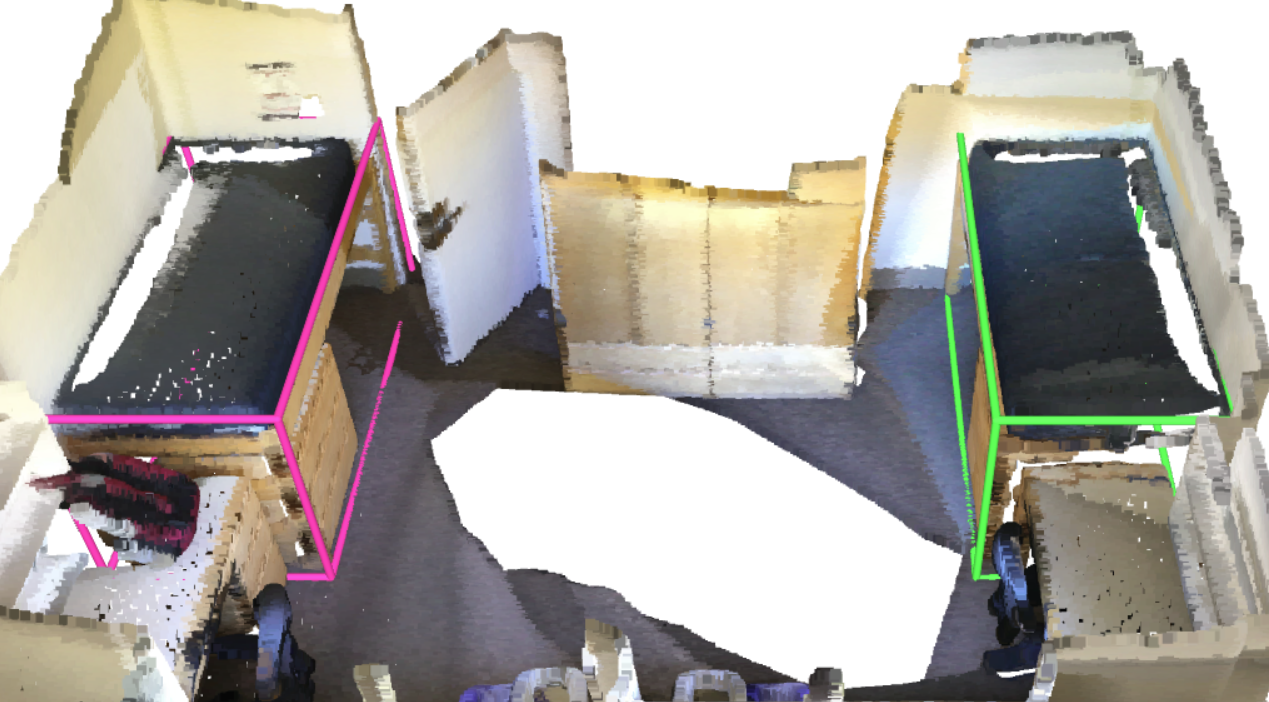} &
\includegraphics[width=1.0\linewidth]{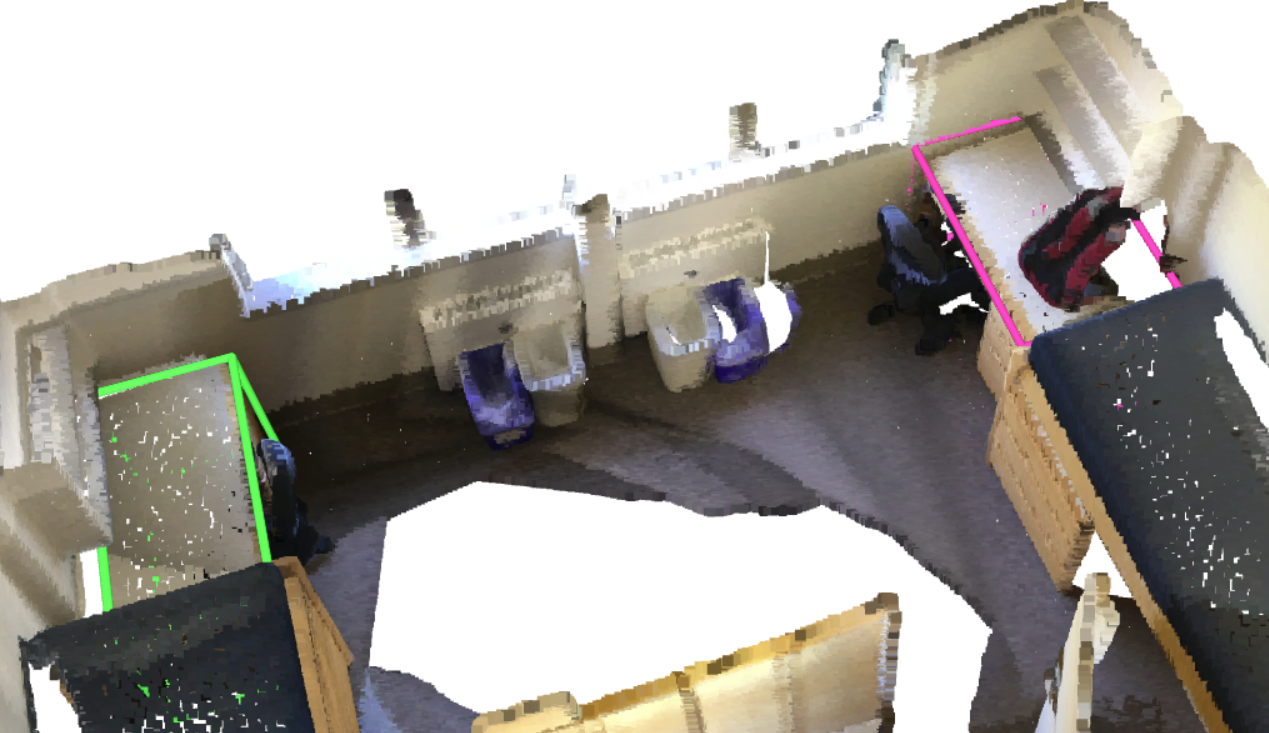} \\
\begin{minipage}{\linewidth}
\begin{lstlisting}
BED_0 = DEF_VAR(labels=["bed"])
DOOR_NEG_0 = DEF_NEG_VAR(labels=["door"])
BESIDE(target=BED_0, anchor=DOOR_NEG_0)
SET_TARGET(BED_0)
\end{lstlisting}
\end{minipage} &
\begin{minipage}{\linewidth}
\begin{lstlisting}
DESK_0 = DEF_VAR(labels=["desk"])
BACKPACK_NEG_0 = DEF_NEG_VAR(labels=["backpack"])
ON(target=BACKPACK_NEG_0, anchor=DESK_0)
SET_TARGET(DESK_0)
\end{lstlisting}
\end{minipage} \\
\querycell{the bed not beside the door.} &
\querycell{the desk without any backpack on it.} \\
\\
\includegraphics[width=1.0\linewidth]{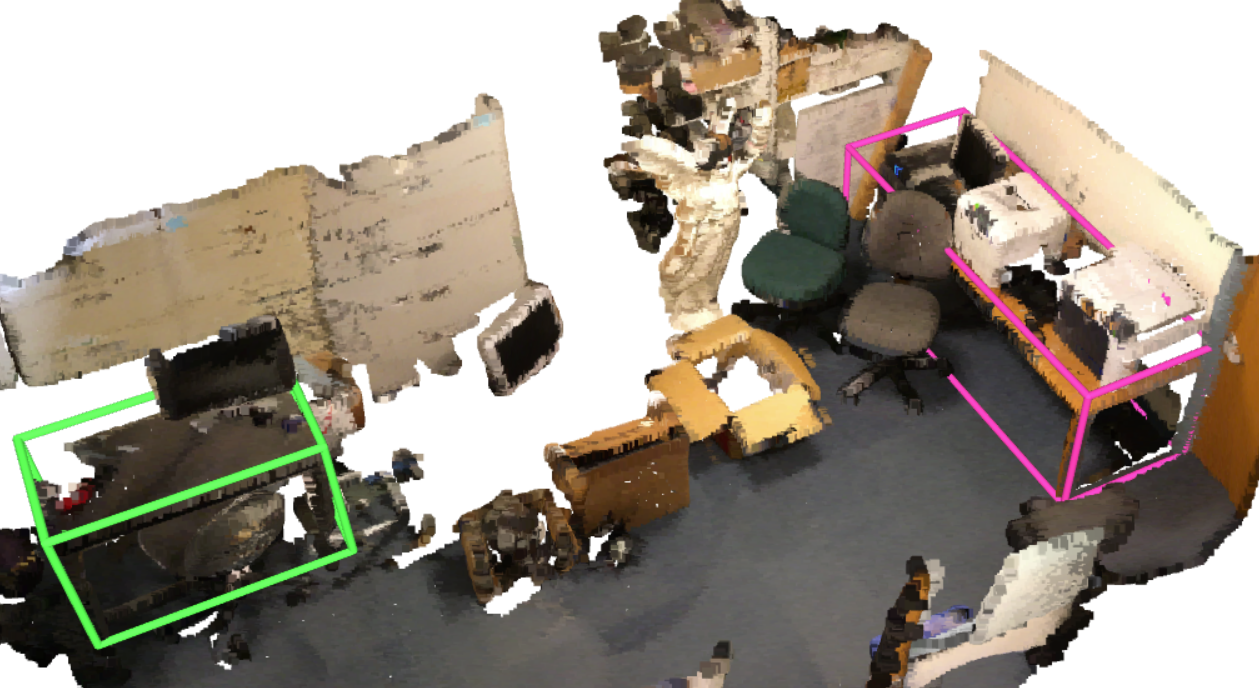} &
\includegraphics[width=1.0\linewidth]{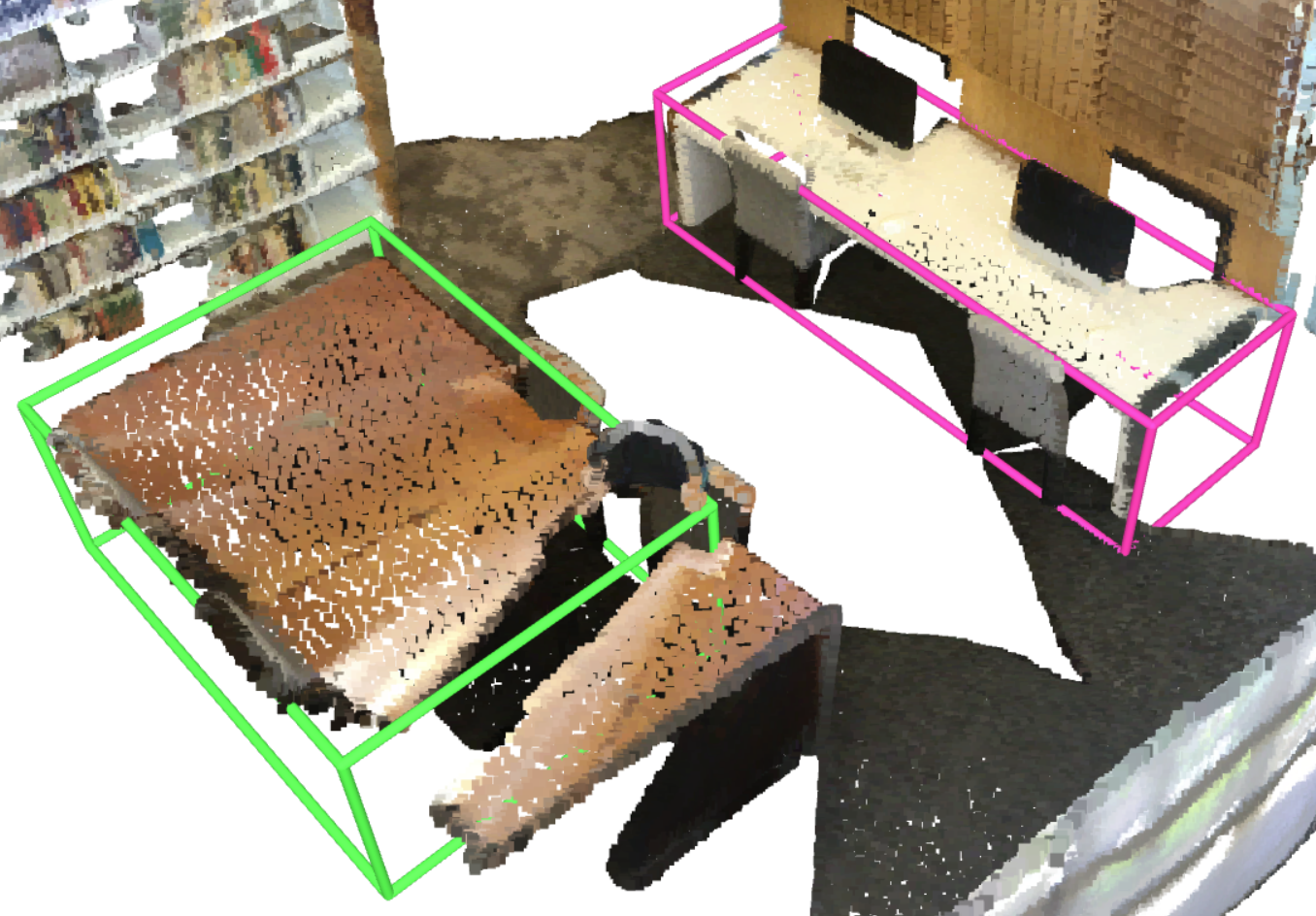} \\
\begin{minipage}{\linewidth}
\begin{lstlisting}
DESK_0 = DEF_VAR(labels=["desk", "table"])
PRINTER_NEG_0 = DEF_NEG_VAR(labels=["printer"])
ON(target=PRINTER_NEG_0, anchor=DESK_0)
SET_TARGET(DESK_0)
\end{lstlisting}
\end{minipage} &
\begin{minipage}{\linewidth}
\begin{lstlisting}
TABLE_0 = DEF_VAR(labels=["table"])
MONITOR_NEG_0 = DEF_NEG_VAR(labels=["monitor"])
ON(target=MONITOR_NEG_0, anchor=TABLE_0)
SET_TARGET(TABLE_0)
\end{lstlisting}
\end{minipage} \\
\querycell{the desk without printers on it.} &
\querycell{the table without monitors on it.} \\
\end{tabular}
\caption{Demonstration of CSVG with negative variables. Distractors (object with the same label as the target): \textcolor[rgb]{1.0,0.3,0.7}{magenta box}. Correct prediction: \textcolor[rgb]{0.3,1.0,0.3}{green box}. Anchor objects discovered by CSVG: \textcolor[rgb]{1.0,0.7,0.0}{orange box}. In the generated programs, the {\ttfamily DEFINE\_VARIABLE} function is abbreviated as {\ttfamily DEF\_VAR}, {\ttfamily DEFINE\_NEGATIVE\_VARIABLE} is replaced by {\ttfamily DEF\_NEG\_VAR}, and we also remove the {\ttfamily CONSTRAINT\_} prefix in constraints for simplicity.}
\label{fig:more_example_csvg_with_neg}
\end{figure*}

\begin{figure*}
\setlength\tabcolsep{2pt}
\centering
\begin{tabular}{p{0.45\linewidth}p{0.45\linewidth}}
\includegraphics[width=1.0\linewidth]{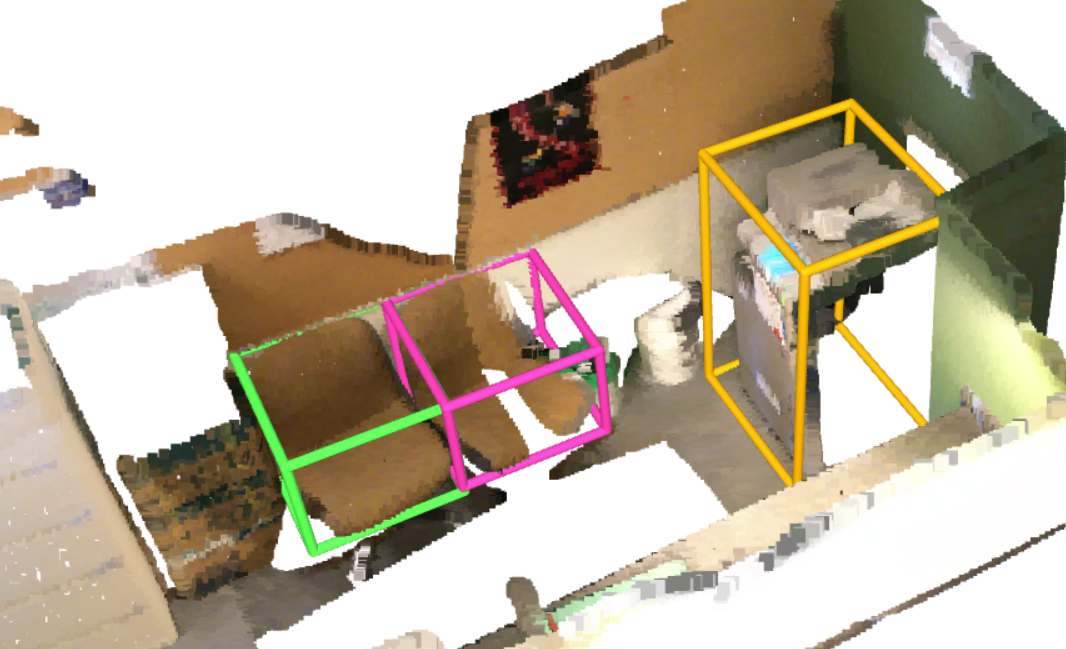} &
\includegraphics[width=1.0\linewidth]{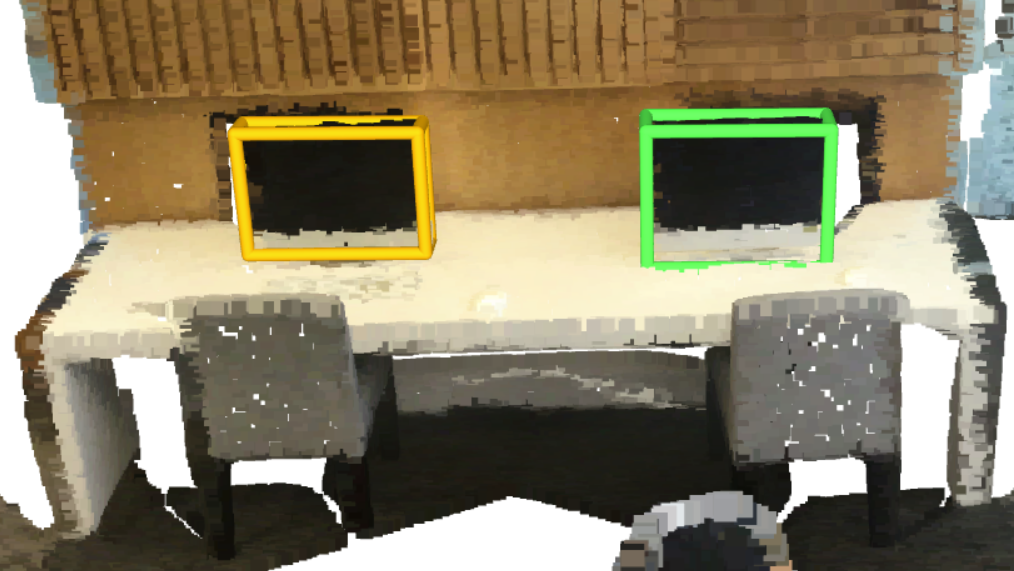} \\
\begin{minipage}{\linewidth}
\begin{lstlisting}
CHAIR_0 = DEFINE_VARIABLE(labels=["chair"])
CHAIR_1 = DEFINE_VARIABLE(labels=["chair"])
COPIER_0 = DEFINE_VARIABLE(labels=["copier"])
CONSTRAINT_MORE(
    target=CHAIR_1, reference=CHAIR_0, 
    anchor=COPIER_0, score_func="distance")
SET_TARGET(CHAIR_1)
\end{lstlisting}
\end{minipage} &
\begin{minipage}{\linewidth}
\begin{lstlisting}
MONITOR_0 = DEFINE_VARIABLE(labels=["monitor"])
MONITOR_1 = DEFINE_VARIABLE(labels=["monitor"])
CONSTRAINT_RIGHT(target=MONITOR_1, anchor=MONITOR_0)
SET_TARGET(MONITOR_1)



\end{lstlisting}
\end{minipage} \\
\querycell{the second chair from the copier.} &
\querycell{the second monitor from the left.}
\end{tabular}

\begin{tabular}{p{0.9\linewidth}}
\includegraphics[width=1.0\linewidth]{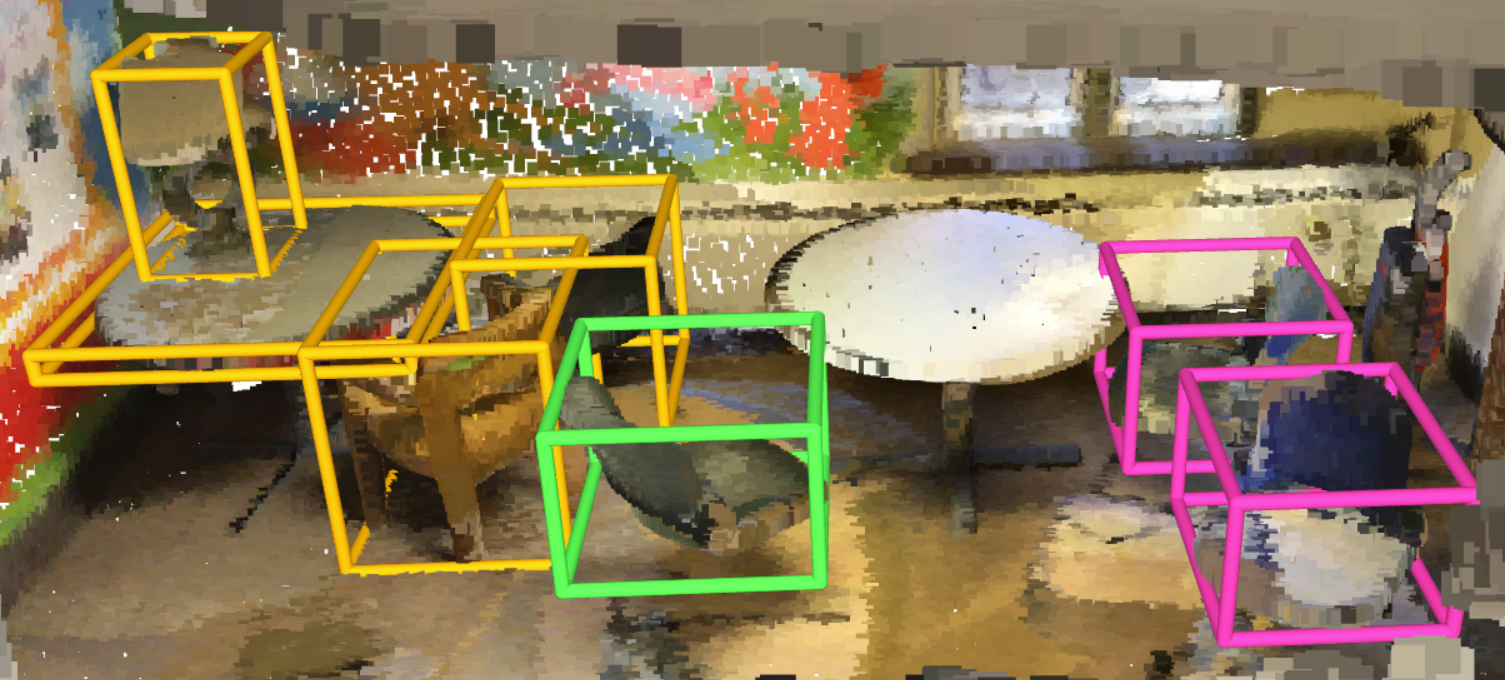} \\
\begin{minipage}{\linewidth}
\begin{lstlisting}
CHAIR_0 = DEFINE_VARIABLE(labels=["chair"])
CHAIR_1 = DEFINE_VARIABLE(labels=["chair"])
CHAIR_2 = DEFINE_VARIABLE(labels=["chair"])
TABLE_WITH_LAMP = DEFINE_VARIABLE(labels=["table"])
LAMP_ON_TABLE = DEFINE_VARIABLE(labels=["lamp"])
CONSTRAINT_ON(target=LAMP_ON_TABLE, anchor=TABLE_WITH_LAMP)
CONSTRAINT_MORE(target=CHAIR_2, reference=CHAIR_1, score_func="distance", anchor=TABLE_WITH_LAMP)
CONSTRAINT_MORE(target=CHAIR_1, reference=CHAIR_0, score_func="distance", anchor=TABLE_WITH_LAMP)
SET_TARGET(CHAIR_2)
\end{lstlisting}
\end{minipage} \\
\querycell{the third chair away from the table with a lamp on it.} \\
\end{tabular}

\caption{Demonstration of the proposed CSVG with counting-based queries. Distractors (object with the same label as the target): \textcolor[rgb]{1.0,0.3,0.7}{magenta box}. Correct prediction: \textcolor[rgb]{0.3,1.0,0.3}{green box}. Anchor objects discovered by CSVG: \textcolor[rgb]{1.0,0.7,0.0}{orange box}.}
\label{fig:more_example_csvg_with_counting}
\end{figure*}

\newcommand{\algovar}[1]{{\fontfamily{lmss}\selectfont \textit{#1}}}
\newcommand{\algofunc}[1]{{\textnormal{\textsc{#1}}}}

\begin{algorithm*}
\caption{Solving CSP with Backtracking}\label{alg:csp_solver}
\SetKwInOut{Input}{Input}\SetKwInOut{Output}{Output}
\Input{CSP generated by the LLM}
\Output{Predicted bounding box for the target, \algovar{target-bbox}}
\algovar{vars} $\gets$ normal (not negative) variables of the CSP\;
\algovar{cons} $\gets$ spatial constraints of the CSP\;
\algovar{minmax-cons} $\gets$ min/max constraints of the CSP\;
\algovar{solutions} $\gets$ empty set\;
\For{\algovar{solution} in \algofunc{Generate-Solutions}(vars)}{
    \For{\algovar{con} in \algovar{cons}}{
        \algovar{neg-var} $\gets$ the negative variable used in \algovar{con}\;
        \eIf{\algovar{neg-var} is empty}{
            \algovar{success} $\gets$ \algofunc{Check-Solution}(\algovar{solution}, \algovar{con})\;
            \If{\textbf{not} \algovar{success}}{
                break and skip this solution\;
            }
        }{
            \For{\algovar{candidate} in \algofunc{Get-Variable-Domain}(\algovar{neg-var})}{
                \algofunc{Update-Solution}(\algovar{solution}, \algovar{neg-var}, \algovar{candidate})\;
                \algovar{success} $\gets$ \algofunc{Check-Solution}(\algovar{solution}, \algovar{con})\;
                \If{\algovar{success}}{
                    break and skip this solution\;
                }
            }
        }
    }
    \If{\algovar{solution} is not skipped}{
        add \algovar{solution} to \algovar{solutions})\;
    }
}
\algovar{valid-solutions} $\gets$ empty set\;
\If{\algovar{solutions} is not empty}{
    \For{\algovar{con} in \algovar{minmax-cons}}{
        \algovar{target-var} $\gets$ \algofunc{Get-Target-Variable}(\algovar{con})\;
        \algovar{anchor-vars} $\gets$ \algofunc{Get-Anchor-Variables}(\algovar{con})\;
        \algovar{con-groups} $\gets$ \algofunc{Group-Solutions}(\algovar{solutions}, \algovar{anchor-vars})\;
        \For{\algovar{con-group} in \algovar{con-groups}}{
            \algovar{solution} $\gets$ \algofunc{Select-Solution-Minmax}(\algovar{con-group}, \algovar{con}, \algovar{target-var})\;
            add \algovar{solution} to \algovar{valid-solutions}\;
        }
    }
}
\algovar{final-solution} $\gets$ \algofunc{Select-Solution-With-Heuristic}(\algovar{valid-solutions})\;
\algovar{target-bbox} $\gets$ \algofunc{Extract-Target-Bounding-Box}(\algovar{final-solution})\;
\end{algorithm*}

\begin{algorithm*}
\caption{Grounding through Local Constraints}\label{alg:non_csp_solver}
\SetKwInOut{Input}{Input}\SetKwInOut{Output}{Output}
\Input{CSP generated by the LLM}
\Output{Predicted bounding box for the target, \algovar{target-bbox}}
\algovar{var-candidates} $\gets$ empty mapping\;
\For{\algovar{csp-var} in normal variables of the CSP}{
    \algovar{var-candidates}[\algovar{csp-var}] = \algofunc{Get-Variable-Domain}(\algovar{csp-var})\;
}
\While{$\exists$ \algovar{constraint}: \algovar{constraint} has not been processed}{
    \For{\algovar{con} in constraints of the CSP}{
        \If{\algovar{con} is already processed}{
            continue\;
        }
        \For{\algovar{anchor-var} in \algofunc{Get-Anchor-Variables}(\algovar{con})}{
            \If{$\exists$ \algovar{any-con}: \algovar{any-con} has not been processed $\land$ \algofunc{Get-Target-Variable}(\algovar{any-con}) = \algovar{anchor-var}}{
                continue\;
            }
        }
        \algovar{target-var} $\gets$ \algofunc{Get-Target-Variable}(\algovar{con})\;
        \algovar{var-candidates}[\algovar{target-var}] $\gets$ \algofunc{Filter-Variable-Candidates}(\algovar{var-candidates}, \algovar{con})\;
    }
    \If{no new constraints are marked as processed}{
        return failure\;
    }
}
\algovar{target-var} $\gets$ target variable of the CSP\;
\algovar{target-bbox} $\gets$ \algofunc{Get-Instance-Bounding-Box}(first element in \algovar{var-candidates}[\algovar{target-var}])\;
\end{algorithm*}

\lstset{style=mystyle}

\begin{lstlisting}[float=*, caption={LLM prompt: system message.}, label={lst:system_msg}]
<[SYSTEM]>
You are given the 3D visual grounding task. The input is a language query describing some object(s) in the 3D scene. Besides that, you are also given a list of object labels that are relevant to this query. Any other objects should not be considered. 
Your task is to generate a Python program that locate the target object specified by the query. You don't have to use all the relevant object labels given to you. Some of them may actually be irrelevant. What's important is, you should always use the labels of the relevant objects given to you.
The Python program you generate should solve a CSP (Constraint Satisfication Problem). Solving the CSP will result in the correct grounding of the target objects. The variables are objects and constraints are spatial relationships among objects or appearance information of objects.
You should first try to do some spatial reasoning of the input query and simplify the logic. Some relations between objects can be converted into simpler ones. You should always first try to reduce the complexity of the input query. E.g. an object that has one object on the left and another on the right is between those two objects. So you should use a single relation "between", instead of two relations "left" and "right". There are many similar cases.
There are two types of variables: normal variables (DEFINE_VARIABLE), which represent a single object, and negative variables (DEFINE_NEGATIVE_VARIABLE), which indicate the abscence of an object satisfying some spatial relations. You should only use negative variables when the query mentions some negative conditions with e.g., 'do not' or 'without'.
Your should only use the following predefined functions in your program. No other functions are allowed.
<[REGISTERED_FUNCTIONS_PLACEHOLDER]>
Some of the predefined functions above have a "score_func" parameter, which specifies how to compare objects. The following score functions are available.
<[REGISTERED_SCORE_FUNCTIONS_PLACEHOLDER]>
Your output should be a valid Python program, which means that any additional explanation should be comments (start with #). You only have to output the python code. Please DO NOT follow markdown conventions. You DO NOT have to enclose the code with ```.
Some extra tips:
- "North", "south", "east" and "west" should always be translated into "near".
- If something is at the side of a room, it should be far from the room center.
- If an object you need is not in the relevant object list, please ignore information about that object, and do not use it.
- You should not ignore relations that can be represented with the given relevant object labels and available functions.
- Relations like "farthest", "northwestern-most" in reference to the room should be translated to being farthest to the room center.
- You should identify only ONE target, to which the given 3D visual grounding query is referring to. You should always set the target at the very end. No function calls are allowed after setting the target.
- Please DO NOT define variables that you will not use later. Every variables defined must be set in relation with at least one other variable.
- Sometimes multiple labels refer to similar objects. For example, different queries may use "desk" or "table" to refer to the same thing. So when you need a "desk", and only "table" are available, you should use table, and vice versa.
- For each object, you either define a normal variable or a negative variable. Please do not define both.
\end{lstlisting}

\begin{lstlisting}[float=*, caption={LLM prompt: in-context example 1.}, label={lst:in_context_example_1}, language=Python, moredelim={[is]{STARTRC}{ENDRC}}]
STARTRC<[USER]>

QUERY:
this is a grey chair. it is at the end of the shorter coffee table closest to the window.

RELEVANT OBJECT LABELS:
[0] chair
[1] window
[2] coffee table

<[ASSISTANT]>ENDRC

# define variables for the CSP
CHAIR_0 = DEFINE_VARIABLE(labels=["chair"])
WINDOW_0 = DEFINE_VARIABLE(labels=["window"])
COFFEE_TABLE_0 = DEFINE_VARIABLE(labels=["coffee table"])

# add a constraint: the grey chair is at the end of the shorter coffee table
CONSTRAINT_NEAR(target=CHAIR_0, anchor=COFFEE_TABLE_0)

# add a constraint: the coffee table is the one closest to the window
CONSTRAINT_MIN_OF(target=COFFEE_TABLE_0, anchor=WINDOW_0, score_func="distance")

# finally set the CSP variable representing the target object
SET_TARGET(CHAIR_0)
\end{lstlisting}

\begin{lstlisting}[float=*, caption={LLM prompt: in-context example 2.}, label={lst:in_context_example_2}, language=Python, moredelim={[is]{STARTRC}{ENDRC}}]
STARTRC<[USER]>

QUERY:
the desk is one of three in the row. there is a desk to the right of it, and a desk to the left of it.

RELEVANT OBJECT LABELS:
[0] desk

<[ASSISTANT]>ENDRC

# since there are three desks in total, I'll create an object to represent each of them
DESK_MIDDLE = DEFINE_VARIABLE(labels=["desk"])
DESK_LEFT = DEFINE_VARIABLE(labels=["desk"])
DESK_RIGHT = DEFINE_VARIABLE(labels=["desk"])

# the target desk has one desk to the right, one to the left, therefore by a simple reasoning I know that it is the desk between two other desks
CONSTRAINT_BETWEEN(target=DESK_MIDDLE, anchors={DESK_LEFT, DESK_RIGHT})

# set the variable for the target object
SET_TARGET(DESK_MIDDLE)
\end{lstlisting}

\begin{lstlisting}[float=*, caption={LLM prompt: in-context example 3.}, label={lst:in_context_example_3}, language=Python, moredelim={[is]{STARTRC}{ENDRC}}]
STARTRC<[USER]>

QUERY:
it is the tallest chair at the center of the room.

RELEVANT OBJECT LABELS:
[0] room center
[1] chair

<[ASSISTANT]>ENDRC

# define a variable for the target chair object
CHAIR_0 = DEFINE_VARIABLE(labels=["chair"])

# define the room center variable
ROOM_CENTER = DEFINE_VARIABLE(labels=["room center"])

# add a constraint: the chair is near the room center
# we pass the set of all chairs in order to keep only chairs at the room center
CONSTRAINT_NEAR(target=CHAIR_0, anchor=ROOM_CENTER)

# add a constraint: the chair is the tallest one, i.e. the one with largest height
# we find the chair with the largest extent along z-axis in the set of chairs at the room center
CONSTRAINT_MAX_OF(target=CHAIR_0, score_func="size-z")

# set the final target
SET_TARGET(CHAIR_0)
\end{lstlisting}

\begin{lstlisting}[float=*, caption={LLM prompt: in-context example 4.}, label={lst:in_context_example_4}, language=Python, moredelim={[is]{STARTRC}{ENDRC}}]
STARTRC<[USER]>

QUERY:
this is a blue trash can that is beside a orange one. it is not the trash can far from other trash cans.

RELEVANT OBJECT LABELS:
[0] trash can

<[ASSISTANT]>ENDRC

# define a variable for the blue trash can (target)
TRASH_CAN_BLUE = DEFINE_VARIABLE(labels=["trash can"])

# define a variable for the orange trash can
TRASH_CAN_ORANGE = DEFINE_VARIABLE(labels=["trash can"])

# define a variable for the far away trash can
TRASH_CAN_FAR = DEFINE_VARIABLE(labels=["trash can"])

# the blue trash can should be beside the orange one
CONSTRAINT_NEAR(target=TRASH_CAN_BLUE, anchor=TRASH_CAN_ORANGE)

# the far away trash can should be far from the other trash cans
CONSTRAINT_FAR(target=TRASH_CAN_FAR, anchor=TRASH_CAN_BLUE)
CONSTRAINT_FAR(target=TRASH_CAN_FAR, anchor=TRASH_CAN_ORANGE)

# finally set the variable for the target object
SET_TARGET(TRASH_CAN_BLUE)
\end{lstlisting}

\begin{lstlisting}[float=*, caption={LLM prompt: in-context example 5.}, label={lst:in_context_example_5}, language=Python, moredelim={[is]{STARTRC}{ENDRC}}]
STARTRC<[USER]>

QUERY:
it is the smallest cup on the largest table.

RELEVANT OBJECT LABELS:
[0] cup
[1] table

<[ASSISTANT]>ENDRC

# define a variable for the largest table
TABLE_0 = DEFINE_VARIABLE(labels=["table"])

# define a variable for the smallest cup on the largest table
CUP_0 = DEFINE_VARIABLE(labels=["cup"])

# constraint: the largest table
CONSTRAINT_MAX_OF(target=TABLE_0, score_func="size")

# constraint: the cup is on the table
CONSTRAINT_ON(target=CUP_0, anchor=TABLE_0)

# constraint: the smallest cup
CONSTRAINT_MIN_OF(target=CUP_0, score_func="size")

# set the variable for the target object
SET_TARGET(CUP_0)
\end{lstlisting}

\begin{lstlisting}[float=*, caption={LLM prompt: in-context example 6.}, label={lst:in_context_example_6}, language=Python, moredelim={[is]{STARTRC}{ENDRC}}]
STARTRC<[USER]>

QUERY:
this is the third chair from the left.

RELEVANT OBJECT LABELS:
[0] chair

<[ASSISTANT]>ENDRC

# this is an example of how to handle counting-based query

# since the target is the third chair from the left, there must be at least three chairs in total
# so we define a variable for each of them
CHAIR_0 = DEFINE_VARIABLE(labels=["chair"])
CHAIR_1 = DEFINE_VARIABLE(labels=["chair"])
CHAIR_2 = DEFINE_VARIABLE(labels=["chair"])

# constraint: the target chair has a chair on the left, which also has another chair to its left
CONSTRAINT_LEFT(target=CHAIR_1, anchor=CHAIR_0)
CONSTRAINT_LEFT(target=CHAIR_2, anchor=CHAIR_1)

# set the variable for the target object
SET_TARGET(CHAIR_0)
\end{lstlisting}

\begin{lstlisting}[float=*, caption={LLM prompt: in-context example 7.}, label={lst:in_context_example_7}, language=Python, moredelim={[is]{STARTRC}{ENDRC}}]
STARTRC<[USER]>

QUERY:
this is the chair third closest to the corner of the room.

RELEVANT OBJECT LABELS:
[0] chair
[1] room corner

<[ASSISTANT]>ENDRC

# this is another example of how to handle counting-based query
# it is also a comparison-based query

# since the target is the third chair from the room corner, there must be at least three chairs in total
# so we define a variable for each of them
CHAIR_0 = DEFINE_VARIABLE(labels=["chair"])
CHAIR_1 = DEFINE_VARIABLE(labels=["chair"])
CHAIR_2 = DEFINE_VARIABLE(labels=["chair"])

# define a variable for the corner of the room
ROOM_CORNER = DEFINE_VARIABLE(labels=["room corner"])

# constraint: the target chair (CHAIR_2) is farther from the room corner than another chair (CHAIR_1), which (CHAIR_1) again is farther from the room corner than yet a third chair (CHAIR_0)
CONSTRAINT_MORE(target=CHAIR_2, reference=CHAIR_1, anchor=ROOM_CORNER, score_func="distance")
CONSTRAINT_MORE(target=CHAIR_1, reference=CHAIR_0, anchor=ROOM_CORNER, score_func="distance")

# set the variable for the target object
SET_TARGET(CHAIR_2)
\end{lstlisting}

\begin{lstlisting}[float=*, caption={LLM prompt: in-context example 8.}, label={lst:in_context_example_8}, language=Python, moredelim={[is]{STARTRC}{ENDRC}}]
STARTRC<[USER]>

QUERY:
there are some windows on the wall. this is the window in the middle.

RELEVANT OBJECT LABELS:
[0] wall
[1] window

<[ASSISTANT]>ENDRC

# define variables for the window and the wall
WALL_0 = DEFINE_VARIABLE(labels=["wall"])
WINDOW_0 = DEFINE_VARIABLE(labels=["window"])

# the windows should be near the wall
CONSTRAINT_NEAR(target=WINDOW_0, anchor=WALL_0)

# the target window should be in the middle, i.e., the "distance-to-center" score must be minimized
CONSTRAINT_MIN_OF(target=WINDOW_0, score_func="distance-to-center")

# set the variable for the target object
SET_TARGET(WINDOW_0)
\end{lstlisting}

\begin{lstlisting}[float=*, caption={LLM prompt: in-context example 9.}, label={lst:in_context_example_9}, language=Python, moredelim={[is]{STARTRC}{ENDRC}}]
STARTRC<[USER]>

QUERY:
this is a chair. it does not have a teddy bear sitting on it.

RELEVANT OBJECT LABELS:
[0] chair
[1] teddy bear

<[ASSISTANT]>ENDRC

# this is an example of using negative variables

# define variables for the chair and the teddy bear
CHAIR_0 = DEFINE_VARIABLE(labels=["chair"])

# define a negative variable to express that the chair has no teddy bear on it
TEDDY_BEAR_NEG = DEFINE_NEGATIVE_VARIABLE(labels=["teddy bear"])

# first create a normal constraint: the teddy bear is on the chair
CONSTRAINT_ON(target=TEDDY_BEAR_NEG, anchor=CHAIR_0)

# set the variable for the target object
SET_TARGET(CHAIR_0)
\end{lstlisting}

\begin{lstlisting}[float=*, caption={LLM prompt: in-context example 10.}, label={lst:in_context_example_10}, language=Python, moredelim={[is]{STARTRC}{ENDRC}}]
STARTRC<[USER]>

QUERY:
it is the cup not on the table.

RELEVANT OBJECT LABELS:
[0] cup
[1] table

<[ASSISTANT]>ENDRC

# this is another example of using negative variables

# define the cup variable, which is the target
CUP_0 = DEFINE_VARIABLE(labels=["cup"])

# define the table variable as negative, since the cup variable is the target and we want to locate it
TABLE_NEG_0 = DEFINE_NEGATIVE_VARIABLE(labels=["table"])

# the cup is not on the table
CONSTRAINT_ON(target=CUP_0, anchor=TABLE_NEG_0)

# set the variable for the target object
SET_TARGET(CUP_0)
\end{lstlisting}

\begin{lstlisting}[float=*, caption={LLM prompt: in-context example 11.}, label={lst:in_context_example_11}, language=Python, moredelim={[is]{STARTRC}{ENDRC}}]
STARTRC<[USER]>

QUERY:
this is a recycling trash can. its blue and white in color and is on the left of a box. it is also below a wooden counter.

RELEVANT OBJECT LABELS:
[0] trash can
[1] box
[2] counter
[3] plastic trash bin
[4] wastebin

<[ASSISTANT]>ENDRC

# define variables for the trash can. since we have multiple labels denoting trash cans (with synonyms), we include all of them in the label.
TRASH_CAN_0 = DEFINE_VARIABLE(labels=["trash can", "plastic trash bin", "wastebin"])

# define variables for the box and the counter
BOX_0 = DEFINE_VARIABLE(labels=["box"])
COUNTER_0 = DEFINE_VARIABLE(labels=["counter"])

# add a constraint: the trash can is on the left of a box
CONSTRAINT_LEFT(target=TRASH_CAN_0, anchor=BOX_0)

# add a constraint: the trash can is also below a wooden counter
CONSTRAINT_UNDER(target=TRASH_CAN_1, anchor=COUNTER_0)

# set the CSP variable for the target object
SET_TARGET(TRASH_CAN_0)
\end{lstlisting}

\end{document}